\PassOptionsToPackage{table}{xcolor}
\documentclass{article} %
\usepackage{realvr_preprint,times}

\usepackage{amsmath,amsfonts,bm}

\def\eqref#1{equation~\ref{#1}}

\def\1{\bm{1}}

\DeclareMathAlphabet{\mathsfit}{\encodingdefault}{\sfdefault}{m}{sl}
\SetMathAlphabet{\mathsfit}{bold}{\encodingdefault}{\sfdefault}{bx}{n}

\usepackage{hyperref}
\hypersetup{colorlinks=false,pdfborder={0 0 0}}
\usepackage{url}
\usepackage{graphicx}
\usepackage{pifont}
\usepackage{adjustbox}
\usepackage{wrapfig}
\usepackage{placeins}
\usepackage{flafter}
\usepackage{booktabs}
\usepackage{tabularx}
\usepackage{array}
\usepackage{multirow}
\usepackage{xcolor}
\usepackage{tikz}
\usepackage{tcolorbox}
\newcolumntype{Y}{>{\raggedright\arraybackslash}X}
\newcolumntype{C}{>{\centering\arraybackslash}X}
\definecolor{tablegray}{gray}{0.96}
\definecolor{lightblue}{rgb}{0.796, 0.894, 0.9808}
\definecolor{natureblue}{RGB}{0,76,153}
\definecolor{naturepurple}{RGB}{115,65,130}
\definecolor{naturegray}{RGB}{248,248,248}
\definecolor{natureteal}{RGB}{32,128,128}
\definecolor{naturemagenta}{RGB}{170,40,120}
\definecolor{brightpurple}{RGB}{160,105,255}
\definecolor{brightgreen}{RGB}{55,190,95}

\DeclareRobustCommand{\realvr}{\mbox{ReaLVR}}
\newtcolorbox{takeawaybox}[1][]{
  colback=naturegray,
  colframe=natureblue!15,
  title={\textsf{#1}},
  coltitle=black,
  boxrule=1pt,
  arc=8pt,
  left=4pt, right=4pt, top=0pt, bottom=0pt,
}

\newcommand{\appendixTableSetup}{%
  \small
  \setlength{\tabcolsep}{4pt}%
  \setlength{\aboverulesep}{0.28ex}%
  \setlength{\belowrulesep}{0.28ex}%
  \setlength{\extrarowheight}{0pt}%
  \renewcommand{\arraystretch}{1.10}%
}
\newcommand{\appendixWideTableSetup}{%
  \footnotesize
  \setlength{\tabcolsep}{2.6pt}%
  \setlength{\aboverulesep}{0.28ex}%
  \setlength{\belowrulesep}{0.28ex}%
  \setlength{\extrarowheight}{0pt}%
  \renewcommand{\arraystretch}{1.08}%
}

\title{Rethinking Latent Visual Reasoning: Grounding Latent Reasoning in Visual Evidence}

\def\realvrWebsiteURL{https://xixiaouab.github.io/projects/ReaLVR/}
\def\realvrCodeURL{https://github.com/xixiaouab/ReaLVR-code}
\def\realvrModelURL{https://huggingface.co/MarkShaw99/ReaLVR}
\newcommand{\amazonHeaderMark}{{\normalfont\small Preprint.}}
\newcommand{\realvrVersionDate}{2026-09-28}
\newcommand{\realvrHeaderDate}{{\normalfont\small\itshape\realvrVersionDate}}
\renewcommand{\headrulewidth}{0.4pt}

\makeatletter
\long\def\@author{%
  {\bfseries
  \mbox{Xi Xiao$^{1,2}$\thanks{Work done during an internship at Amazon AGI.}\kern0.25em},
  \mbox{Tianchen Zhao$^{2}$},
  \mbox{Youngeun Kim$^{2}$},
  \mbox{Zhuowei Li$^{2}$},
  \mbox{Linghan Xu$^{2}$},\\
  \mbox{Jiaye Wu$^{2}$},
  \mbox{Zheng Zhang$^{2}$},
  \mbox{Xiang Xu$^{2}$},
  \mbox{Xuanbai Chen$^{2}$},\\
  \mbox{Farhan Tejani$^{2}$},
  \mbox{Jakub Zablocki$^{2}$},
  \mbox{Julia Xu$^{2}$},
  \mbox{Yifan Xing$^{2}$}\par}
  \vspace{4pt}
  {\normalfont\fontsize{10}{12.5}\selectfont
  $^{1}$University of Alabama at Birmingham\quad $^{2}$Amazon AGI\par}
  \vspace{3pt}
  {\normalfont\fontsize{10}{12.5}\selectfont
  \ding{41}\hspace{0.3em}\href{mailto:xxiao@uab.edu}{\texttt{xxiao@uab.edu}},\quad\href{mailto:tianchz@amazon.com}{\texttt{tianchz@amazon.com}}\par}
  \vspace{5pt}
  {\normalfont\fontsize{10}{12.5}\selectfont
  \raisebox{-0.4ex}{\includegraphics[height=1em]{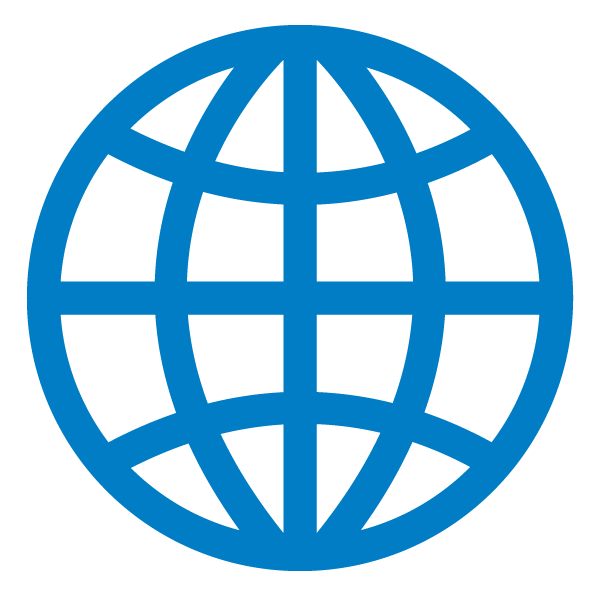}}\hspace{0.3em}%
  \ifx\realvrWebsiteURL\empty\texttt{Website}\else
    \href{\realvrWebsiteURL}{\texttt{Website}}\fi
  \hspace{1em}%
  \raisebox{-0.4ex}{\includegraphics[height=1em]{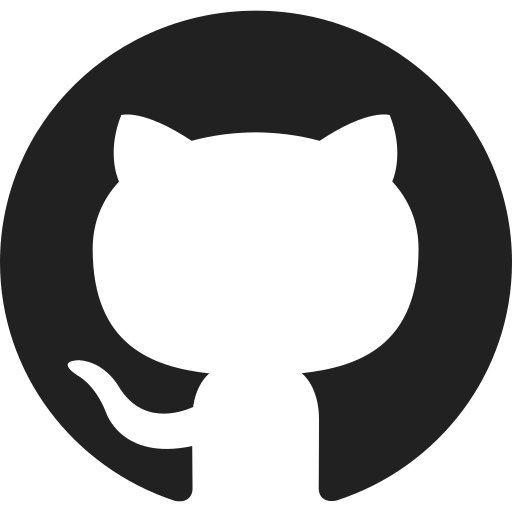}}\hspace{0.3em}%
  \ifx\realvrCodeURL\empty\texttt{Code}\else
    \href{\realvrCodeURL}{\texttt{Code}}\fi
  \hspace{1em}%
  \raisebox{-0.4ex}{\includegraphics[height=1em]{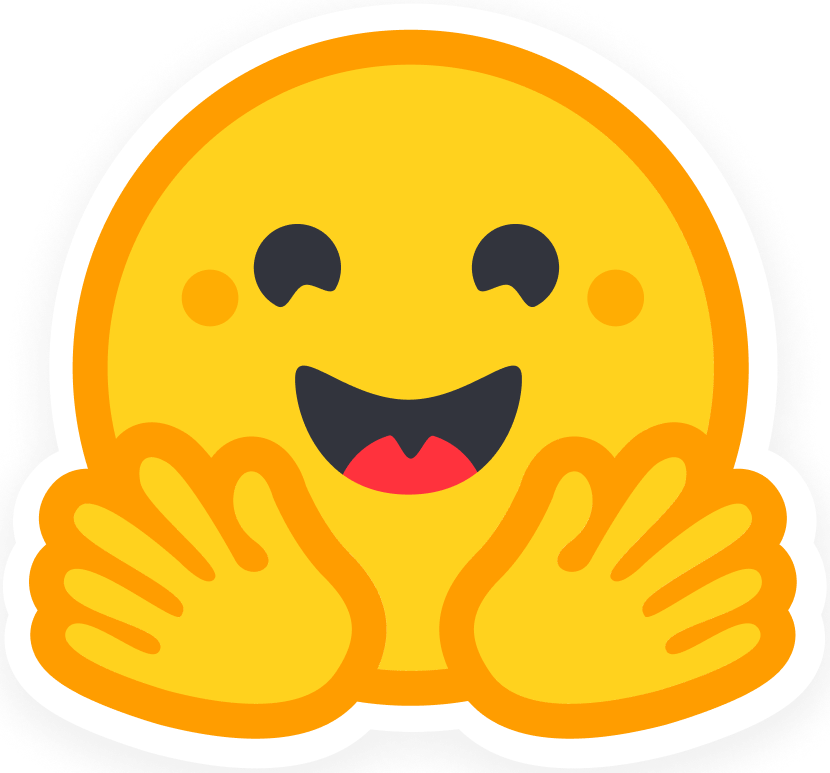}}\hspace{0.3em}%
  \ifx\realvrModelURL\empty\texttt{Model}\else
    \href{\realvrModelURL}{\texttt{Model}}\fi\par}%
}

\makeatother

\makeatletter
\renewcommand{\@maketitle}{%
  \vbox{\hsize\textwidth
    {\raggedright\hyphenpenalty=10000\exhyphenpenalty=10000
     \fontsize{18}{21.5}\selectfont\bfseries\@title\par}
    \vskip 13pt
    {\raggedright\fontsize{10}{13}\selectfont\@author\par}
    \vskip 18pt
  }%
}
\makeatother

\begin{document}

\setcounter{topnumber}{3}
\setcounter{bottomnumber}{2}
\setcounter{totalnumber}{5}
\renewcommand{\topfraction}{0.92}
\renewcommand{\bottomfraction}{0.85}
\renewcommand{\textfraction}{0.08}
\renewcommand{\floatpagefraction}{0.85}
\setlength{\textfloatsep}{8pt plus 2pt minus 2pt}
\setlength{\floatsep}{6pt plus 2pt minus 2pt}
\setlength{\intextsep}{6pt plus 2pt minus 2pt}
\setlength{\abovecaptionskip}{4pt}
\setlength{\parskip}{2pt}
\makeatletter
\setlength{\@fptop}{0pt}
\setlength{\@fpsep}{12pt}
\makeatother

\maketitle
\fancyhead[L]{\amazonHeaderMark}
\fancyhead[R]{\realvrHeaderDate}
\vspace{-6pt}

\begin{abstract}
Latent visual reasoning (LVR) enables multimodal large language models (MLLMs) to perform intermediate computation in continuous latent tokens rather than expressing every reasoning step in words. However, unlike textual CoT, latent reasoning is not directly observable, making it difficult to supervise what latent tokens learn. In this work, we first conduct a thorough analysis of latent-token behavior and identify a \textit{latent evidence-credit gap}: latent tokens respond only weakly to image perturbations that alter the correct answer. We hypothesize that this issue stems from the lack of explicit supervision during GRPO training. These findings suggest that a final-answer reward provides too little guidance on what visual evidence to preserve or how credit should be assigned across latent tokens. To bridge this gap, we propose ReaLVR, which brings visual-evidence supervision to the model’s own free-running latent trajectories. ReaLVR contrasts correct and model-generated wrong answers to determine where stronger supervision is needed, and relevant and mismatched visual evidence to specify what to preserve. Across three model families, ReaLVR consistently outperforms evaluated LVR baselines, achieving the highest five-task average of 63.7\% on Qwen2.5-VL-7B. Crucially, we are the first to scale visual reasoning in latent space, showing that our framework continues to deliver robust improvements at frontier model scales up to 235B. Further analyses show more question-sensitive latent-token positions, stronger alignment with relevant visual regions, and greater fixed-context dependence on the most attended latent tokens.
\end{abstract}

\section{Introduction}
\label{sec:intro}

Latent visual reasoning (LVR) has emerged as an alternative to textual chain-of-thought reasoning in multimodal large language models (MLLMs), performing intermediate computation through continuous latent tokens rather than expressing every reasoning step in words~\citep{li2025latentvisualreasoning,wang2025monet,yang2025mirage,dong2025ilvr,hu2026colt,jeon2026visionaligned,li2026latent}. This approach is especially appealing for problems involving spatial relationships and fine-grained visual details that are difficult to describe step by step. However, the underlying mechanisms of LVR remain unclear: what information the generated latent tokens encode, whether they respond to visual evidence that changes the correct answer, and how they contribute to the final answer? Assessing only final-answer accuracy is insufficient to address these questions.

\begin{figure}[!htbp]
  \centering
  \includegraphics[width=\linewidth]{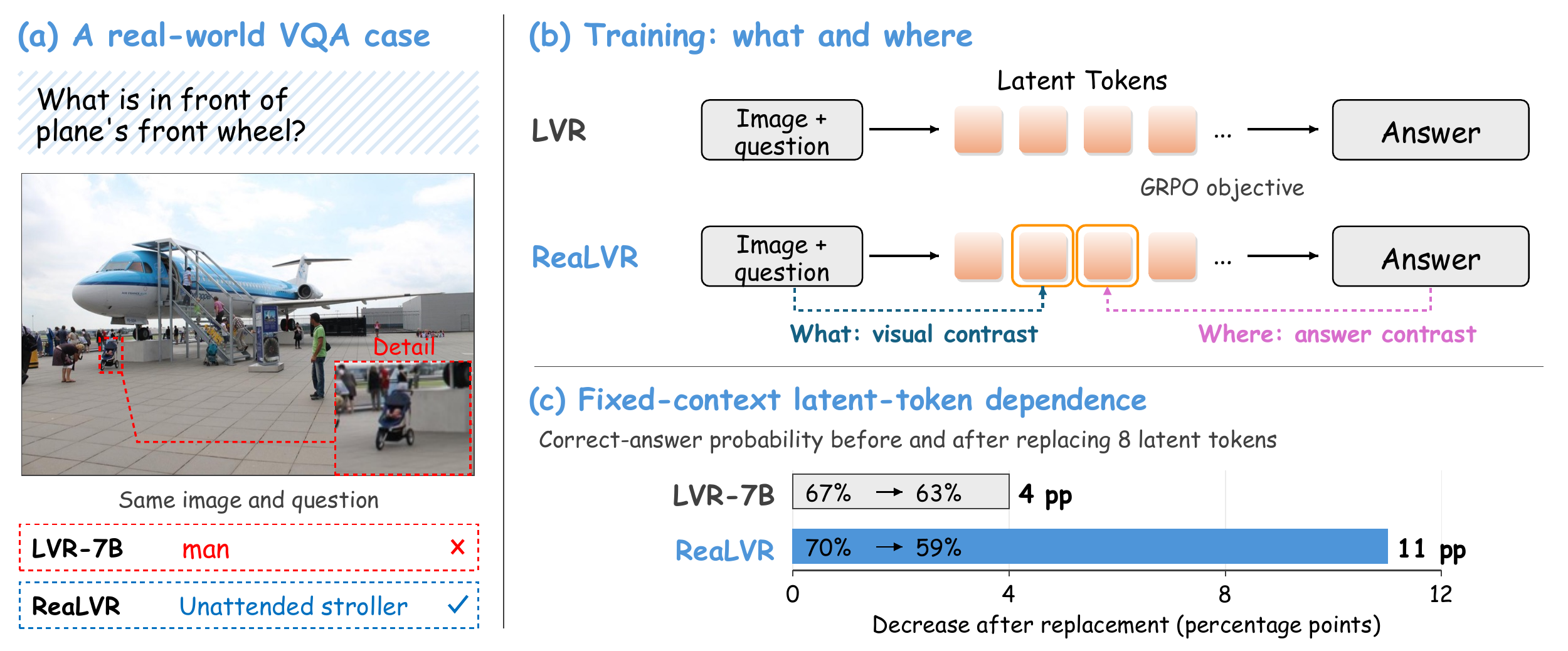}
  \caption{\small\textbf{ReaLVR connects visual evidence to latent reasoning.}
  (a) ReaLVR identifies the stroller missed by LVR-7B in a complex scene; dashed boxes mark image details.
  (b) Visual and answer contrast determine \emph{what to preserve} and \emph{where to supervise}, during training only.
  (c) Replacing the top-8 tokens ranked by answer-to-token attention, with other states fixed, decreases correct-answer probability by 4 and 11 percentage points. The larger ReaLVR drop indicates greater local answer dependence.}
  \label{fig:realvr-teaser}
\end{figure}

Our controlled behavioral and representation tests suggest that vanilla LVR does not reliably preserve answer-relevant visual evidence in its generated trajectory. Motivated by these observations, we examine how the generated trajectory is supervised. In the SFT stage, latent visual states are supervised to match target visual features. During RL and inference, the model instead generates a \emph{free-running latent trajectory} without these targets. Standard GRPO~\citep{shao2024deepseekmath} only optimizes the generated text rather than the latent trajectory itself. Consequently, answer-level feedback provides no direct signal indicating which positions need stronger visual supervision or what evidence they should preserve. We term this missing connection the \emph{latent evidence-credit gap}.

To bridge this gap, we propose ReaLVR, which directly trains free-running latent tokens to preserve answer-relevant visual evidence. ReaLVR regenerates the trajectory with the current model and retains gradients through its generation. To decide where stronger visual supervision is needed, it compares how the ground-truth answer and wrong answers sampled from the behavior policy attend to each latent token. To determine what the supervised tokens should preserve, ReaLVR contrasts answer-relevant visual evidence with mismatched evidence. Figure~\ref{fig:realvr-teaser} provides an overview.

We evaluate ReaLVR on five benchmarks with six backbones spanning three model families (\texttt{Qwen2.5-VL}/\texttt{Qwen3-VL}, \texttt{InternVL3}, and \texttt{Gemma-3}). ReaLVR achieves the highest five-task average among the evaluated latent-reasoning methods on Qwen2.5-VL-7B, reaching 63.7\%. On Qwen3-VL-8B, ReaLVR reaches an average of $61.2\%$, exceeding the evaluated latent-reasoning baselines. The gains extend to Qwen3-VL-30B, where ReaLVR reaches 65.2\%, 1.1 points above LVR. On \texttt{Qwen3-VL-235B}, ReaLVR also improves over LVR-SFT on all three evaluated benchmarks. To our knowledge, this is the first demonstration of visual reasoning in latent space trained on a 235B-parameter multimodal backbone. 

The contributions of this work are threefold. \textbf{First}, we identify the \emph{latent evidence-credit gap} and show that vanilla LVR does not reliably preserve answer-relevant visual evidence in its generated trajectory. \textbf{Second}, we introduce ReaLVR, which uses answer comparison to determine where stronger visual supervision is needed and visual evidence comparison to determine what the supervised tokens should preserve. This additional supervision requires no change to the model architecture or inference procedure. \textbf{Third}, we evaluate six backbones up to 235B parameters. To the best of our knowledge, we provide the first demonstration that continuous latent visual reasoning can be trained at frontier scale. The resulting latent states remain input-dependent and carry answer-relevant information rather than collapsing to a fixed trajectory. Through this work, we call for more attention toward demystifying the internal dynamics of continuous latent reasoning beyond benchmark accuracy, laying a grounded foundation for robust and faithful multimodal systems.

\section{Related Work}
\label{sec:related}

\paragraph{Visual Reasoning in Latent Space.}
Latent visual reasoning (LVR) performs intermediate computation through
continuous latent tokens rather than explicit textual
rationales~\citep{li2025latentvisualreasoning,wang2025monet,yang2025mirage,
dong2025ilvr}. Recent work has explored a range of latent trajectory designs.
Some methods switch or interleave textual and visual reasoning, adapt the
number of latent states, or combine text and image representations within a
shared latent workspace~\citep{tong2025skila,chen2026eva,tong2026swimbird,
chen2026reasoningdark,jiang2026univlr}. Others introduce structured or
coarse-to-fine trajectories~\citep{viveiros2026lantern,wang2026laser}, or
support long, parallel, decomposed, progressive, and multi-hypothesis
reasoning~\citep{wang2026scolar,lu2026deeplatent,zhu2026decompose,
li2026prolavit,huang2026dlwm,tang2026ldpvr}. These methods expand the form and
flexibility of latent computation. However, they leave open how to ensure that
a free-running latent trajectory preserves the visual evidence required for
its answer.

\paragraph{Visually Grounded Multimodal Reasoning.}
A broad line of work grounds multimodal reasoning in observable image evidence.
VisCoT selects relevant regions~\citep{shao2024viscot}; PixelReasoner and
DeepEyes revisit images through pixel-space operations or visual
tools~\citep{su2025pixelreasoner,zheng2025deepeyes}; and Argus and grounded
chain-of-thought methods make regions or coordinates explicit during
reasoning~\citep{man2025argus,wu2026groundedcot,xia2025groundedcot}.
Other methods learn multi-turn grounding from final-answer rewards or guide
policy updates with verifiable perception
questions~\citep{huang2026mgpo,zhang2026pearl}. For continuous latent
reasoning, methods use semantic or attention-trajectory
targets~\citep{xu2026semantic,wu2026lavit}, align states with visual features,
regions, relations, or contrastive objectives~\citep{miao2026gap,
cui2026ris,wang2026regular,ding2026colvr}, or develop latent-specific policy
objectives~\citep{cheng2026hylar,zhu2026silence}. RoT instead uses rendered
textual CoT rather than targets from the input image~\citep{wang2026renderthought}.
Diagnostic studies go beyond accuracy and representation similarity to probe
what latent states encode, how they respond to image evidence, and whether
final answers depend on them~\citep{li2026capimagine,
viveiros2026holdingback,zhang2026visuallatents,
zhang2026cosinemisleads,guo2026beyondvisualmemory,yang2026glaqgroundinglatentqueries,park2026reasonlatentmakinglatent,kang2026lutlatentutilitytraining}. Monet directly optimizes
sampled latent trajectories~\citep{wang2025monet}, CoLVR contrasts latent
trajectories~\citep{ding2026colvr}, and RIS supervises region
evidence~\citep{cui2026ris}. ReaLVR uses correct-versus-wrong answer
readout to weight positions within one differentiable current-model
trajectory, then applies relevant-versus-mismatched visual supervision
at those positions while retaining the LVR inference procedure.

\section{Preliminaries}
\label{sec:preliminaries}

\subsection{Latent visual reasoning}
\label{sec:prelim_lvr}

Each example contains an image--question input $x$, a ground-truth answer
$y^\star$, and optionally an evidence annotation $a$, such as a region of
interest (ROI); $a=\emptyset$ means that no region annotation is available.
Let $\pi_\theta$ denote the autoregressive policy, and let
$\mathcal T_\theta(c)$ denote the decoder hidden state produced from a causal
prefix $c$. The vision stack represents the image in $x$ with $N$ visual tokens
$\mathbf V=\{v_n\}_{n=1}^{N}$, where $v_n\in\mathbb R^d$ and $d$ is the shared
visual-token and decoder hidden-state dimension.

Latent Visual Reasoning (LVR) inserts continuous decoder states between the
input and the textual answer~\citep{li2025latentvisualreasoning}. At each
latent position, the decoder passes its hidden state directly to the next
decoding step instead of mapping it to a vocabulary token. With $K$ latent
tokens, the generation order is
\[
    x
    \rightarrow
    \texttt{<|lvr\_start|>},
    z_1,\ldots,z_K,
    \texttt{<|lvr\_end|>},
    \text{answer},
\]
where $z_t\in\mathbb R^d$. The block $z_{1:K}$ is the \emph{latent span}.
Let $o$ denote the discrete control markers and answer text. A free-running
rollout is $\tau=(z_{1:K},o)\sim\pi_\theta(\cdot\mid x)$.
If $c_t(\tau)$ is the causal prefix before latent position $t$, then
$z_t=\mathcal T_\theta(c_t(\tau))$. Thus, a generated latent token can depend
on the image, question, and earlier latent tokens, but never on future answer
tokens.

\subsection{Two-stage LVR training}
\label{sec:prelim_vanilla_objective}

LVR first initializes its latent states with visual supervision and then
post-trains the model using outcome feedback~\citep{li2025latentvisualreasoning}.

\paragraph{Stage 1: target-conditioned visual supervision.}
An ROI annotation is mapped to an ordered sequence of $T_v$ visual targets
$v_{1:T_v}^\star$. Under teacher forcing (TF), these target visual embeddings
are supplied along the latent span instead of autoregressively feeding back
the model's own generated latent states. The decoder hidden states
$z_{1:T_v}^{\mathrm{TF}}$ are trained to reconstruct this target sequence. We
therefore call them \emph{target-conditioned}, to distinguish them from the
free-running states used in Stage~2. Their visual reconstruction loss is
\begin{equation}
    \mathcal L_{\mathrm{rec}}(\theta)
    =\frac{1}{T_v}\sum_{t=1}^{T_v}
      \left\|z_t^{\mathrm{TF}}-v_t^\star\right\|_2^2.
    \label{eq:lvr_rec}
\end{equation}
Together with the standard next-token prediction loss, this stage produces
parameters $\theta_0$, which are held fixed as the reference policy during
Stage~2. The target length $T_v$ can differ from the free-running length $K$.

\paragraph{Stage 2: free-running outcome optimization.}
The frozen behavior policy $\pi_{\theta_{\mathrm{old}}}$ samples
$\{\tau_i=(z_{i,1:K}^{\mathrm{roll}},o_i)\}_{i=1}^{G}$.
Here $G$ is the group size, $z_{i,1:K}^{\mathrm{roll}}$ are the saved latent
states, and $\widehat y_i$ is the canonical answer parsed from $o_i$, with
$\widehat y_i=\bot$ for a parse failure. We write
$\operatorname{Correct}(\widehat y_i,y^\star)\in\{0,1\}$ for answer
correctness.

The standard reward combines answer correctness and output format. Group
Relative Policy Optimization (GRPO) converts the $G$ rewards into fixed
group-relative advantages
$\widehat{\mathbf A}=(\widehat A_1,\ldots,\widehat A_G)$.
Let $\mathcal J_{\mathrm{clip}}$ denote the clipped text-token GRPO objective,
$\widehat{\mathcal L}_{\mathrm{KL}}^{\mathrm{text}}$ the sampled text-token
Kullback--Leibler (KL) penalty to $\pi_{\theta_0}$, and $\beta\geq0$ its
weight. Vanilla Stage~2 minimizes
\begin{equation}
    \mathcal L_{\mathrm{S2}}^{\mathrm{LVR}}(\theta)
    =-\mathcal J_{\mathrm{clip}}
      (\theta;\theta_{\mathrm{old}},\widehat{\mathbf A})
     +\beta\widehat{\mathcal L}_{\mathrm{KL}}^{\mathrm{text}}
      (\theta;\theta_0).
    \label{eq:lvr_s2}
\end{equation}
Both terms score generated text positions. During policy replay, sampled
latent vectors are treated as fixed context, so the policy loss does not
backpropagate through the process that generated those vectors. Appendix~\ref{app:latent-evidence-credit-gap} explains why this setup leaves
free-running latent states without direct visual-evidence supervision and
distinguishes readout, grounding, and intervention utility.

\section{ReaLVR: Outcome-Contrastive Evidence Credit}
\label{sec:method}

ReaLVR couples answer-contrastive position weighting with
relevant-versus-mismatched visual supervision on one differentiable
trajectory generated by the current model. Correct-versus-wrong answer
readout assigns stronger supervision to selected latent positions; visual
contrast defines the evidence those positions should preserve. The visual
loss backpropagates through latent generation, updating the process used
at inference without changing the architecture or inference procedure.

\begin{figure}[!htbp]
  \centering
    \includegraphics[width=\linewidth]{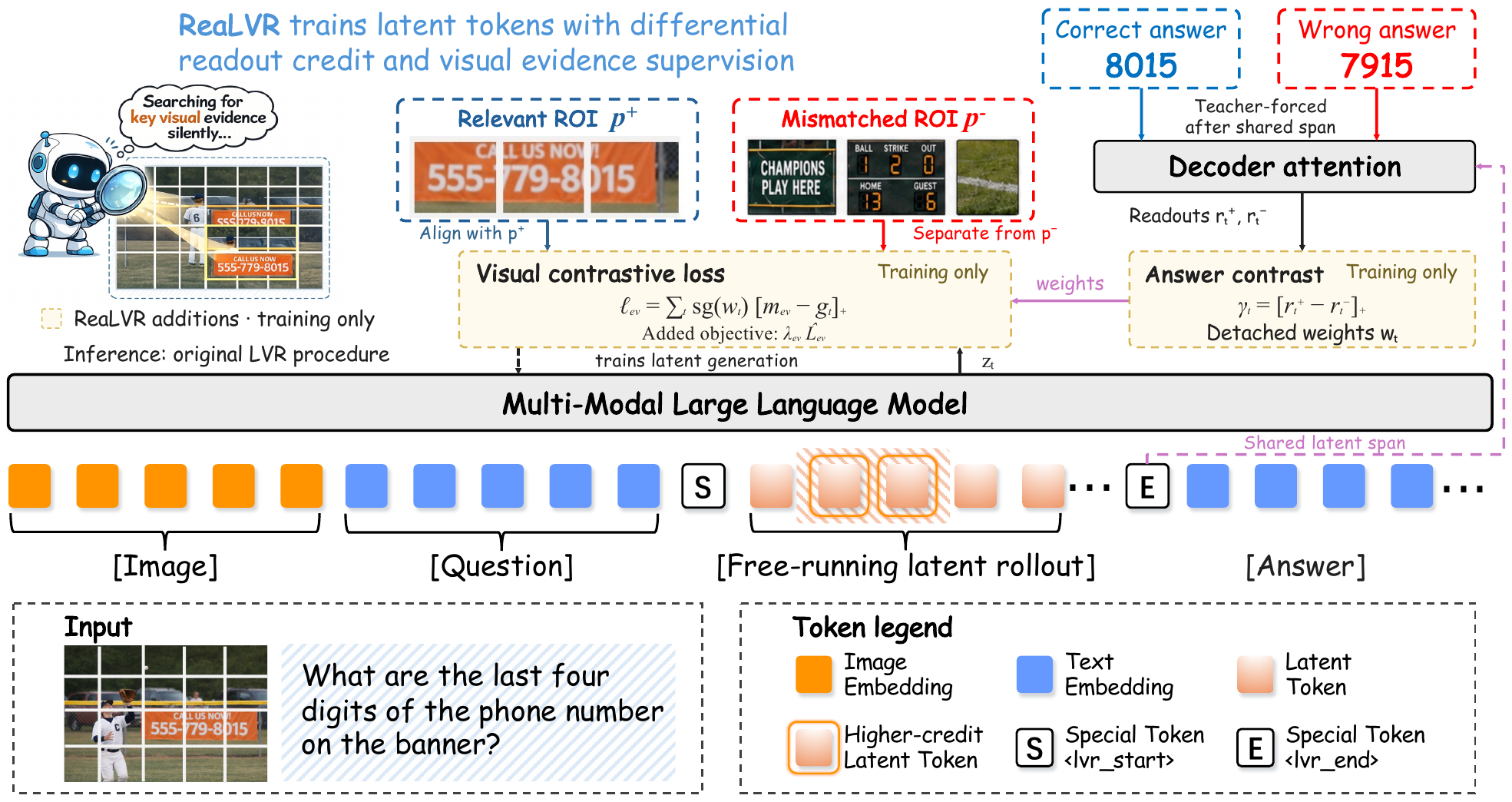}
  \caption{\small\textbf{ReaLVR learns visual grounding on its own latent trajectory.}
  The model first generates continuous latent states from the image and
  question. Relevant and mismatched visual prototypes specify what these
  states should preserve. Correct and wrong answers are separately
  teacher-forced after the shared latent span; their attention contrast
  determines the supervision weights. The weights are detached, and the
  visual loss trains the latent-generation process. Both supervision
  branches are used only during training; inference follows the original
  LVR procedure.}
  \label{fig:realvr-training-overview}
\end{figure}

\subsection{Supervise the model's own latent trajectory}
\label{sec:method_rollout}

Stage~1 teaches latent states under supplied visual prefixes, while inference
requires the model to generate those prefixes itself. This distinction
motivates the central principle of on-policy distillation: provide supervision
on trajectories produced by the learner~\citep{agarwal2024onpolicy}.
ReaLVR applies this principle to visual grounding, using image evidence to
supervise the current model's own latent computation. For each input $x$, the $G$ behavior-policy completions from
Section~\ref{sec:prelim_vanilla_objective} supply the GRPO outcomes and
candidate wrong answers. GRPO replays these completions with their saved
latent inputs fixed. Alongside this policy update, ReaLVR regenerates one
current-model trajectory by recursively feeding back its own hidden states:
\[
    z_t^{\theta}
    =\mathcal T_\theta\!\left(x,\texttt{<|lvr\_start|>},
      z_{1:t-1}^{\theta}\right),
    \qquad t=1,\ldots,K.
\]
We retain gradients through this recurrence, allowing the evidence loss to
update the process that produces the latent states.
The entire latent span is generated before any answer token is supplied.
Correct and wrong answers are then teacher-forced in separate branches after
this shared span to compute the supervision weights. Thus, one differentiable
trajectory supports both visual alignment and answer-conditioned routing.
Figure~\ref{fig:realvr-training-overview} illustrates these two sources of
supervision.

\subsection{Specify what to preserve with visual contrast}
\label{sec:method_evidence_target}

We anchor the generated states to visual evidence from the input image.
The annotation $a$ defines an evidence mask
$a^+=(a_1^+,\ldots,a_N^+)\in[0,1]^N$ over the visual tokens $\mathbf V$.
The positive prototype is the masked mean,
$p^+=\operatorname{Pool}(\mathbf V;a^+)
=\frac{\sum_n a_n^+v_n}{\sum_n a_n^++\varepsilon}$, with
$\varepsilon>0$. If the ROI is unavailable or its mask is empty, we set
$a_n^+=1$ and obtain a whole-image target. All latent positions share this
visual target; their supervision strengths will be determined by answer
contrast. To make the target discriminative, we form a set
$\mathcal N=\{p_s^-\}_{s=1}^{N_-}$ of nonzero visual prototypes pooled from
mismatched examples using the same construction, with $N_-\geq1$.
The margin at latent position $t$ compares the relevant prototype with the
most similar negative:
\[
    g_t
    =\operatorname{sim}(z_t^{\theta},p^+)
     -\max_{p^-\in\mathcal N}
      \operatorname{sim}(z_t^{\theta},p^-),
\]
where $\operatorname{sim}$ is cosine similarity. Increasing this margin trains
the latent state to distinguish the supporting visual content from competing
image features. The vision encoder and connector are frozen, keeping the
prototypes fixed as the language model learns to preserve their content.

\subsection{Locate supervision with answer contrast}
\label{sec:method_readout_gate}

The model's own wrong answers provide a reference for identifying which
latent positions are preferentially read under the correct answer. Subtracting
this reference discounts attention shared across competing outcomes and
concentrates supervision on positions with a stronger correct-answer readout.
From the $G$ behavior-policy completions, we construct $\mathcal Y_x^-$ by
retaining distinct, parseable wrong answers with nonempty answer content.
These answer candidates guide position selection, while the visual negatives
in $\mathcal N$ define the content to distinguish.

For a canonical answer $y$, let
$\operatorname{Fmt}(y)=(\widetilde y_1,\ldots,\widetilde y_{M(y)})$ be its
formatted sequence, and let $\mathcal J(y)$ index the content tokens after
excluding control and format markers. Each candidate is teacher-forced after
$(x,\texttt{<|lvr\_start|>},z_{1:K}^{\theta},\texttt{<|lvr\_end|>})$.
We extract attention at the input position of each content token
$\widetilde y_j$: its query conditions on $\widetilde y_{1:j}$, including
$\widetilde y_j$ itself. For a single-token multiple-choice answer, this is
the query after \texttt{A} or \texttt{B} has been supplied as input.

Let $A_{j,t}^{(\ell,h)}(y)$ denote the resulting post-softmax attention to
latent position $t$. Averaging over decoder layers
$\ell\in\mathcal L_{\mathrm{dec}}$, heads $h\in\mathcal H$, and content
positions $j\in\mathcal J(y)$ gives the readout
$r_t(y)=\operatorname{mean}_{\ell,h,j}A_{j,t}^{(\ell,h)}(y)$.
We retain the raw attention mass without renormalizing it within the latent
span. All candidate branches use the current model and the same regenerated
latent states; their role is to compute routing weights. Write $r_t^+=r_t(y^\star)$ for the correct-answer readout and
$r_t^-=|\mathcal Y_x^-|^{-1}\sum_{y^-\in\mathcal Y_x^-}r_t(y^-)$ for the
mean wrong-answer readout. The selective credit is their positive difference,
$\gamma_t=[r_t^+-r_t^-]_+$, where $[u]_+=\max(u,0)$.
When no valid wrong answer is available, we set $r_t^-=r_t^+$, giving
$\gamma_t=0$. We keep the magnitude of this contrast so that it expresses both
the preferred positions and the strength of the routing signal.

\subsection{Train with readout-weighted visual evidence}
\label{sec:method_objective}

We combine selective credit with a uniform baseline,
$w_t=\eta/K+(1-\eta)\gamma_t$, where $\eta\in[0,1]$ controls the baseline
strength. The total weight is
$\sum_t w_t=\eta+(1-\eta)\sum_t\gamma_t$:
stronger answer contrast increases the evidence supervision assigned to the
example. With no selective signal, the weights reduce to $w_t=\eta/K$,
retaining uniform supervision whenever $\eta>0$. Hyperparameters are listed in Appendix~\ref{app:hparams}. We detach $w_t$ when optimizing the visual margin. This makes the weights
allocate supervision while the gradient improves the evidence representation,
preventing a shortcut through reducing the weight itself.
Let $\operatorname{sg}$ denote stop-gradient and let
$m_{\mathrm{ev}}\in(0,2]$ be the target margin. The per-example evidence loss
and the Stage~2 objective are
\begin{align}
    \ell_{\mathrm{ev}}
    &=\sum_{t=1}^{K}\operatorname{sg}(w_t)
      [m_{\mathrm{ev}}-g_t]_+,
    \label{eq:realvr_batch_evidence}\\
    \mathcal L_{\mathrm{S2}}^{\mathrm{ReaLVR}}(\theta)
    &=\mathcal L_{\mathrm{S2}}^{\mathrm{LVR}}(\theta)
      +\lambda_{\mathrm{ev}}\widehat{\mathcal L}_{\mathrm{ev}}(\theta),
    \label{eq:realvr_difference}
\end{align}
where $\widehat{\mathcal L}_{\mathrm{ev}}
=B^{-1}\sum_{b=1}^{B}\ell_{\mathrm{ev}}^{(b)}$ averages over a minibatch of
$B$ examples and $\lambda_{\mathrm{ev}}\geq0$ controls the added objective.
Gradients of Equation~\ref{eq:realvr_batch_evidence} flow through the visual
margins and recurrent latent generation, with answer strings, prototypes, and
weights fixed. Equation~\ref{eq:realvr_difference} preserves GRPO rewards and
advantages while training latent states on discriminative visual evidence.
Appendices~\ref{app:uniform-bootstrap} and~\ref{app:detached-credit} detail
the weight allocation and gradient decomposition. At inference, ReaLVR generates $K$ latent tokens and decodes the answer with
the original LVR architecture and procedure; both supervision branches are
used only during training.

\section{Experiments}
\label{sec:experiments}

\subsection{Experimental Setup}
\label{sec:exp-setup}
We evaluate ReaLVR on five benchmarks of visual discrimination, spatial
reasoning, and high-resolution perception: \texttt{MMVP}~\citep{tong2024eyes},
\texttt{BLINK}~\citep{fu2024blink}, \texttt{HRBench-4K/8K}~\citep{wang2025hrbench},
and \texttt{MME-RealWorld}~\citep{zhang2025mmerealworld}.
Six backbones span Qwen2.5-VL and Qwen3-VL~\citep{bai2025qwen25vl,bai2025qwen3vl},
InternVL3~\citep{zhu2025internvl3}, and Gemma~3~\citep{gemmateam2025gemma3}.
We compare with Pixel Reasoner~\citep{su2025pixelreasoner},
Vision-R1~\citep{huang2026visionr1}, LVR~\citep{li2025latentvisualreasoning},
ILVR~\citep{dong2025ilvr}, and Monet~\citep{wang2025monet}. We report task accuracy and the unweighted five-benchmark
mean when available. Training used 800 AMD MI250X GPUs ($128$\,GB each), please see Appendix~\ref{app:hparams} for detailed settings.
Appendix~\ref{app:experimental-settings} covers evaluation protocols and
aggregation; Appendix~\ref{app:diagnostics} defines the analysis estimators;
Appendix~\ref{app:benchmark-case-studies} gives benchmark cases.

\subsection{Main Results}
\label{sec:main-results}

\providecommand{\miss}{\textcolor{gray}{--}}
\providecommand{\tbd}{\textcolor{gray}{tbd}}
\providecommand{\rep}{\textcolor{gray}{rep.}}
\providecommand{\directrow}[1]{\textcolor{black!58}{#1}}
\providecommand{\basemethod}[1]{\textcolor{black!58}{#1}}
\providecommand{\stdmethod}[1]{\cellcolor{brightpurple!18}#1}
\providecommand{\latentmethod}[1]{\cellcolor{brightgreen!18}#1}
\providecommand{\baselegend}[1]{\textcolor{black!58}{#1}}
\providecommand{\stdlegend}[1]{\begingroup\setlength{\fboxsep}{1.2pt}\colorbox{brightpurple!18}{\strut #1}\endgroup}
\providecommand{\latentlegend}[1]{\begingroup\setlength{\fboxsep}{1.2pt}\colorbox{brightgreen!18}{\strut #1}\endgroup}
\providecommand{\modelsep}{\specialrule{0.45pt}{2.4pt}{2.4pt}}
\providecommand{\methodlabel}[2]{#1\,\textcolor{lightgray}{\scriptsize{[#2]}}}

\begin{table*}[!htbp]
\caption{\textbf{Comparison with SOTA methods.} Five-benchmark accuracy on
\texttt{Qwen2.5-VL-7B}. We report mean $\pm$ standard deviation over three
random seeds.
Purple method cells mark non-latent baselines, green method cells mark
latent-reasoning baselines.}
\vspace{-0.5pt}
\centering
\begingroup
\renewcommand{\bfdefault}{bx}
\normalfont\fontfamily{lmr}\selectfont
\definecolor{mainoursblue}{RGB}{232,241,249}
\definecolor{mainbandgray}{RGB}{242,244,246}
\definecolor{mainaccent}{RGB}{28,72,106}
\definecolor{mainstdblue}{RGB}{76,130,176}
\renewcommand{\directrow}[1]{#1}
\renewcommand{\methodlabel}[2]{#1\,\textcolor{black!45}{\scriptsize[#2]}}
\newcommand{\scorestd}[2]{$#1_{{\color{mainstdblue}\pm #2}}$}
\renewcommand{\modelsep}{\cmidrule(l){2-9}\addlinespace[1pt]}
\newcommand{\mainsectionrow}[1]{%
  \rowcolor{mainbandgray}\multicolumn{9}{c}{%
    \rule{0pt}{2.5ex}\textit{#1}\rule[-0.8ex]{0pt}{0pt}}\\[2pt]}
\renewcommand{\arraystretch}{1.02}
\setlength{\tabcolsep}{3.5pt}
\setlength{\aboverulesep}{0.28ex}
\setlength{\belowrulesep}{0.28ex}
\setlength{\extrarowheight}{0pt}
\adjustbox{max width=\textwidth,max totalheight=0.9\textheight,center}{%
\begin{tabular}{lcl ccccc c}
\toprule
\multirow{2}{*}{Model} &
\multicolumn{2}{c}{Training recipe} &
\multicolumn{2}{c}{Visual reasoning} &
\multicolumn{3}{c}{High-res. / real-world} &
\multirow{2}{*}{\shortstack{\textbf{Average}$\uparrow$\\[-1pt]{\scriptsize (5 tasks)}}} \\
\cmidrule(r){2-3} \cmidrule(lr){4-5} \cmidrule(lr){6-8}
& Stage & Method
& \texttt{MMVP}$\uparrow$
& \texttt{BLINK}$\uparrow$
& \texttt{HR-4K}$\uparrow$
& \texttt{HR-8K}$\uparrow$
& \texttt{MME-RW}$\uparrow$ \\
\midrule
\multirow{9}{*}{\directrow{\texttt{Qwen2.5-VL-7B}}} & RL & \stdmethod{\methodlabel{Pixel Reasoner}{NeurIPS'25}} &
\scorestd{66.8}{0.4} & \scorestd{53.3}{0.3} & \scorestd{69.8}{0.3} & \scorestd{64.1}{0.3} & \scorestd{49.7}{0.2} & \scorestd{60.7}{0.2} \\
& RL & \stdmethod{\methodlabel{Vision-R1}{ICLR'26}} &
\scorestd{51.5}{0.5} & \scorestd{52.7}{0.4} & \scorestd{62.9}{0.4} & \scorestd{58.6}{0.4} & \scorestd{44.2}{0.3} & \scorestd{54.0}{0.3} \\
& SFT & \latentmethod{\methodlabel{LVR-SFT}{ICLR'26}} &
\scorestd{63.6}{0.3} & \scorestd{53.2}{0.2} & \scorestd{69.0}{0.3} & \scorestd{63.3}{0.2} & \scorestd{49.5}{0.2} & \scorestd{59.7}{0.2} \\
& RL & \latentmethod{\methodlabel{LVR-RL}{ICLR'26}} &
\scorestd{64.2}{0.4} & \scorestd{53.6}{0.3} & \scorestd{69.6}{0.3} & \scorestd{64.4}{0.3} & \scorestd{50.1}{0.3} & \scorestd{60.4}{0.2} \\
& S1 & \latentmethod{\methodlabel{ILVR-Stage1}{ACL'26}} &
\scorestd{68.1}{0.3} & \scorestd{55.4}{0.3} & \scorestd{70.2}{0.3} & \scorestd{66.0}{0.3} & \scorestd{49.1}{0.2} & \scorestd{61.8}{0.2} \\
& S2 & \latentmethod{\methodlabel{ILVR-Stage2}{ACL'26}} &
\scorestd{69.4}{0.4} & \scorestd{\mathbf{56.8}}{0.3} & \scorestd{71.0}{0.2} & \scorestd{\mathbf{66.9}}{0.3} & \scorestd{50.3}{0.3} & \scorestd{62.9}{0.2} \\
& SFT & \latentmethod{\methodlabel{Monet-SFT}{CVPR'26}} &
\scorestd{68.4}{0.3} & \scorestd{51.2}{0.4} & \scorestd{68.9}{0.3} & \scorestd{64.8}{0.3} & \scorestd{50.7}{0.2} & \scorestd{60.8}{0.2} \\
& RL & \latentmethod{\methodlabel{Monet-RL}{CVPR'26}} &
\scorestd{69.9}{0.4} & \scorestd{52.4}{0.4} & \scorestd{71.3}{0.3} & \scorestd{66.0}{0.3} & \scorestd{51.5}{0.2} & \scorestd{62.2}{0.3} \\
\rowcolor{mainoursblue}\cellcolor{white}
& \cellcolor{white}RL & \textcolor{mainaccent}{\textbf{\realvr{}}} &
\scorestd{\mathbf{72.0}}{0.4} & \scorestd{55.8}{0.3} & \scorestd{\mathbf{71.8}}{0.3} & \scorestd{66.6}{0.3} & \scorestd{\mathbf{52.2}}{0.2} &
\scorestd{\mathbf{63.7}}{0.2} \\
\bottomrule
\end{tabular}
 }
\endgroup
\label{tab:main-backbone-results}
\vspace{-6pt}
\end{table*}

\providecommand{\miss}{\textcolor{gray}{--}}
\providecommand{\tbd}{\textcolor{gray}{tbd}}
\providecommand{\rep}{\textcolor{gray}{rep.}}
\providecommand{\directrow}[1]{\textcolor{black!58}{#1}}
\providecommand{\basemethod}[1]{\textcolor{black!58}{#1}}
\providecommand{\stdmethod}[1]{\cellcolor{brightpurple!18}#1}
\providecommand{\latentmethod}[1]{\cellcolor{brightgreen!18}#1}
\providecommand{\baselegend}[1]{\textcolor{black!58}{#1}}
\providecommand{\stdlegend}[1]{\begingroup\setlength{\fboxsep}{1.2pt}\colorbox{brightpurple!18}{\strut #1}\endgroup}
\providecommand{\latentlegend}[1]{\begingroup\setlength{\fboxsep}{1.2pt}\colorbox{brightgreen!18}{\strut #1}\endgroup}
\providecommand{\modelsep}{\specialrule{0.45pt}{2.4pt}{2.4pt}}
\providecommand{\methodlabel}[2]{#1\,\textcolor{lightgray}{\scriptsize{[#2]}}}

\begin{table*}[!htbp]
\caption{\textbf{Applicability across model sizes and families.} The 235B evaluation covers MMVP, BLINK, and MME-RealWorld; a five-task average is reported only when all five scores are available.}
\vspace{-0.5pt}
\centering
\begingroup
\renewcommand{\bfdefault}{bx}
\normalfont\fontfamily{lmr}\selectfont
\definecolor{mainoursblue}{RGB}{232,241,249}
\definecolor{mainbandgray}{RGB}{242,244,246}
\definecolor{mainaccent}{RGB}{28,72,106}
\renewcommand{\directrow}[1]{#1}
\renewcommand{\methodlabel}[2]{#1\,\textcolor{black!45}{\scriptsize[#2]}}
\renewcommand{\modelsep}{\cmidrule(l){2-9}\addlinespace[1pt]}
\newcommand{\mainsectionrow}[1]{%
  \rowcolor{mainbandgray}\multicolumn{9}{c}{%
    \rule{0pt}{2.5ex}\textit{#1}\rule[-0.8ex]{0pt}{0pt}}\\[2pt]}
\renewcommand{\arraystretch}{1.02}
\setlength{\tabcolsep}{3.5pt}
\setlength{\aboverulesep}{0.28ex}
\setlength{\belowrulesep}{0.28ex}
\setlength{\extrarowheight}{0pt}
\adjustbox{max width=\textwidth,max totalheight=0.9\textheight,center}{%
\begin{tabular}{lcl ccccc c}
\toprule
\multirow{2}{*}{Model} &
\multicolumn{2}{c}{Training recipe} &
\multicolumn{2}{c}{Visual reasoning} &
\multicolumn{3}{c}{High-res. / real-world} &
\multirow{2}{*}{\shortstack{\textbf{Average}$\uparrow$\\[-1pt]{\scriptsize (5 tasks)}}} \\
\cmidrule(r){2-3} \cmidrule(lr){4-5} \cmidrule(lr){6-8}
& Stage & Method
& \texttt{MMVP}$\uparrow$
& \texttt{BLINK}$\uparrow$
& \texttt{HR-4K}$\uparrow$
& \texttt{HR-8K}$\uparrow$
& \texttt{MME-RW}$\uparrow$ \\
\midrule
\mainsectionrow{Across Qwen3-VL model sizes}
\multirow{4}{*}{\directrow{\texttt{Qwen3-VL-8B}}} & SFT & \latentmethod{\methodlabel{LVR-SFT}{ICLR'26}} &
65.5 & 52.9 & 68.4 & 61.0 & 47.7 & 59.1 \\
& RL & \latentmethod{\methodlabel{LVR-RL}{ICLR'26}} &
66.9 & 54.2 & 69.9 & 62.2 & 49.7 & 60.6 \\
& RL & \latentmethod{\methodlabel{Monet}{CVPR'26}} &
67.4 & 55.1 & 68.5 & 61.9 & 48.4 & 60.3 \\
\rowcolor{mainoursblue}\cellcolor{white}
& \cellcolor{white}RL & \textcolor{mainaccent}{\textbf{\realvr{}}} &
68.7 & 54.9 & 70.4 & 62.9 & 49.3 & \textbf{61.2} \\
\modelsep
\multirow{3}{*}{\directrow{\texttt{Qwen3-VL-30B}}} & SFT & \latentmethod{\methodlabel{LVR-SFT}{ICLR'26}} &
77.0 & 47.5 & 73.8 & 66.8 & 53.4 & 63.7 \\
& RL & \latentmethod{\methodlabel{LVR-RL}{ICLR'26}} &
77.6 & 47.8 & 73.9 & 67.2 & 53.9 & 64.1 \\
\rowcolor{mainoursblue}\cellcolor{white}
& \cellcolor{white}RL & \textcolor{mainaccent}{\textbf{\realvr{}}} &
77.7 & 51.4 & 74.1 & 67.9 & 54.7 & \textbf{65.2} \\
\modelsep

& \directrow{--} & \basemethod{Direct} &
\directrow{80.0} & \directrow{73.5} & \miss & \miss & 69.8 & \miss \\
& SFT & \latentmethod{\methodlabel{LVR-SFT}{ICLR'26}} &
80.9 & 74.2 & \miss & \miss & 70.4 & \miss \\
\rowcolor{mainoursblue}
\cellcolor{white}\multirow{-3}{*}{\texttt{Qwen3-VL-235B}}
& \cellcolor{white}RL & \textcolor{mainaccent}{\textbf{\realvr{}}} &
81.9 & 75.4 & \miss & \miss & 71.0 & \miss \\
\addlinespace[3pt]
\mainsectionrow{Across model families}
& SFT & \latentmethod{\methodlabel{LVR-SFT}{ICLR'26}} &
71.0 & 46.1 & 60.0 & 53.6 & 39.0 & 53.9 \\
& RL & \latentmethod{\methodlabel{LVR-RL}{ICLR'26}} &
71.4 & 46.6 & 61.2 & 54.0 & 41.6 & 55.0 \\
\rowcolor{mainoursblue}
\cellcolor{white}\multirow{-3}{*}{\directrow{\texttt{InternVL3-8B}}}
& \cellcolor{white}RL & \textcolor{mainaccent}{\textbf{\realvr{}}} &
72.3 & 52.4 & 64.4 & 55.4 & 45.9 & \textbf{58.1} \\
\modelsep
& SFT & \latentmethod{\methodlabel{LVR-SFT}{ICLR'26}} &
58.3 & 32.9 & 36.1 & 34.6 & 28.1 & 38.0 \\
& RL & \latentmethod{\methodlabel{LVR-RL}{ICLR'26}} &
59.1 & 33.3 & 36.7 & 36.3 & 29.6 & 39.0 \\
\rowcolor{mainoursblue}
\cellcolor{white}\multirow{-3}{*}{\directrow{\texttt{Gemma-3-12B}}}
& \cellcolor{white}RL & \textcolor{mainaccent}{\textbf{\realvr{}}} &
60.0 & 40.6 & 37.4 & 38.0 & 31.8 & \textbf{41.6} \\
\bottomrule
\end{tabular}
 }
\endgroup
\label{tab:cross-backbone-results}
\vspace{-6pt}
\end{table*}

ReaLVR reaches the highest five-task average on \texttt{Qwen2.5-VL-7B},
$63.7$ (Table~\ref{tab:main-backbone-results}): $+4.0$ points over LVR-SFT,
$+3.3$ over LVR-RL, $+2.9$ over Monet-SFT, $+1.5$ over Monet-RL, and $+0.8$
over ILVR, the strongest competing latent-reasoning row by average accuracy.
Gains over LVR-SFT cover all five tasks, led by \texttt{MMVP} ($+8.4$),
\texttt{HR-8K} ($+3.3$), and \texttt{HR-4K} ($+2.8$), spanning subtle visual
discrimination and high-resolution evidence. Across model sizes (Table~\ref{tab:cross-backbone-results}), applying the
same objective and inference procedure to
\texttt{Qwen3-VL-8B}, \texttt{Qwen3-VL-30B}, and \texttt{Qwen3-VL-235B-A22B}
(235B total parameters) yields five-task averages of $61.2$ and $65.2$ at
$8$B and $30$B. At $235$B, ReaLVR scores $81.9$ on MMVP, $75.4$ on BLINK,
and $71.0$ on MME-RealWorld. HRBench was not evaluated, so no five-task mean
is reported. Across model families, ReaLVR yields five-task means of $58.1$
on \texttt{InternVL3-8B} and $41.6$ on \texttt{Gemma-3-12B} without
architecture or inference changes. On these backbones, ReaLVR exceeds LVR-RL
by $3.1$ and $2.6$ points in five-task mean accuracy, respectively.
These families use different vision encoders and language backbones; Gemma
uses a fixed $256$-token, single-tile
image representation. Together, these results show applicability across
three model families and up to 235B total parameters.
Appendix~\ref{sec:mablation} isolates the components, and
Appendix~\ref{app:latent-length-ablation}
sweeps inference budgets $K=0$--$20$; a short span captures much of the
benefit, and $K=8$ gives the highest mean accuracy.
Appendix~\ref{app:mass-matched-factorial} reports a further
supervision-mass-matched $2\times2$ test of answer-based position
allocation and positive-versus-negative visual evidence.

\subsection{Mechanism Analysis: Variation, Grounding, and Use}
\label{sec:representation-analysis}

\begin{figure}[!htbp]
  \centering
  \includegraphics[width=\linewidth]{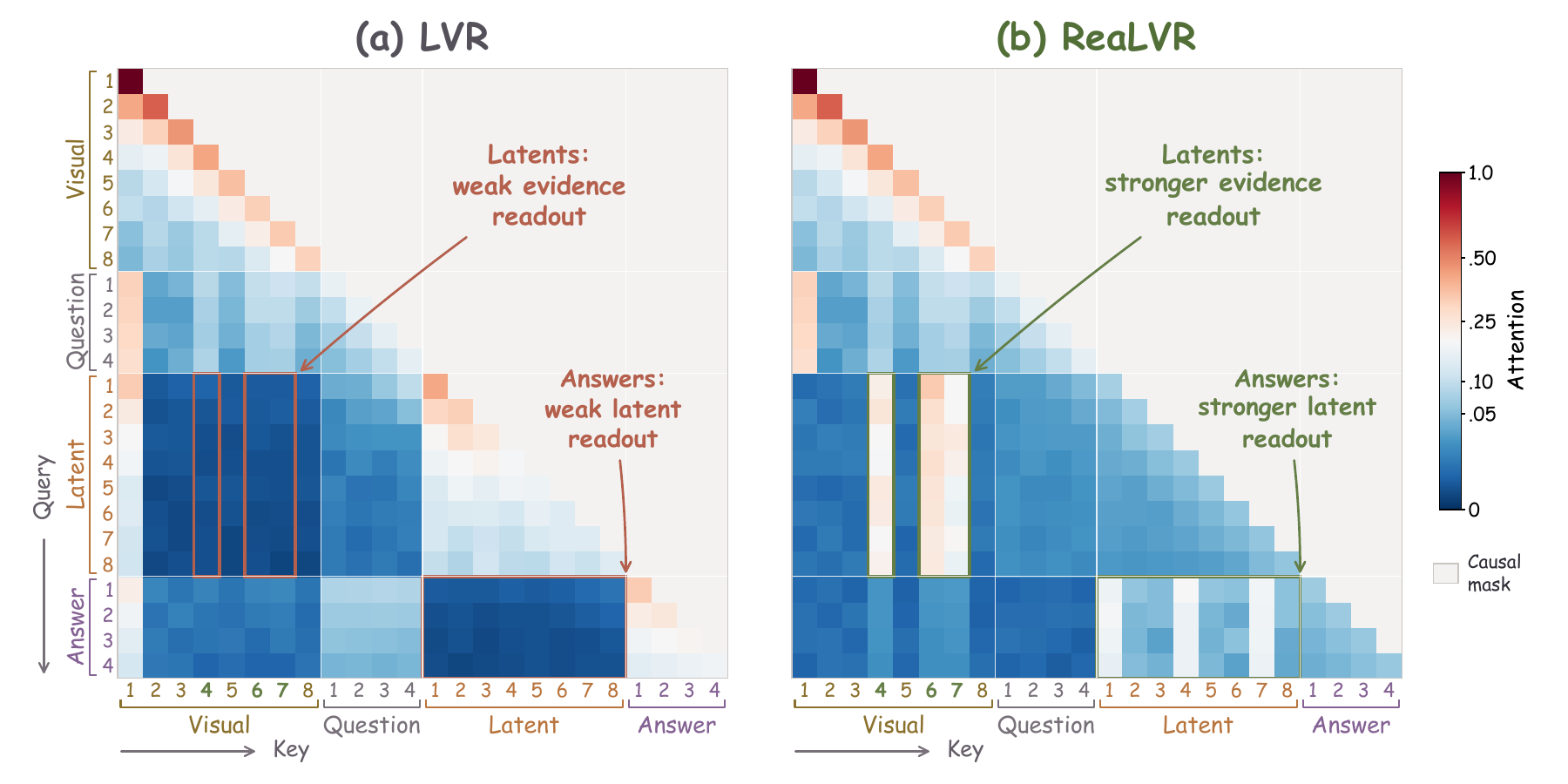}
\caption{\small\textbf{Visual evidence and latent-token readout.} Attention of
  LVR (a) and ReaLVR (b) on \texttt{Qwen2.5-VL-7B},
  averaged over all layers and heads and 200 HR-Bench-4K
  examples, with image, question, and answer tokens averaged into
  8/4/4 bins and the $K=8$ latent positions shown individually.
  The gray upper triangle is the causal mask: each query
  can attend only to its own and earlier positions~\citep{vaswani2017attention}.
  Each row sums to one over allowed keys.
  Green tick labels mark the same visual-evidence keys ($4$, $6$, and $7$),
  the image bins overlapping the annotated ROI.
  The narrow frames show latent queries reading these visual keys; the bottom
  frames show answer queries reading latent states. Panel (a) shows weak readout along both links; panel (b) shows stronger readout along both. Please see Appendix~\ref{app:attention-designs} for further attention analyses.}
  \label{fig:attention-grounding-schematic}
\end{figure}

Visual attention can link generated words to image regions~\citep{xu2015show}.
We examine visual-evidence readout into latent states and answer readout from
them (Figure~\ref{fig:attention-grounding-schematic}; Appendix~\ref{app:attention-designs}),
using variation, grounding, and fixed-context replacement to test answer dependence.Figure~\ref{fig:attention-grounding-schematic} shows weak LVR attention from
latent queries to target-overlapping visual bins and from answer queries to
latent states; ReaLVR strengthens both links. The latter link motivates the
dependence test below.

\begin{figure}[!htbp]
  \centering
  \includegraphics[width=\linewidth]{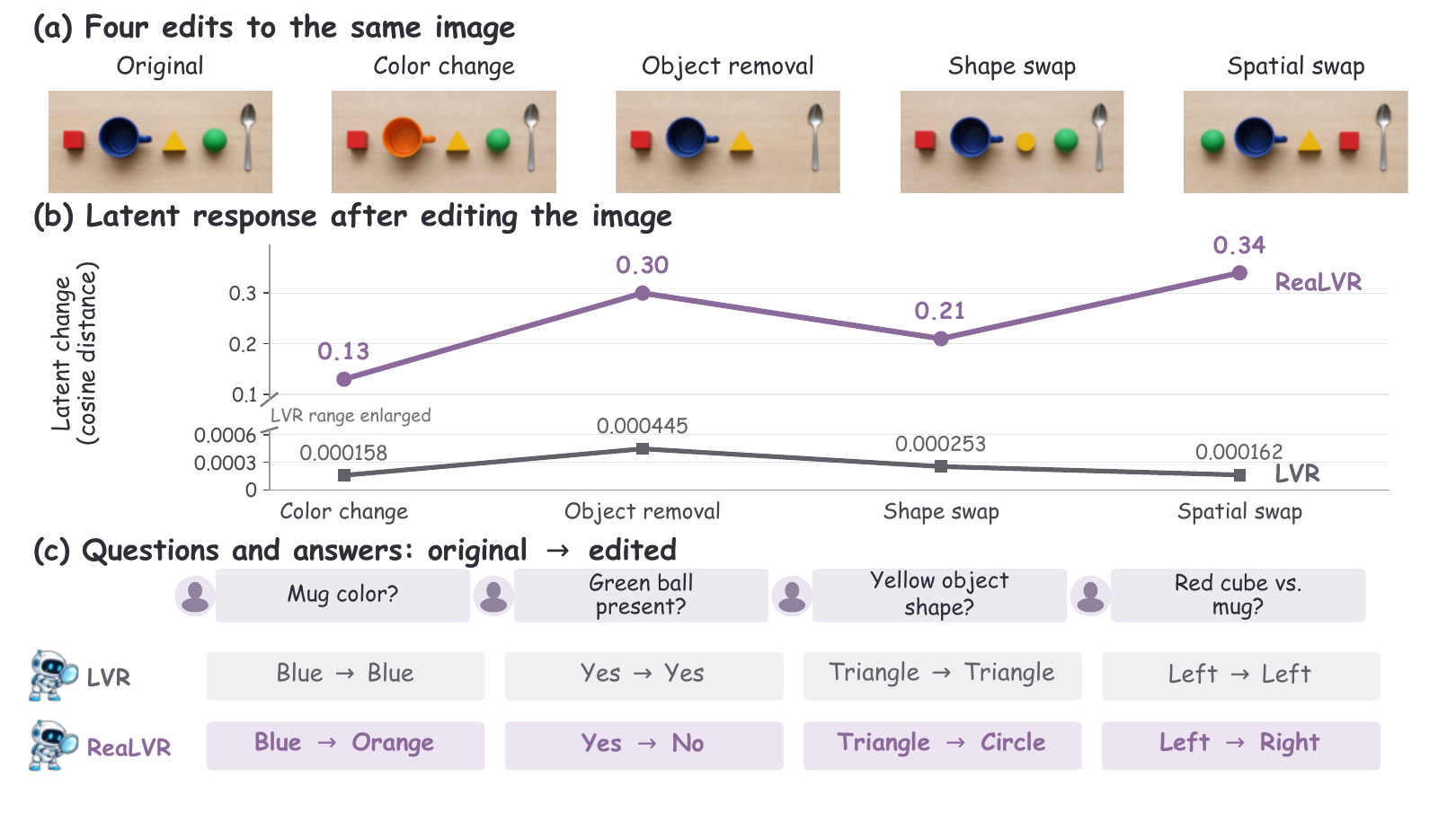}
  \caption{\small\textbf{Latent changes and answer updates after image edits.}
  (a) Four edits of the same reference image.
  (b) Cosine distance between the mean-pooled latent trajectories for the original and edited inputs; zero denotes no change. Each point is the average over 512 original--edited pairs per edit type with the question held fixed, measured for both LVR and ReaLVR on \texttt{Qwen2.5-VL-7B}. The broken vertical axis enlarges the LVR range to make its small fluctuations visible; values are unchanged.
  (c) Each user question is followed by the two model replies, read from
  the original image to the edited image. See Appendix~\ref{app:qualitative-cases} for more analysis.}
  \label{fig:lvr-counterfactual-diagnostic}
\end{figure}

\begin{figure}[!htbp]
  \centering
  \includegraphics[width=\linewidth]{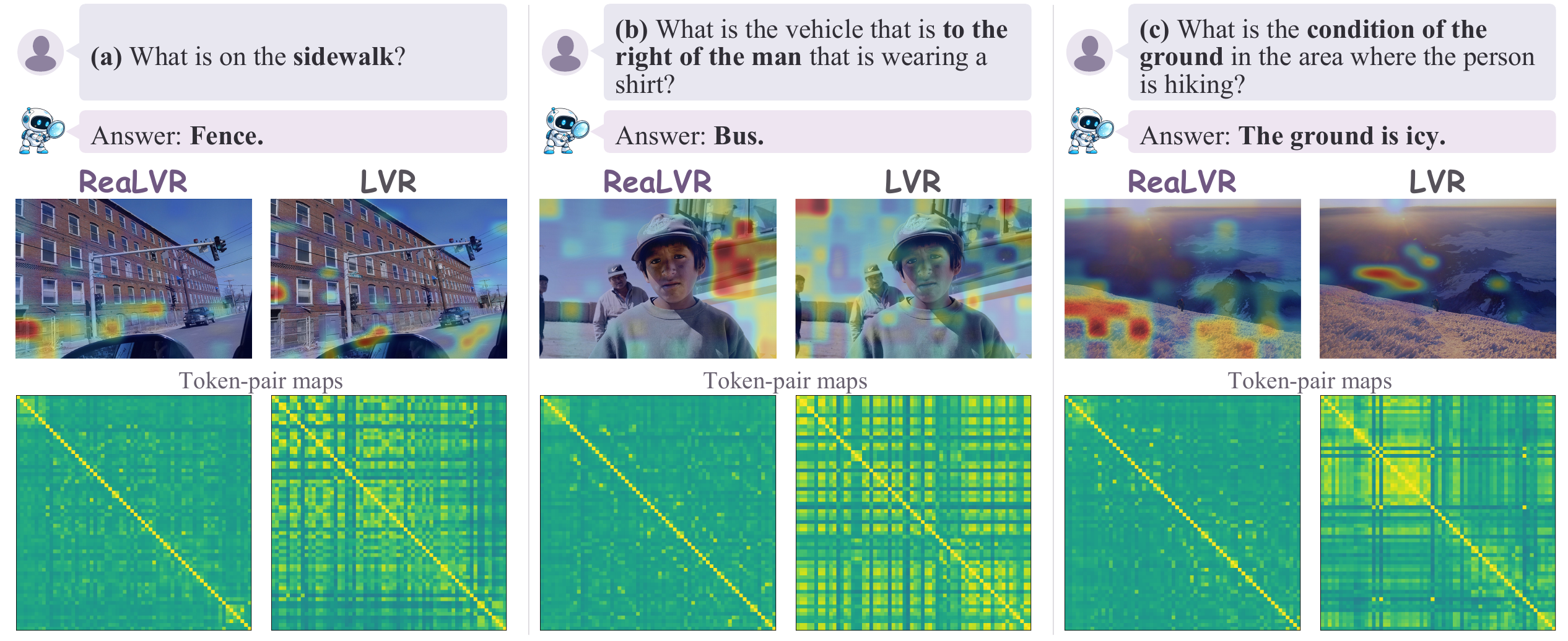}
  \caption{\small\textbf{Image and token views of the same three visual questions.}
  ReaLVR and LVR on the same fence, bus, and icy-ground questions:
  image saliency overlays (top) and token-pair maps (bottom).
  Each pair is labeled with its question and answer.}
  \label{fig:token-saliency-cases}
  \label{fig:image-evidence-heatmaps-full}
\end{figure}

\paragraph{Vanilla LVR has weak counterfactual sensitivity.}
We test whether generated latent states respond to edits that change
answer-relevant evidence. Each of four edit types contains 512
original--edited pairs with a fixed question. The correct answer changes
in $81.45\%$--$86.33\%$ of pairs, but LVR changes its prediction in only
$5.66\%$--$13.09\%$ (Table~\ref{tab:diag-counterfactual}). These edits
preserve much of the scene while altering a decisive cue, as in
counterfactual VQA evaluations~\citep{agarwal2020causalvqa,dancette2021shortcuts}. Figure~\ref{fig:lvr-counterfactual-diagnostic}(c) illustrates the
failure: LVR answers the original views correctly but retains its answers
after changes to mug color, ball presence, object shape, or left--right
relation. ReaLVR updates its answer in each case. Across the full sets,
the gap between ground-truth and LVR prediction-change rates is
$68.36$--$80.67$ percentage points. In panel (b), ReaLVR's mean-pooled
latent distance between original and edited inputs is $0.13$--$0.34$,
versus below $0.0005$ for LVR. \textit{LVR's trajectory may retain shared
scene information while responding weakly to the cue needed to revise
the answer.} Together with its weak free-running alignment to visual
targets (Table~\ref{tab:diag-teacher-forcing}), this motivates
supervising generated latent states on answer-relevant evidence.
Appendices~\ref{app:complete-counterfactual} and~\ref{app:latent-exchange}
report paired Direct/LVR/ReaLVR answer-change tests and latent
interventions to separate correct answer revision from local dependence
on the latent span.

\paragraph{Latent positions respond differently to input changes.}
Figure~\ref{fig:slot-variance-map} compares latent-position variation across
image--question examples (a) and across questions about one fixed image (b). ReaLVR's variation is more concentrated at particular positions than
Monet's or the LVR variants'. Panel (c) reports a top-token variation
gap of $0.07$ for ReaLVR, $0.02$ for Monet, and $0.01$ for each LVR
variant. Mean pooling can obscure changes concentrated in a few latent
states~\citep{ennadir2025pool}. This pattern is consistent with ReaLVR's
position-specific training signal: \textit{positions share a visual
target within an example, while correct-versus-wrong-answer readout
assigns them different supervision weights.} Figure~\ref{fig:token-saliency-cases} adds saliency views for the fence,
bus, and icy-ground questions. ReaLVR focuses more on the relevant
regions, whereas LVR's saliency is more diffuse. Because each overlay
is normalized within its example, the comparison concerns spatial
focus. Together, the variation map, saliency cases, and fixed-context
replacement examine \textit{how latent states change, where visual
evidence is read, and whether selected states affect the answer.}

\begin{figure}[!h]
  \centering
  \includegraphics[width=\linewidth]{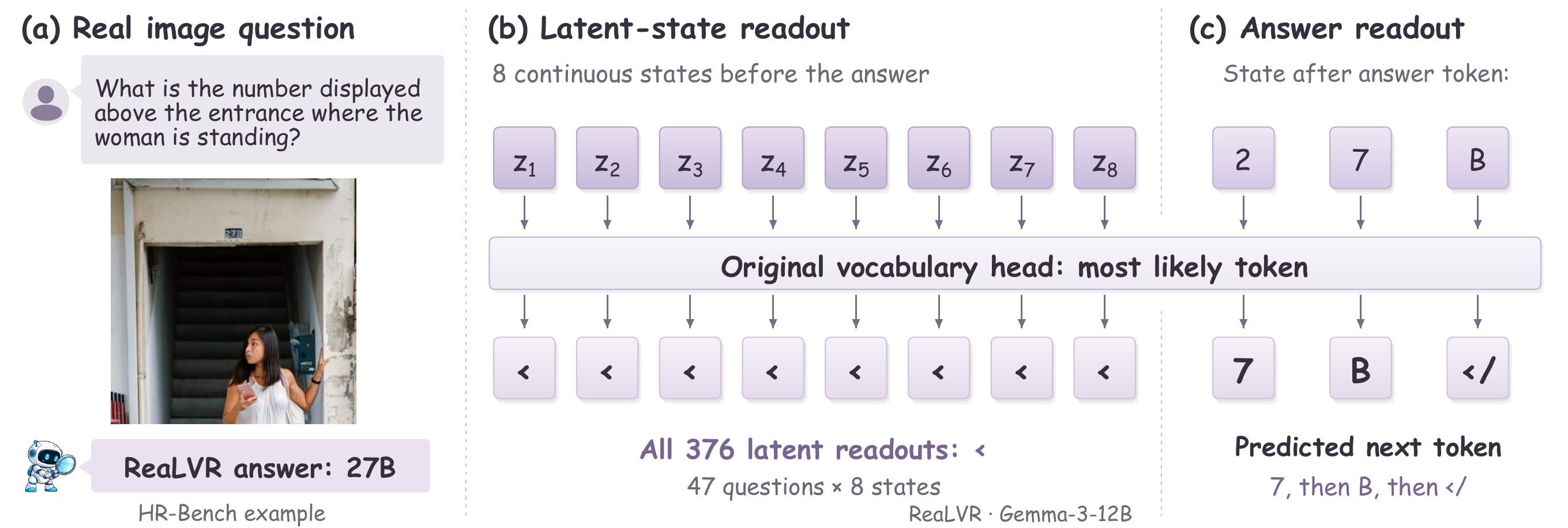}
  \caption{\small\textbf{A real visual answer and its latent-state text readout.}
  ReaLVR (\texttt{Gemma-3-12B}) answers ``27B'' from an HR-Bench image (a).
  The original vocabulary head reads each of its eight latent states as
  \texttt{<}, as it does all $376$ states from $47$ questions (b), yet
  predicts \texttt{7} after \texttt{2}, \texttt{B} after \texttt{7}, and
  \texttt{</} after \texttt{B} during answer generation (c). The latent
  readouts do not form an intermediate explanation.}
  \label{fig:latent-vocabulary-readout}
\end{figure}

\FloatBarrier
\paragraph{Useful latent computation need not be verbal.}
ReaLVR trains continuous states to preserve visual evidence, rather than
produce intermediate sentences. We inspect their textual readout by applying
the \texttt{Gemma-3-12B} vocabulary head to actual latent vectors from
$47$ questions with $8$ states each, with special tokens masked. All $376$
projections have \texttt{<} as their top-1 token with probability $1.0$,
while the same head gives correct next-token predictions for the answer
``27B'' in Figure~\ref{fig:latent-vocabulary-readout}. We attribute this
readout to the closing-tag prior learned in Stage~1: \texttt{<} begins the
literal \texttt{<|lvr\_end|>} that follows every latent span, and the latent
states align with its unembedding direction (cosine
$0.41$ versus $0.02$ for a random
vocabulary row). Top-1 vocabulary readout therefore does not expose a
language rationale, yet the states are not empty: a linear probe recovers
the BLINK task label from the mean latent with $99.9\%$ accuracy
(Appendix~\ref{app:diag-counterfactual}). Continuous states can thus support answers without expressing a
rationale~\citep{hao2024continuous}.

A fixed-context replacement test probes whether the answer uses these states:
replacing the eight most answer-attended tokens while holding other states
fixed lowers ReaLVR's correct-answer probability from $0.70$ to $0.59$
(Figure~\ref{fig:accuracy-utility}; Appendix~\ref{app:diag-latent-dependence}).
Thus, nonverbal latent states affect answer likelihood locally;
regenerating later states could introduce further effects.

\paragraph{ReaLVR attends more strongly to target regions.}
Target-region enrichment divides answer attention on an annotated region
by that on same-area background windows, reducing region-size effects.
A value of $1$ means equal attention; $2$ means twice as much on the target.
ReaLVR reaches about $2$ in middle layers, versus Monet near $1.6$ and
LVR at or below $1.3$ (Figure~\ref{fig:attention-ratio-look}). Correct
and incorrect ReaLVR responses have similar curves, so spatial alignment
alone cannot explain correctness, consistent with prior VLM
observations~\citep{liu2026seeing}. Together with fixed-context replacement,
this probes \textit{where answers attend and whether selected latent states
affect answer likelihood under controlled replacement with other states
held fixed}~\citep{reich2023measuring}.

\begin{samepage}
\section{Conclusion}
\label{sec:conclusion}

In this work, we study a fundamental challenge in latent visual reasoning: a correct final answer does not guarantee that the preceding latent tokens have learned to preserve the visual evidence necessary to produce it. We identify this missing link as the \emph{latent evidence-credit gap}. To address it, we introduce \text{ReaLVR}, which contrasts visual evidence to teach latent tokens \emph{what} to preserve, while comparing how the correct and wrong answers attend to these tokens to decide \emph{where} supervision is most critical. Across the tested backbones up to 235B, evidence supervision improves the available LVR baselines without changing model architectures or inference procedures. These results support visual-evidence supervision as a way to improve the grounding and use of continuous latent states.

\par\end{samepage}

\FloatBarrier

%
%

\setlength{\bibsep}{7pt plus 2pt minus 2pt}
\bibliography{references}
\bibliographystyle{realvr_preprint}

\appendix
\clearpage
\phantomsection
\pdfbookmark[0]{Technical Appendices}{app:technical-appendix}
\label{app:technical-appendix}

\hypersetup{linktoc=all}
\fancyhead[L]{\amazonHeaderMark}
\fancyhead[R]{\realvrHeaderDate}
\raggedbottom
\setcounter{table}{0}
\setcounter{figure}{0}
\setcounter{equation}{0}
\makeatletter
\@addtoreset{table}{section}
\@addtoreset{figure}{section}
\@addtoreset{equation}{section}
\renewcommand{\theHtable}{\thesection.\arabic{table}}
\renewcommand{\theHfigure}{\thesection.\arabic{figure}}
\renewcommand{\theHequation}{\thesection.\arabic{equation}}
\setlength{\@fptop}{0pt}
\setlength{\@fpsep}{12pt}
\newcommand{\realvrAppendixEntry}[3]{%
  \addtocontents{atoc}{\protect\contentsline{#1}{\protect\numberline{#2}#3}{\thepage}{\@currentHref}}}
\let\realvrAppendixSection\section
\renewcommand{\section}[1]{%
  \FloatBarrier
  \realvrAppendixSection{#1}%
  \realvrAppendixEntry{section}{\thesection}{#1}}
\let\realvrAppendixSubsection\subsection
\renewcommand{\subsection}[1]{%
  \realvrAppendixSubsection{#1}%
  \realvrAppendixEntry{subsection}{\thesubsection}{#1}}
\makeatother
\renewcommand{\thetable}{\Alph{section}.\arabic{table}}
\renewcommand{\thefigure}{\Alph{section}.\arabic{figure}}
\renewcommand{\theequation}{\Alph{section}.\arabic{equation}}
\setlength{\floatsep}{10pt plus 2pt minus 2pt}
\setlength{\textfloatsep}{12pt plus 2pt minus 2pt}
\setlength{\intextsep}{10pt plus 2pt minus 2pt}
\setlength{\abovecaptionskip}{5pt}
\setlength{\belowcaptionskip}{3pt}

\noindent\rule{\linewidth}{1.5pt}\par
\vspace{7pt}
\begin{center}
  {\fontsize{25}{29}\selectfont\bfseries Technical Appendices}
\end{center}
\vspace{4pt}
\noindent\rule{\linewidth}{0.6pt}\par
\vspace{20pt}
{\fontsize{16}{20}\selectfont\bfseries Table of Contents}\par
\vspace{6pt}
\begingroup
\parskip=0pt
\fontsize{10}{13}\selectfont
\setcounter{tocdepth}{2}
\makeatletter
\def\@pnumwidth{1.9em}
\def\@tocrmarg{2.6em}
\renewcommand*{\l@section}[2]{%
  \addpenalty{-\@highpenalty}%
  \addvspace{8pt}%
  \begingroup
    \parindent\z@ \rightskip\@pnumwidth \parfillskip-\@pnumwidth
    \@tempdima=1.8em
    \leavevmode\bfseries
    \advance\leftskip\@tempdima \hskip-\leftskip
    #1\nobreak\hfil\nobreak\hb@xt@\@pnumwidth{\hss #2}\par
  \endgroup}
\renewcommand*{\l@subsection}{\@dottedtocline{2}{1.8em}{2.5em}}
\@starttoc{atoc}
\makeatother
\endgroup
\vspace{24pt}

\section{Training Details and Hyperparameters}
\label{app:hparams}

All backbones are trained with the two-stage recipe of Section~\ref{sec:prelim_lvr}:
Stage~1 (target-conditioned visual supervision) followed by
Stage~2 (free-running outcome optimization with the ReaLVR evidence loss). The recipe and all optimization hyperparameters are
shared across backbones; only the number of GPUs, and hence the global batch
size, changes with model scale.

\paragraph{Hardware.}
All models are trained on a cluster of AMD Instinct MI250X accelerators.
An MI250X is a dual-die package: it holds two Graphics Compute Dies (GCDs),
each with its own 64\,GB of HBM2e memory, and the ROCm runtime exposes every
GCD as a separate device. Each node in our cluster contains four MI250X
accelerators and therefore provides eight independently addressable GPUs
with 64\,GB each. Following common practice on MI250X systems, we refer to a
GCD as a GPU when referring to ROCm-visible devices. The 235B model is
trained on 200 nodes, i.e.\ 800 MI250X accelerators exposed as 1{,}600
GCDs; the 7B, 8B, and 30B backbones use 64 nodes (256 MI250X
accelerators, 512 GCDs). World sizes and per-device batch counts below
refer to GCDs.

\paragraph{Distributed training.}
We use DeepSpeed ZeRO-3 with full parameter, gradient, and optimizer-state
partitioning across all GPUs and no CPU offload; communication is overlapped
with computation. Table~\ref{tab:hp-dist} summarizes the configuration.

\begin{table}[ht]
\caption{\textbf{Hardware and distributed training configuration.} Each MI250X
accelerator contributes two ROCm-visible 64\,GB GCDs.}
\label{tab:hp-dist}
\centering
\appendixTableSetup
\begin{tabularx}{\linewidth}{@{}>{\raggedright\arraybackslash}p{0.43\linewidth}>{\raggedright\arraybackslash}X@{}}
\toprule
\textbf{Item} & \textbf{Value} \\
\midrule
Accelerator & AMD Instinct MI250X (2 GCDs per accelerator) \\
Per node & 4 MI250X $=$ 8 GPUs (GCDs), 64\,GB HBM2e per GPU \\
Accelerators, 235B & 200 nodes $\times$ 4 $=$ 800 MI250X (1{,}600 GCDs) \\
Accelerators, $\leq$30B & 64 nodes $\times$ 4 $=$ 256 MI250X (512 GCDs) \\
Parallelism & DeepSpeed ZeRO-3 (parameters, gradients, optimizer states) \\
Offload & None \\
\texttt{overlap\_comm} & True \\
\texttt{stage3\_max\_live\_parameters} & $1\times10^{9}$ \\
Precision & bf16 \\
Gradient checkpointing & Enabled \\
Attention kernel & PyTorch SDPA \\
Launcher & Multi-node DeepSpeed launcher (hostfile) \\
\bottomrule
\end{tabularx}
\end{table}

\paragraph{Stage 1: target-conditioned visual supervision.}
Stage~1 trains the language model on the next-token loss plus the visual
reconstruction loss, with the vision encoder and the
vision--language merger frozen. Latent states are fed back as continuous
hidden states (Section~\ref{sec:prelim_lvr}); no separate latent projection head
is used. Sequences are packed to 4{,}096 tokens, and each image contributes
between 128 and 5{,}120 visual tokens depending on its resolution. Each GPU
processes one packed sequence per step without gradient accumulation, so the
global batch equals the number of GPUs.

\begin{table}[ht]
\caption{\textbf{Stage-1 hyperparameters.}}
\label{tab:hp-sft}
\centering
\appendixTableSetup
\begin{tabularx}{\linewidth}{@{}>{\raggedright\arraybackslash}p{0.43\linewidth}>{\raggedright\arraybackslash}X@{}}
\toprule
\textbf{Hyperparameter} & \textbf{Value} \\
\midrule
Per-GPU batch $\times$ grad.\ accumulation & $1 \times 1$ \\
Global batch (sequences) & 1{,}600 (235B) / 512 ($\leq$30B) \\
Optimizer & AdamW \\
Peak learning rate & $1\times10^{-5}$ \\
LR schedule / warmup ratio / weight decay & cosine / 0.03 / 0.1 \\
Reconstruction loss & MSE, weight $0.1$ \\
Latent feedback & continuous hidden state \\
Latent projection head & none \\
Trainable modules & language model \\
Frozen modules & vision tower, merger \\
Sequence packing / max packed tokens & yes / 4{,}096 \\
Max sequence length & 4{,}096 \\
Visual tokens per image & 128--5{,}120 \\
\bottomrule
\end{tabularx}
\end{table}

\paragraph{Stage 2: free-running outcome optimization with evidence credit.}
Stage~2 optimizes the objective. For every prompt the
behavior policy samples $G=8$ completions at temperature 0.6, which supply
both the GRPO advantages and the wrong-answer set $\mathcal{Y}^-_x$. We set the KL weight $\beta$ to 0, so the Stage-1 model serves only as the initialization.
Images are capped at 2{,}560 visual tokens ($\approx$2.0\,MP). Training
prompts are drawn from a mixture of ViRL39K and Visual-CoT. Stage~2 runs for
100 optimizer steps with a checkpoint every 25 steps. And all baselines such as LVR and Monet use the same data and supervision for training. The evidence loss uses $K={8}$ latent tokens, weight
$\lambda_{\mathrm{ev}}={0.2}$, margin $m_{\mathrm{ev}}={0.5}$,
uniform baseline $\eta={0.3}$, and $N^-={16}$ negative prototypes
pooled from other examples in the same global batch; the same $K$ is used at
inference.

\begin{table}[ht]
\caption{\textbf{Stage-2 hyperparameters (GRPO with ReaLVR evidence supervision).}}
\label{tab:hp-rl}
\centering
\appendixTableSetup
\begin{tabularx}{\linewidth}{@{}>{\raggedright\arraybackslash}p{0.43\linewidth}>{\raggedright\arraybackslash}X@{}}
\toprule
\textbf{Hyperparameter} & \textbf{Value} \\
\midrule
Objective & GRPO $+$ evidence loss \\
Group size $G$ (rollouts per prompt) & 8 \\
Sampling temperature / top-$p$ / top-$k$ & 0.6 / 1.0 / off \\
KL weight $\beta$ & 0 \\
Per-GPU batch $\times$ grad.\ accumulation & $1 \times 1$ \\
Prompts per step & 1{,}600 (235B) / 512 ($\leq$30B) \\
Optimizer & AdamW \\
Peak learning rate & $5\times10^{-7}$ \\
LR schedule / warmup ratio / weight decay & cosine / 0.03 / 0.1 \\
Max prompt length & 4{,}096 \\
Max completion length & 192 \\
Max visual tokens per image & 2{,}560 \\
Training steps & 100 \\
Checkpoint interval & 25 steps \\
Latent length $K$ (training and inference) & 8 \\
Evidence loss weight $\lambda_{\mathrm{ev}}$ & 0.2 \\
Target margin $m_{\mathrm{ev}}$ & 0.5 \\
Uniform baseline $\eta$ & 0.3 \\
Negative prototypes per example $N^-$ & 16 (pooled from other examples in the global batch) \\
Training data & ViRL39K $+$ Visual-CoT mixture \\
\bottomrule
\end{tabularx}
\end{table}

\section{The Latent Evidence-Credit Gap}
\label{app:latent-evidence-credit-gap}

Equation~\ref{eq:lvr_s2} assigns an outcome to an entire completion. The reward
identifies a successful answer, while leaving open which latent tokens
supported it and what visual evidence they preserved. Vanilla LVR supplies
visual targets only to target-conditioned Stage~1 states, leaving the
free-running states used in Stage~2 and at inference without direct
visual-evidence supervision.

We separate three questions about a generated latent token. \emph{Readout}
asks whether the answer decoder attends to it. \emph{Grounding} asks whether
it represents the evidence relevant to the image--question pair rather than
mismatched evidence. \emph{Utility} asks whether intervening on the token
changes the answer. These questions require distinct measurements: a token
can receive attention without representing relevant evidence or affecting the
answer. Readout shared by correct and behavior-policy wrong answers can also
reflect formatting or transition behavior common to both outcomes.

ReaLVR addresses the training gap with outcome-contrastive readout credit.
It supervises one regenerated current-model trajectory with visual evidence
and gives greater weight to positions read more strongly under the correct
answer than under sampled wrong answers. Both readouts use the current model
and the same regenerated latent span. Intervention utility remains a separate
evaluation criterion.

\begin{figure}[!htbp]
  \centering
  \includegraphics[width=\linewidth]{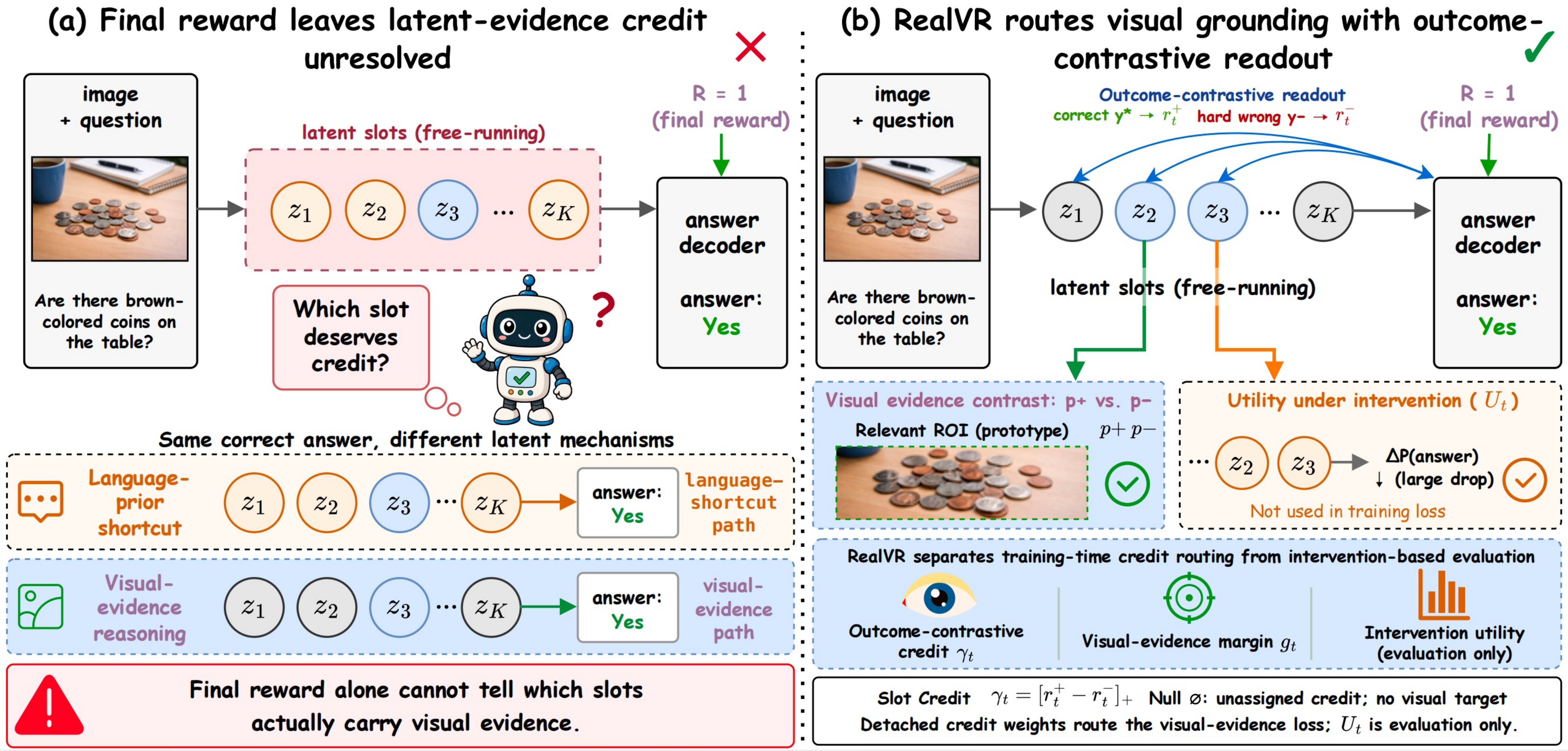}
  \caption{\textbf{From final reward to latent evidence credit.} The output
  reward indicates whether the answer is correct. ReaLVR uses relevant and
  mismatched visual evidence to determine what the latent tokens should
  preserve, then contrasts ground-truth and wrong-answer readouts to determine
  where supervision should be stronger. Intervention separately evaluates
  whether the answer depends on the latent tokens and is not part of the
  training objective.}
  \label{fig:realvr-method}
\end{figure}

Figure~\ref{fig:realvr-method} separates the roles of outcome reward, visual
evidence supervision, and intervention-based evaluation.

\section{Experimental and Training Details}
\label{app:experimental-settings}
\label{app:credit-construction}

\subsection{Benchmarks and Model Coverage}
\label{app:benchmark-model-coverage}

\texttt{MMVP}~\citep{tong2024eyes} tests subtle visual discrimination,
while \texttt{BLINK}~\citep{fu2024blink} evaluates visual perception tasks,
including counting, jigsaw, spatial relations, and depth.
\texttt{HRBench-4K} and \texttt{HRBench-8K}~\citep{wang2025hrbench}
test high-resolution perception, and
\texttt{MME-RealWorld}~\citep{zhang2025mmerealworld} tests understanding
of complex real-world scenes. Together, these benchmarks cover visual
reasoning at different image resolutions.

Five backbones have complete five-benchmark results:
\texttt{Qwen2.5-VL-7B}~\citep{bai2025qwen25vl},
\texttt{Qwen3-VL-8B} and \texttt{Qwen3-VL-30B}~\citep{bai2025qwen3vl},
\texttt{InternVL3-8B}~\citep{zhu2025internvl3}, and
\texttt{Gemma-3-12B}~\citep{gemmateam2025gemma3}.
Here, \texttt{Qwen3-VL-30B} abbreviates \texttt{Qwen3-VL-30B-A3B}.
We additionally evaluate \texttt{Qwen3-VL-235B-A22B}, abbreviated as
\texttt{Qwen3-VL-235B}, on \texttt{MMVP}, \texttt{BLINK}, and
\texttt{MME-RealWorld}, comparing direct decoding, LVR-SFT, and ReaLVR.

\subsection{Evaluation Protocol and Metrics}
\label{app:evaluation-protocol}

Table~\ref{tab:main-backbone-results} compares Pixel
Reasoner~\citep{su2025pixelreasoner}, Vision-R1~\citep{huang2026visionr1},
LVR~\citep{li2025latentvisualreasoning}, ILVR~\citep{dong2025ilvr}, and
Monet~\citep{wang2025monet} at the training stages indicated in each row.
Table~\ref{tab:cross-backbone-results} reports the Qwen3-VL size series
and the InternVL3 and Gemma-3 family comparisons. The latter use a matched
evaluation protocol within each family.

We report task accuracy and compute an unweighted mean only for rows with
all five benchmark scores. Average gains are differences between the
displayed one-decimal means. The three-benchmark $235$B evaluation is
reported task by task and has no five-benchmark average. Mechanism analyses
examine latent variation, visual grounding, and answer dependence;
Appendix~\ref{app:diagnostics} defines the region, saliency, and
fixed-context token-replacement estimators.

The training-time construction rules below complete the derivation in
Section~\ref{sec:method}, using the notation of
Section~\ref{sec:preliminaries}.

\subsection{Visual Prototypes}
\label{app:construction-prototypes}

If the ROI annotation is absent or its visual-token mask is empty, we set
$a_n^+=1$ for all $n$, giving a whole-image target $p^+$. We stabilize the
pooling denominator and require nonzero prototypes for the cosine margin.
The negative set $\mathcal N$ contains at least one valid prototype; each
$p_s^-$ uses the same pooling rule on a designated mismatched example.
Frozen vision and connector weights keep prototypes fixed while the evidence
loss updates the language model through the regenerated trajectory.

\subsection{Wrong-Answer Construction}
\label{app:construction-wrong-answers}

We require $\mathcal J(y^\star)\neq\emptyset$. After parsing and
canonicalization, the retained set is
\[
    \mathcal Y_x^-
    =\operatorname{Unique}\!\left\{
      \widehat y_i:
      i\in\{1,\ldots,G\},
      \widehat y_i\neq\bot,
      \operatorname{Correct}(\widehat y_i,y^\star)=0,
      \mathcal J(\widehat y_i)\neq\emptyset
      \right\}.
\]
This rule removes parse failures, correct answers, candidates without
answer-content positions, and duplicates. The retained strings come from the
behavior policy and supply comparison outcomes, regardless of their
probability under the updated model.

Every formatted candidate is teacher-forced after the same regenerated latent
span. The readout in Section~\ref{sec:method_readout_gate} averages the selected
decoder layers, heads, and answer-content positions.

\subsection{Dependencies and Gradient Flow}
\label{app:construction-gradient-flow}

For minibatch example $b$, the group outputs
$\{o_{b,i}\}_{i=1}^{G}$ construct $\mathcal Y_{x_b}^-$. The loss also depends
on $(x_b,a_b,y_b^\star)$ and the supplied visual negative set $\mathcal N_b$;
we leave these dependencies implicit in $w_{b,t}$ and $g_{b,t}$.
Gradients pass through the autoregressive latent-generation process, so a
loss term at position $t$ can update earlier generation steps and shared
parameters. A token's weight specifies its contribution to the loss, not an
update restricted to that position.

\section{Uniform Bootstrap and Unassigned Mass}
\label{app:uniform-bootstrap}

\subsection{Selective Credit and Its Remainder}
\label{app:bootstrap-remainder}

Section~\ref{sec:method_readout_gate} defines
$\gamma_t=[r_t^+-r_t^-]_+$. Because $\gamma_t\leq r_t^+$ and the raw
attention mass over the latent span is at most one,
$\sum_t\gamma_t\leq1$. We therefore define the nonnegative bookkeeping
remainder
\[
    \gamma_{\emptyset}=1-\sum_{t=1}^{K}\gamma_t.
\]
Here $\emptyset$ labels unassigned mass; it is distinct from $a=\emptyset$,
the notation for a missing ROI annotation.

\subsection{Uniform Allocation and Boundary Cases}
\label{app:bootstrap-allocation}

Section~\ref{sec:method_objective} mixes selective credit with a uniform
component. The full allocation is
\[
    w_t
    =\frac{\eta}{K}+(1-\eta)\gamma_t,
    \qquad
    w_{\emptyset}=(1-\eta)\gamma_{\emptyset}.
\]
For $0<\eta<1$, the first term gives every latent token a nonzero routing
weight, while the second term preserves selective routing; $\eta=0$ and
$\eta=1$ recover the selective-only and uniform-only endpoints. Because
$\gamma_{\emptyset}+\sum_t\gamma_t=1$, the complete allocation satisfies
$w_{\emptyset}+\sum_t w_t=1$. The remainder $w_{\emptyset}$ is bookkeeping
only: it introduces no token, module, or loss term, and the token weights
retain their original mass without renormalization.

If $\mathcal{Y}_x^-=\emptyset$, then $\gamma_t=0$ and $w_t=\eta/K$.
Thus, $\eta>0$ retains uniform supervision when no valid wrong-answer
comparison is available, while $\eta=0$ assigns zero evidence weight to that
example.

\section{Detached-Credit Gradient Decomposition}
\label{app:detached-credit}

Detaching the readout-derived weights lets them allocate supervision while
the evidence gradient improves the visual margin. The following decomposition
shows which gradient path detachment removes.

\subsection{Local Derivative of the Undetached Objective}
\label{app:gradient-undetached}

At a given optimization update, we condition
on the sampled candidate-answer set, its tokenized answer sequences, the
selected decoder layers and heads, the answer-position masks, the visual
prototypes, and $\eta$. These quantities are fixed for the local gradient
calculation; the dependence on $\theta$ below comes from the regenerated
trajectory and its current-model readout. For one regenerated trajectory,
define the per-token evidence violation
\[
    h_t(\theta)
    =[m_{\mathrm{ev}}-g_t(\theta)]_+,
\]
and temporarily view
$w_t(\theta)=\eta/K+(1-\eta)\gamma_t(\theta)$ as an ordinary
differentiable function. The corresponding hypothetical undetached objective
is
\[
    \ell_{\mathrm{ev}}^{\mathrm{undet}}(\theta)
    =\sum_{t=1}^{K}w_t(\theta)h_t(\theta).
\]
Away from the kink points of the positive-part operators, the product rule
gives
\[
    \nabla_\theta\ell_{\mathrm{ev}}^{\mathrm{undet}}
    =
    \sum_{t=1}^{K}
    \underbrace{w_t\nabla_\theta h_t}_{
      \text{update the visual evidence margin}}
    +
    \sum_{t=1}^{K}
    \underbrace{h_t\nabla_\theta w_t}_{
      \text{update the credit router}}.
\]
At a hinge or ReLU kink, or when multiple visual negatives attain the same
maximum, automatic differentiation selects a subgradient and the same two
computational-graph paths remain. The first term is the intended weighted
evidence update. The second changes the router in proportion to the current
evidence violation. In particular,
$\partial\ell_{\mathrm{ev}}^{\mathrm{undet}}/\partial w_t=h_t\geq0$:
when the hinge is active, gradient descent has a local path to reduce the loss
by lowering the token's weight. More explicitly, at smooth points,
\[
    \nabla_\theta h_t
    =-\mathbf 1\{g_t<m_{\mathrm{ev}}\}\nabla_\theta g_t,
    \qquad
    \nabla_\theta w_t
    =(1-\eta)\mathbf 1\{r_t^+>r_t^-\}
      (\nabla_\theta r_t^+-\nabla_\theta r_t^-).
\]
Thus, the routing term can lower $r_t^+$ or raise $r_t^-$ where the
readout difference is active, reducing the mass assigned to the token and
increasing the bookkeeping remainder without improving the token's visual
evidence margin. The decomposition therefore identifies a local optimization
shortcut through the router.

\subsection{The Detached Update}
\label{app:gradient-detached-update}

The stop-gradient operator preserves the forward value,
$\operatorname{sg}(w_t)=w_t$, but sets its derivative to zero. At
optimization step $k$, it is equivalent for this branch to differentiating
the local surrogate
\[
    \widetilde\ell_{\mathrm{ev},k}(\theta)
    =\sum_{t=1}^{K}w_t(\theta_k)h_t(\theta).
\]
Its gradient at the current parameters is
\[
    \left.\nabla_\theta\widetilde\ell_{\mathrm{ev},k}(\theta)
    \right|_{\theta=\theta_k}
    =
    \sum_{t=1}^{K}
    w_t(\theta_k)
    \left.\nabla_\theta h_t(\theta)\right|_{\theta=\theta_k}.
\]
The weights are recomputed at each update, so this surrogate describes the
local gradient at step $k$. The evidence branch uses the current readout to
allocate supervision and improves the visual margin through the latent
trajectory. Readout still evolves across updates through GRPO, KL
regularization, and shared-parameter changes. Detachment removes only the
evidence loss's direct gradient through the routing weights.

\section{Latent-Length Ablation}
\label{app:latent-length-ablation}
\label{sec:latent-length-ablation}

\subsection{Evaluation Setup}
\label{app:latent-length-setup}

We hold the \texttt{Qwen2.5-VL-7B} ReaLVR checkpoint fixed within this
ablation and vary the prescribed inference-time latent length $K$.
The $K=0$ setting skips the latent span and decodes the answer directly.

\subsection{Task-Dependent Budgets}
\label{app:latent-length-results}

\begin{table}[!htbp]
\caption{\textbf{Sensitivity to the inference-time latent budget.}
The ReaLVR (\texttt{Qwen2.5-VL-7B}) checkpoint is held fixed while the latent-token
budget $K$ is varied at inference; $K=0$ decodes the answer without a latent span.
$K=8$ is the budget used during training and serves as the reference ($\dagger$):
entries are differences in accuracy (percentage points) from that column, which is
therefore $0.0$ by construction. The sweep is run on a fixed evaluation subset
so that all budgets are
scored identically; its absolute scores are consequently not on the same scale as
Table~\ref{tab:main-backbone-results}, and the differences reported here are meaningful only within
this sweep. Under the full evaluation protocol of Table~\ref{tab:main-backbone-results}, the $K=8$
setting scores $72.0$ on \texttt{MMVP}, $55.8$ on \texttt{BLINK}, $66.6$ on
\texttt{HR-8K} and $52.2$ on \texttt{MME-RealWorld}. Bold marks each row's largest
value, including ties. The mean is the unweighted average across the four listed
benchmarks.}
\label{tab:latent-length-ablation}
\centering
\appendixTableSetup
\begin{tabular*}{\linewidth}{@{\extracolsep{\fill}}lrrrrrrr@{}}
\toprule
\textbf{Benchmark} & \multicolumn{7}{c}{\textbf{Latent-token budget} $K$ \textit{(difference from} $K=8$\textit{)}} \\
\cmidrule(l){2-8}
& $0$ & $2$ & $4$ & $8^{\dagger}$ & $12$ & $16$ & $20$ \\
\midrule
\texttt{HR-8K}  & $-3.0$ & $-2.0$ & $-1.5$ & $\phantom{+}\mathbf{0.0}$ & $-1.0$ & $-0.5$ & $\phantom{+}\mathbf{0.0}$ \\
\texttt{MMVP}   & $-0.7$ & $-1.0$ & $-0.3$ & $\phantom{+}\mathbf{0.0}$ & $-1.0$ & $-1.0$ & $\phantom{+}\mathbf{0.0}$ \\
\texttt{BLINK}  & $-4.0$ & $\phantom{+}0.0$ & $-1.1$ & $\phantom{+}0.0$ & $+1.2$ & $\mathbf{+2.3}$ & $+0.9$ \\
\texttt{MME-RealWorld} & $-4.4$ & $+1.9$ & $\mathbf{+2.1}$ & $\phantom{+}0.0$ & $-0.2$ & $-1.4$ & $-2.1$ \\
\midrule
\textbf{Mean}   & $-3.0$ & $-0.3$ & $-0.2$ & $\phantom{+}\mathbf{0.0}$ & $-0.2$ & $-0.1$ & $-0.3$ \\
\bottomrule
\end{tabular*}
\end{table}
\paragraph{Additional steps help different tasks to different degrees.}

Table~\ref{tab:latent-length-ablation} shows that the best observed length is
task dependent. \texttt{HR-8K} first reaches its maximum at $K=8$, while
\texttt{BLINK} peaks at $K=16$, improving by $6.3$ points over $K=0$. In
contrast, \texttt{MME-RealWorld} peaks at $K=4$ and then declines by $4.2$
points by $K=20$. \texttt{MMVP} varies by
only $1.0$ point across all tested lengths and shows no consistent trend.
Together, these results favor task-dependent budgets over uniformly longer
trajectories.

\textbf{A short latent span captures much of the benefit.}
Across the four tasks in this ablation, $K=8$ gives the highest mean accuracy;
the closest alternative, $K=16$, is 0.1 points below it, and removing the latent
span entirely ($K=0$) costs 3.0 points. Selecting the best tested $K$ separately
for each benchmark gives a post-hoc task-level oracle 1.1 points above $K=8$,
which motivates adaptive latent budgets. Even $K=2$ captures 4.0 of the
6.3-point maximum gain on \texttt{BLINK} and 6.3 of the 6.5-point maximum gain
on \texttt{MME-RealWorld}.

\subsection{Components Ablations}
\label{sec:mablation}

Table~\ref{tab:mablation} removes each ingredient of ReaLVR on
\texttt{Qwen2.5-VL-7B}. Every variant stays above LVR-RL, so the evidence loss
helps in any form, but the full method is best on all five benchmarks.

\textbf{Answer-contrast routing matters.} Replacing the routing weights with
a uniform $1/K$ ($\eta=1$) costs {1.3} points on average and
{2.5} on \texttt{MMVP}: the same visual supervision, spread evenly
over the latent span, is markedly less effective than supervision
concentrated where the correct answer reads. Subtracting the wrong-answer
readout is part of this effect; routing with raw correct-answer attention
alone recovers only {62.9}, because attention shared by correct and
wrong answers (formatting, transitions) then receives credit. Removing the
uniform floor entirely ($\eta=0$) is slightly worse than the full model
({63.2} vs.\ $63.7$), consistent with its role of keeping supervision
alive on examples without a valid wrong answer.

\textbf{Negatives make the target discriminative.} Aligning latents to $p^+$
without mismatched prototypes drops {1.6} points, the largest loss
among the ``what'' and ``where'' components; a plain alignment objective
pulls latents toward generic image content rather than toward what
distinguishes this image from others.

\textbf{On-policy regeneration and detachment are both necessary.}
Supervising the saved rollout latents instead of a regenerated trajectory
loses {1.8} points, the largest drop in the table, confirming that the
loss must reach the process that produces the latents at inference. Removing
the stop-gradient on $w_t$ loses {1.1} points, matching the shortcut
identified in Appendix~\ref{app:detached-credit}: the router lowers weights on hard
positions instead of improving their evidence margin.

\begin{table}[t]
\caption{\textbf{Component ablation on \texttt{Qwen2.5-VL-7B}.} Each row
removes or replaces one ingredient of ReaLVR; all other settings follow
Appendix~\ref{app:hparams}. ``Uniform routing'' sets $\eta=1$ so every latent
position receives weight $1/K$ and the answer contrast is unused.
``Selective only'' sets $\eta=0$. ``No negatives'' replaces the margin with a
plain cosine alignment to $p^+$. ``Raw attention'' routes with $r_t^+$ instead
of $[r_t^+-r_t^-]_+$. ``Undetached'' removes the stop-gradient on $w_t$.
``Off-policy'' applies the evidence loss to the saved rollout latents instead
of a regenerated trajectory. LVR-RL is the $\lambda_{\mathrm{ev}}=0$ endpoint.}
\label{tab:mablation}
\centering
\appendixTableSetup
\begin{tabularx}{\linewidth}{@{}>{\raggedright\arraybackslash}X*{6}{r}@{}}
\toprule
\textbf{Variant} & \texttt{MMVP} & \texttt{BLINK} & \texttt{HR-4K} & \texttt{HR-8K} & \texttt{MME-RW} & \textbf{Avg.} \\
\midrule
LVR-RL (no evidence loss) & 64.2 & 53.6 & 69.6 & 64.4 & 50.1 & 60.4 \\
\midrule
\multicolumn{7}{l}{\textit{Where to supervise (answer contrast)}} \\
\quad Uniform routing ($\eta=1$)        & {69.5} & {54.6} & {71.0} & {65.6} & {51.3} & {62.4} \\
\quad Raw attention ($\gamma_t=r_t^+$)  & {70.6} & {55.1} & {71.3} & {66.0} & {51.7} & {62.9} \\
\quad Selective only ($\eta=0$)         & {71.2} & {55.3} & {71.4} & {66.2} & {51.9} & {63.2} \\
\midrule
\multicolumn{7}{l}{\textit{What to preserve (visual contrast)}} \\
\quad No negatives ($\mathcal{N}=\emptyset$) & {69.0} & {54.4} & {70.6} & {65.3} & {51.0} & {62.1} \\
\midrule
\multicolumn{7}{l}{\textit{Training mechanics}} \\
\quad Undetached weights  & {70.1} & {54.8} & {71.1} & {65.5} & {51.4} & {62.6} \\
\quad Off-policy targets  & {68.8} & {54.2} & {70.4} & {65.1} & {50.8} & {61.9} \\
\midrule
\textbf{ReaLVR (full)} & 72.0 & 55.8 & 71.8 & 66.6 & 52.2 & \textbf{63.7} \\
\bottomrule
\end{tabularx}
\end{table}

\section{Additional Diagnostic Results}
\label{app:diagnostics}

The diagnostics examine answer dependence, visual grounding, representation
geometry, and sensitivity to image edits. Each subsection defines the
quantity being measured before presenting its results.

\subsection{Fixed-Context Latent-Token Dependence}
\label{app:diag-latent-dependence}
\begin{figure}[!htbp]
  \centering
  \includegraphics[width=\linewidth]{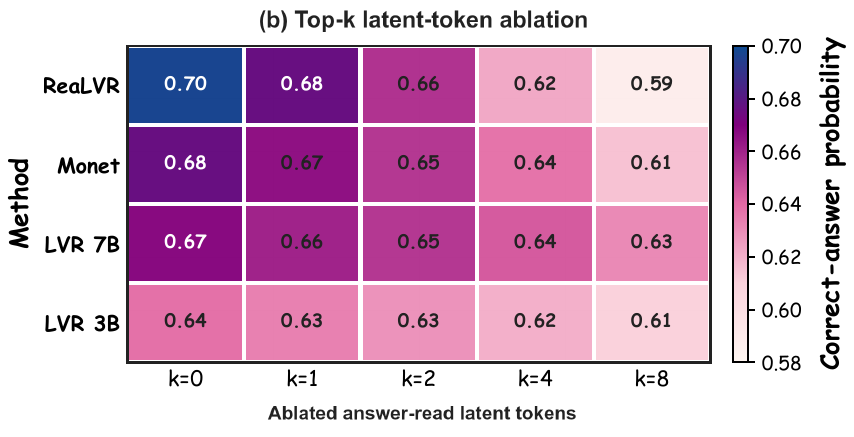}
  \caption{\textbf{Fixed-context latent-token dependence.}
  Each cell gives the correct-answer probability after replacing the top-$k$
  answer-read latent tokens, with all remaining latent states held fixed.
  Columns increase $k$ from $0$ to $8$. ReaLVR's probability falls from
  $0.70$ to $0.59$, the largest endpoint drop among the four methods.}
  \label{fig:accuracy-utility}
\end{figure}

Readout attention ranks latent tokens, but it does not show whether the answer
depends on them. The fixed-context audit replaces one latent token while
holding the others fixed and measures the change in target-answer
log-likelihood. Let
$o_{\mathrm{ans}}^\star$ be the canonical formatted target answer, whose
probability is computed by teacher forcing over the entire answer sequence,
and let $\bar z_t$ be a replacement token vector. Write
$\widetilde z_{1:K}^{(t\leftarrow \bar z_t)}$ for the latent sequence obtained
by replacing $z_t$ with $\bar z_t$. The fixed-context intervention score is
\begin{equation}
    u_t(\bar z_t)
    =\log\pi_\theta(o_{\mathrm{ans}}^\star\mid x,z_{1:K})
     -\log\pi_\theta\!\left(
       o_{\mathrm{ans}}^\star\mid
       x,\widetilde z_{1:K}^{(t\leftarrow \bar z_t)}
      \right).
    \label{eq:slot_utility}
\end{equation}
A positive value means that the original latent token gives the target answer
higher likelihood than its replacement. This is an evaluation metric, not part
of the \realvr{} training loss. Figure~\ref{fig:accuracy-utility}
reports the fixed-context intervention.

\paragraph{Answer-read latent tokens become more load-bearing.}
We rank latent tokens by answer-to-token attention, replace the top-$k$ tokens
while holding the remaining context fixed, and measure the correct-answer
probability. Figure~\ref{fig:accuracy-utility} shows the resulting
probability changes. For
ReaLVR, the correct-answer probability falls monotonically from $0.70$ at
$k=0$ to $0.59$ at $k=8$, a drop of $0.11$ that exceeds those of Monet and
both LVR variants. Together with
ReaLVR's higher five-benchmark average, this result is consistent with the
model placing more answer-relevant computation in the latent tokens that the
answer subsequently reads.

This drop quantifies local answer dependence on the selected latent tokens.

\subsection{Target-Region Attention Enrichment}
\label{app:diag-target-region}
For generated answer positions $\mathcal{J}$, heads $\mathcal{H}$, and image
token indices $\mathcal{R}\subseteq\{1,\ldots,N\}$, let
$A_{j,n}^{(\ell,h)}$ denote the post-softmax attention from answer position
$j$ to image token $n$ at layer $\ell$ and head $h$. We aggregate this
attention as
\[
    s_{\ell}(\mathcal{R})
    =
    \frac{1}{|\mathcal{J}|\,|\mathcal{H}|}
    \sum_{j\in\mathcal{J}}\sum_{h\in\mathcal{H}}
    \sum_{n\in\mathcal{R}} A_{j,n}^{(\ell,h)}.
\]
Let $\mathcal{M}$ be the annotated target region and let
$\Omega(\mathcal{M})$ contain same-area background windows outside it. We
report
\[
    \rho_{\ell}
    =
    \frac{s_{\ell}(\mathcal{M})}
    {\mathbb{E}_{\mathcal{B}\sim\Omega(\mathcal{M})}
       [s_{\ell}(\mathcal{B})]+\epsilon}.
\]
Here $\epsilon>0$ stabilizes the denominator, and $\rho_{\ell}=1$ indicates no
enrichment.

\begin{figure}[!ht]
  \centering
  \includegraphics[width=\linewidth,trim=0bp 100.56bp 0bp 0bp,clip]{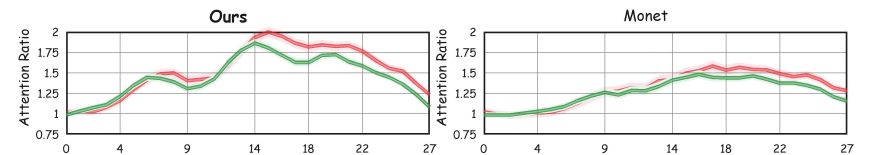}
  \par\vspace{-2pt}
  {\fontsize{9}{11}\selectfont
  \hspace*{0.035\linewidth}\makebox[0.50\linewidth][c]{Layer index}\makebox[0.465\linewidth][c]{Layer index}\par
  \vspace{3pt}
  \textcolor[RGB]{48,145,71}{\rule[0.5ex]{15pt}{1pt}}\hspace{4pt}Correctly answered
  \hspace{16pt}
  \textcolor[RGB]{222,38,51}{\rule[0.5ex]{15pt}{1pt}}\hspace{4pt}Incorrectly answered\par}
  \vspace{3pt}
  \includegraphics[width=\linewidth,trim=0bp 10.56bp 0bp 92.4bp,clip]{figures/attention_ratio_ours_monet_lvr_comic.pdf}
  \par\vspace{-2pt}
  {\fontsize{9}{11}\selectfont
  \hspace*{0.035\linewidth}\makebox[0.50\linewidth][c]{Layer index}\makebox[0.465\linewidth][c]{Layer index}\par}
  \caption{\textbf{Target-region attention enrichment.}
  Layer-wise answer attention on the annotated region relative to matched
  background windows. Values above $1$ indicate target-region enrichment;
  green and red curves correspond to correct and incorrect generations,
  respectively. Their overlap shows that alignment alone does not determine
  answer correctness.}
  \label{fig:attention-ratio-look}
\end{figure}

\paragraph{Layer-wise target-region attention results.}
\label{sec:where-to-look}
We next ask whether the answer attends to the visual region needed by the
question. The diagnostic compares answer-to-image attention on the annotated
target with same-area background windows, where a ratio of $1$ denotes no
enrichment. In Figure~\ref{fig:attention-ratio-look}, ReaLVR rises from near
$1$ in the lower layers to approximately $2$ in the middle layers, then
retains substantial enrichment through most upper layers. Monet peaks near
$1.6$, while the LVR curves remain at or below approximately $1.3$. The
correctly and incorrectly answered curves nevertheless track each other
closely, and the incorrect curve is sometimes higher. Thus, ReaLVR strengthens
spatial alignment, but looking at the target is not sufficient for answering
correctly. This distinction motivates separating a scalable grounding signal
from the answer-level learning signal.

\FloatBarrier

\subsection{Answer-Conditioned Saliency}
\label{app:diag-saliency}
After generating an answer
$o^{\mathrm{attr}}=(o_1^{\mathrm{attr}},\ldots,o_T^{\mathrm{attr}})$, we run a
teacher-forced attribution pass on its content positions $\mathcal{J}$. The
cross-entropy target is that generated sequence, aggregated as
\[
    \mathcal L_{\mathrm{CE}}(o^{\mathrm{attr}})
    =-\sum_{j\in\mathcal{J}}
      \log\pi_\theta\!\left(
        o_j^{\mathrm{attr}}
        \mid x,z_{1:K},o_{<j}^{\mathrm{attr}}
      \right).
\]
The score for image token $n$ is
\[
    S(n)=
    \frac{1}{|\mathcal{L}_{\mathrm{dec}}|\,|\mathcal{H}|\,|\mathcal{J}|}
    \sum_{\ell\in\mathcal{L}_{\mathrm{dec}}}
    \sum_{h\in\mathcal{H}}\sum_{j\in\mathcal{J}}
    \left|
    A^{(\ell,h)}_{j,n}
    \frac{\partial\mathcal L_{\mathrm{CE}}(o^{\mathrm{attr}})}
    {\partial A^{(\ell,h)}_{j,n}}
    \right|.
\]
We resize the visual-token grid to the image and normalize the scores within
each example. For panels explicitly labeled Saliency in
Figure~\ref{fig:token-saliency-full} and the additional cases, token-level
attribution retains the query-key matrix instead of projecting it onto
image patches: rows index queries and columns index keys. Each case shows
ReaLVR attention, LVR-7B attention, ReaLVR saliency, and LVR-7B saliency,
in that order. Color intensity is normalized within each panel.

\FloatBarrier
\subsection{Similarity and Representation Geometry}
\label{app:diag-geometry}
High cosine similarity can reflect several properties of a representation.
Table~\ref{tab:diag-raw-similarity} measures similarity within LVR
trajectories and compares latent tokens with same-prefix dummy tokens and
ordinary continuation tokens under different normalizations.
Table~\ref{tab:diag-rcs} then measures stability under perturbations and
effective rank. The latent trajectories remain stable and highly similar
while occupying fewer principal directions than the visual embeddings.
These measurements describe representation geometry; similarity alone does
not establish useful visual computation.

\begin{table}[!htbp]
\caption{\textbf{Latent-token similarity diagnostics.}
(a) Raw within-trajectory cosine similarity on two VISCOT examples.
(b) Context-normalized similarity on BLINK ($N{=}697$), with same-prefix dummy
and ordinary continuation tokens as references.}
\label{tab:diag-similarity}
\label{tab:diag-raw-similarity}
\label{tab:diag-context-normalized}
\centering
\appendixTableSetup
{\bfseries (a) Raw within-trajectory similarity}\par\vspace{0.18em}
\begin{tabular*}{\linewidth}{@{\extracolsep{\fill}}lrrrr@{}}
\toprule
\textbf{Example} & \textbf{Latent tokens} & \textbf{Off-diag. cos.} & \textbf{Adjacent cos.} & \textbf{Max cos.} \\
\midrule
VISCOT 18 & 126 & 0.7342 & 0.9417 & 0.9951 \\
VISCOT 23 & 63 & 0.8408 & 0.9215 & 0.9958 \\
\bottomrule
\end{tabular*}

\vspace{0.58em}
{\bfseries (b) Context-normalized similarity}\par\vspace{0.18em}
\begin{tabularx}{\linewidth}{@{}>{\raggedright\arraybackslash}Xrrrr@{}}
\toprule
\textbf{Comparison} & \textbf{LVR} & \textbf{Dummy} & \textbf{Ordinary} & \textbf{LVR--Ord.} \\
\midrule
Raw adjacent cosine & 0.8877 & 0.9923 & 0.4817 & +0.4059 \\
Centered adjacent cosine & 0.8166 & 0.9978 & 0.5210 & +0.2956 \\
Whitened adjacent cosine & 0.6074 & 0.6481 & 0.0952 & +0.5122 \\
Residualized LVR adjacent cosine & 0.7760 & -- & -- & -- \\
Same-task LVR mean cosine & 0.9960 & -- & -- & -- \\
Different-task LVR mean cosine & 0.9907 & -- & -- & -- \\
\bottomrule
\end{tabularx}
\end{table}

\begin{table}[!htbp]
\caption{\textbf{Representation stability and effective rank.}
(a) Representation consistency (RCS) is the mean pairwise cosine between
repeated mean LVR trajectories under each condition.
(b) Rank90, Rank95, and Rank99 count the principal directions needed to
explain 90\%, 95\%, and 99\% of the variance; PR is the participation ratio.}
\label{tab:diag-geometry}
\label{tab:diag-rcs}
\label{tab:diag-rank}
\centering
\appendixTableSetup
{\bfseries (a) Representation consistency under perturbations}\par\vspace{0.18em}
\begin{tabularx}{\linewidth}{@{}l>{\raggedright\arraybackslash}Xr@{}}
\toprule
\textbf{Family} & \textbf{Condition} & \textbf{Mean RCS} \\
\midrule
Reference & No augmentation & 0.99999997 \\
Image augmentation & Brightness/contrast/rotation/crop & 0.99917778 \\
Mask sweep & Strength 0.05 & 0.99987373 \\
Mask sweep & Strength 0.30 & 0.99974626 \\
Blur+jitter & Strength 0.10 & 0.99994875 \\
Blur+jitter & Strength 0.50 & 0.99988020 \\
Affine sweep & Tested affine range & 0.99917--0.99939 \\
\bottomrule
\end{tabularx}

\vspace{0.58em}
{\bfseries (b) Effective-rank diagnostics}\par\vspace{0.18em}
\begin{tabularx}{\linewidth}{@{}l>{\raggedright\arraybackslash}Xrrrr@{}}
\toprule
\textbf{Dataset} & \textbf{Representation} & \textbf{Shape} & \textbf{Rank90} & \shortstack{\textbf{Rank95 /}\\\textbf{Rank99}} & \textbf{PR} \\
\midrule
\multirow[t]{4}{*}{BLINK} & visual embedding & $163868{\times}3584$ & 428 & 713 / 1697 & 54.31 \\
& LVR latent tokens & $11152{\times}3584$ & 4 & 5 / 24 & 2.60 \\
& LVR mean & -- & 4 & 6 / 56 & 1.61 \\
& last LVR token & -- & 2 & 3 / 17 & 1.35 \\
\midrule
\multirow[t]{4}{*}{VISCOT} & visual embedding & $242899{\times}3584$ & 404 & 675 / 1641 & 58.55 \\
& LVR latent tokens & $81945{\times}3584$ & 51 & 169 / 909 & 5.70 \\
& LVR mean & -- & 21 & 65 / 289 & 4.70 \\
& last LVR token & -- & 36 & 111 / 414 & 5.66 \\
\bottomrule
\end{tabularx}
\end{table}

\FloatBarrier

\subsection{Target Alignment and Answer Readout}
\label{app:diag-target-readout}
Table~\ref{tab:diag-teacher-forcing} compares autoregressively generated
latent trajectories with the visual targets used during reconstruction
training. Each mixed schedule supplies the first $k$ target vectors and
lets the remaining latent states free-run. Supplying more targets increases
cosine similarity, while fully free-running states remain weakly aligned
with the targets. The fully forced trajectory matches the target by
construction; it does not measure learned alignment during inference.

\begin{table}[!htbp]
\caption{\textbf{Alignment with visual targets under target forcing.}
The first $k$ latent positions receive the visual target vectors; later
positions are generated autoregressively. MSE and cosine similarity compare
the resulting trajectory with the visual targets.}
\label{tab:diag-teacher-forcing}
\centering
\appendixTableSetup
\begin{tabular*}{\linewidth}{@{\extracolsep{\fill}}lrrrr@{}}
\toprule
& \multicolumn{2}{c}{\textbf{BLINK}} & \multicolumn{2}{c}{\textbf{VISCOT}} \\
\cmidrule(lr){2-3}\cmidrule(l){4-5}
\textbf{Generation schedule} & \textbf{MSE} & \textbf{Cosine} & \textbf{MSE} & \textbf{Cosine} \\
\midrule
Free-running & 1.4218 & 0.2145 & 1.7700 & 0.2863 \\
Force $k{=}4$ & -- & 0.4221 & -- & 0.4898 \\
Force $k{=}8$ & -- & 0.6145 & -- & 0.6619 \\
Force all $k{=}16$ & -- & 1.0000 & -- & 1.0000 \\
\bottomrule
\end{tabular*}
\end{table}

Table~\ref{tab:diag-credit-residual} summarizes answer readout and residual
injection. The attention measurements describe associations with answer
correctness; residual injection measures local sensitivity. Neither alone
assigns causal credit to a latent token.

\begin{table}[!htbp]
\caption{\textbf{Answer readout and residual-injection diagnostics.}
Summary statistics from generated rollouts. Attention deltas compare correct
and incorrect predictions.}
\label{tab:diag-credit-residual}
\centering
\appendixTableSetup
\begin{tabularx}{\linewidth}{@{}>{\raggedright\arraybackslash}p{0.25\linewidth}>{\raggedright\arraybackslash}p{0.21\linewidth}Y@{}}
\toprule
\textbf{Diagnostic} & \textbf{Evaluation} & \textbf{Reported values} \\
\midrule
Answer-to-latent-token attention & \texttt{MMVP} & Acc. $64.00\%$; answer-to-LVR mass $\Delta{=}+0.0099$; top-1 attention $\Delta{=}+0.0307$. \\
& \texttt{BLINK} & Acc. $49.28\%$; mass $\Delta{=}+0.0029$; top-1 attention $\Delta{=}+0.0217$. \\
& \texttt{HR-4K} / \texttt{HR-8K} & Acc. $56.88\%$ / $48.62\%$; correct-answer mass and top-1 mass are higher, with smaller deltas at 8K. \\
\midrule
Residual injection & $47{,}012$ generated rollout records & Best $\Delta$ toward counterfactual prediction: $+0.0068$ to $+0.0137$; logit-margin deltas up to $+0.0358$. \\
\bottomrule
\end{tabularx}
\end{table}

\FloatBarrier

\subsection{Task Structure and Counterfactual Sensitivity}
\label{app:diag-counterfactual}
Table~\ref{tab:diag-task-structure}(a) tests whether representations encode
the BLINK task label. Adjusted Rand index (ARI) measures cluster alignment
with task labels; probe accuracy uses 5-fold cross-validation. 

\begin{table}[!htbp]
\caption{\textbf{Task decodability and sensitivity to answer-changing edits.}
(a) Task-label structure on BLINK ($N{=}697$).
(b) Sensitivity to synthetic image edits ($N{=}2{,}048$; 512 pairs per edit).
LVR distance uses the mean-pooled latent trajectory; final distance uses the
final hidden representation.}
\label{tab:diag-task-counterfactual}
\label{tab:diag-task-structure}
\label{tab:diag-counterfactual}
\centering
\appendixTableSetup
{\bfseries (a) Task-label decodability}\par\vspace{0.18em}
\begin{tabular*}{\linewidth}{@{\extracolsep{\fill}}lrrrr@{}}
\toprule
\textbf{Representation} & \textbf{ARI} & \shortstack{\textbf{Bootstrap}\\\textbf{ARI}} & \shortstack{\textbf{Linear}\\\textbf{accuracy}} & \shortstack{\textbf{MLP}\\\textbf{accuracy}} \\
\midrule
$h_{\mathrm{pre}}$ & 0.9489 & 0.9492 & 0.9994 & 0.9960 \\
LVR mean & 0.3885 & 0.3905 & 0.9991 & 0.9822 \\
$h_{\mathrm{lvr,last}}$ & 0.1298 & 0.1430 & 0.9966 & 0.9684 \\
Final hidden & 0.2714 & 0.2558 & 0.9954 & 0.9641 \\
\bottomrule
\end{tabular*}

\vspace{0.58em}
{\bfseries (b) Counterfactual sensitivity}\par\vspace{0.18em}
\begin{tabular*}{\linewidth}{@{\extracolsep{\fill}}lrrrrr@{}}
\toprule
\textbf{Edit} & \shortstack{\textbf{Should}\\\textbf{flip}} & \shortstack{\textbf{Model}\\\textbf{flip}} & \shortstack{\textbf{Correct}\\\textbf{flip}} & \shortstack{\textbf{LVR}\\\textbf{distance}} & \shortstack{\textbf{Final}\\\textbf{distance}} \\
\midrule
Color change & 86.33\% & 5.66\% & 5.86\% & 0.000158 & 0.004764 \\
Object removal & 81.45\% & 13.09\% & 6.45\% & 0.000445 & 0.022313 \\
Shape swap & 84.57\% & 12.70\% & 5.27\% & 0.000253 & 0.012438 \\
Spatial swap & 84.96\% & 11.33\% & 4.69\% & 0.000162 & 0.008934 \\
\bottomrule
\end{tabular*}
\end{table}

Panel (b) edits the image while holding the question fixed. ``Should flip''
is the ground-truth answer-change rate, and ``Model flip'' is the prediction
change rate. The separately reported ``Correct flip'' rate counts predictions
that reach the edited ground-truth answer. Both distance columns report
cosine distance. The correct answer changes for most pairs, but LVR's answers
change infrequently and its mean-pooled trajectories move only slightly.

\FloatBarrier

\FloatBarrier

\section{Additional Qualitative Results}
\label{app:qualitative-cases}
The visualizations below examine individual generated answers and illustrate
the visual reasoning tasks covered by our benchmarks.
Appendix~\ref{app:diagnostics} reports the dataset-level diagnostics.

\subsection{Image Saliency and Token Views}
\label{app:qualitative-original-views}
Each model first generates an answer, after which we compute
$|\mathrm{attention}\times\mathrm{gradient}|$ saliency for that sequence.
Figure~\ref{fig:token-saliency-cases} in the main text shows three paired
image and token views. The image overlays use answer-conditioned
$|\mathrm{attention}\times\mathrm{gradient}|$ saliency. The additional cases
below show explicitly labeled attention and saliency panels.

\FloatBarrier

\subsection{Attention and Saliency Cases}
\label{app:qualitative-attention-saliency}
\label{app:qualitative-answer-comparison}
\label{app:qualitative-additional-cases}
\begin{figure}[!htbp]
  \centering
  \setlength{\parskip}{0pt}
  \includegraphics[width=\linewidth]{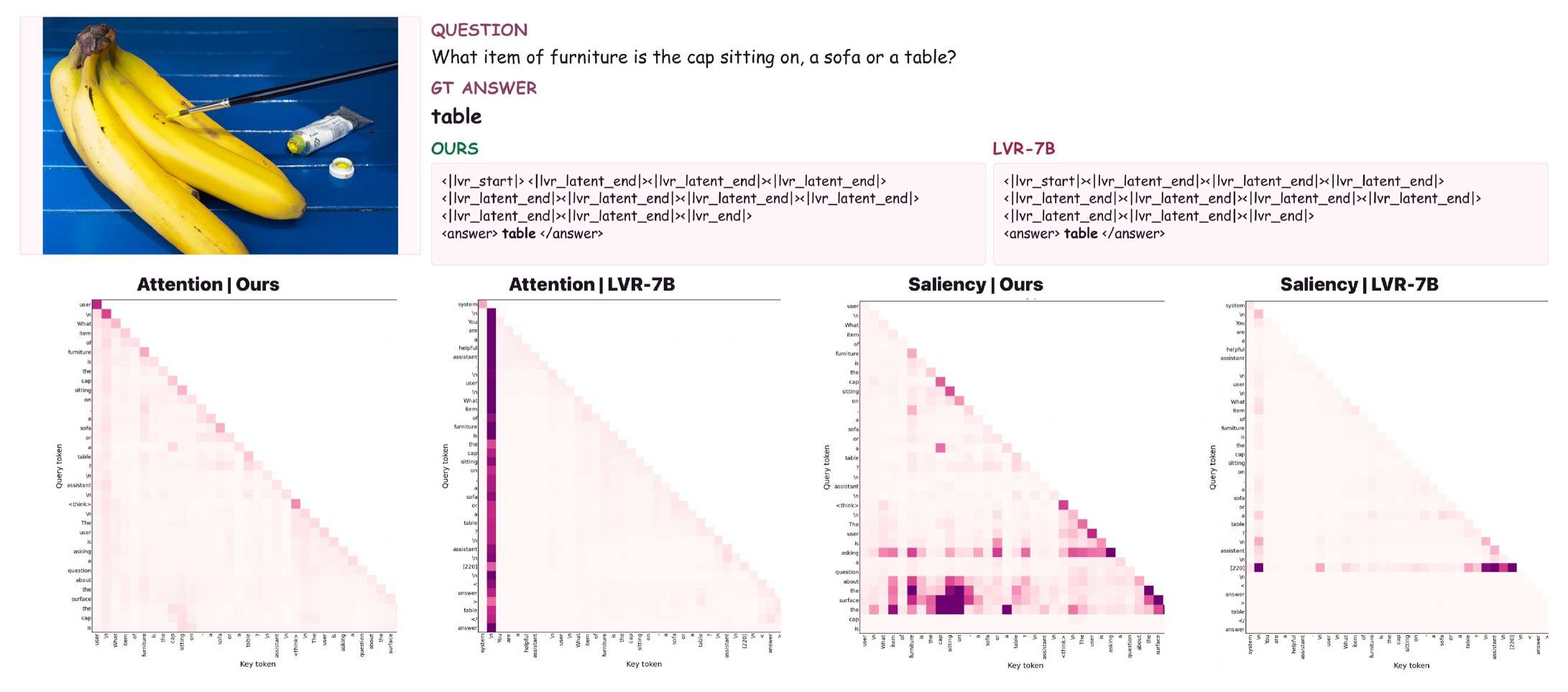}
  \par\vspace{4pt}
  \includegraphics[width=\linewidth]{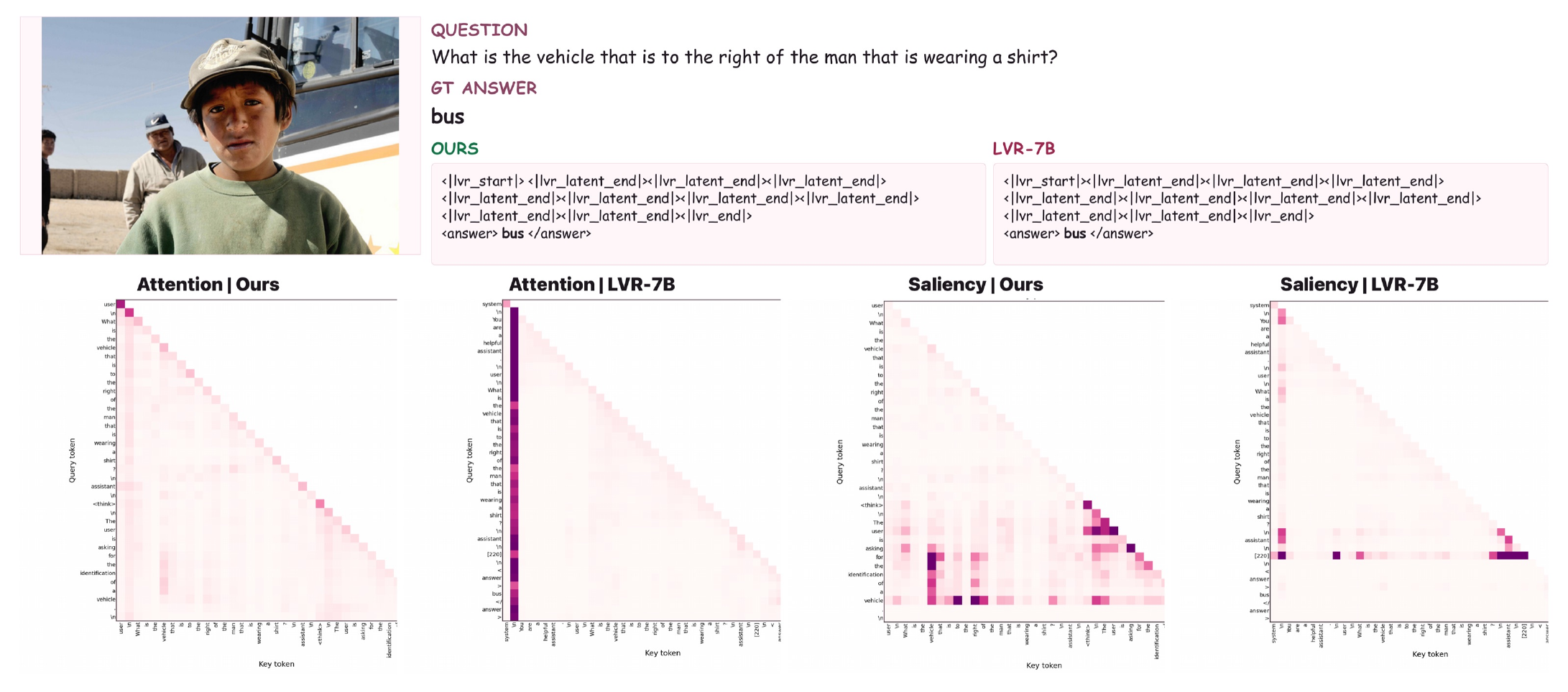}
  \par\vspace{4pt}
  \includegraphics[width=\linewidth]{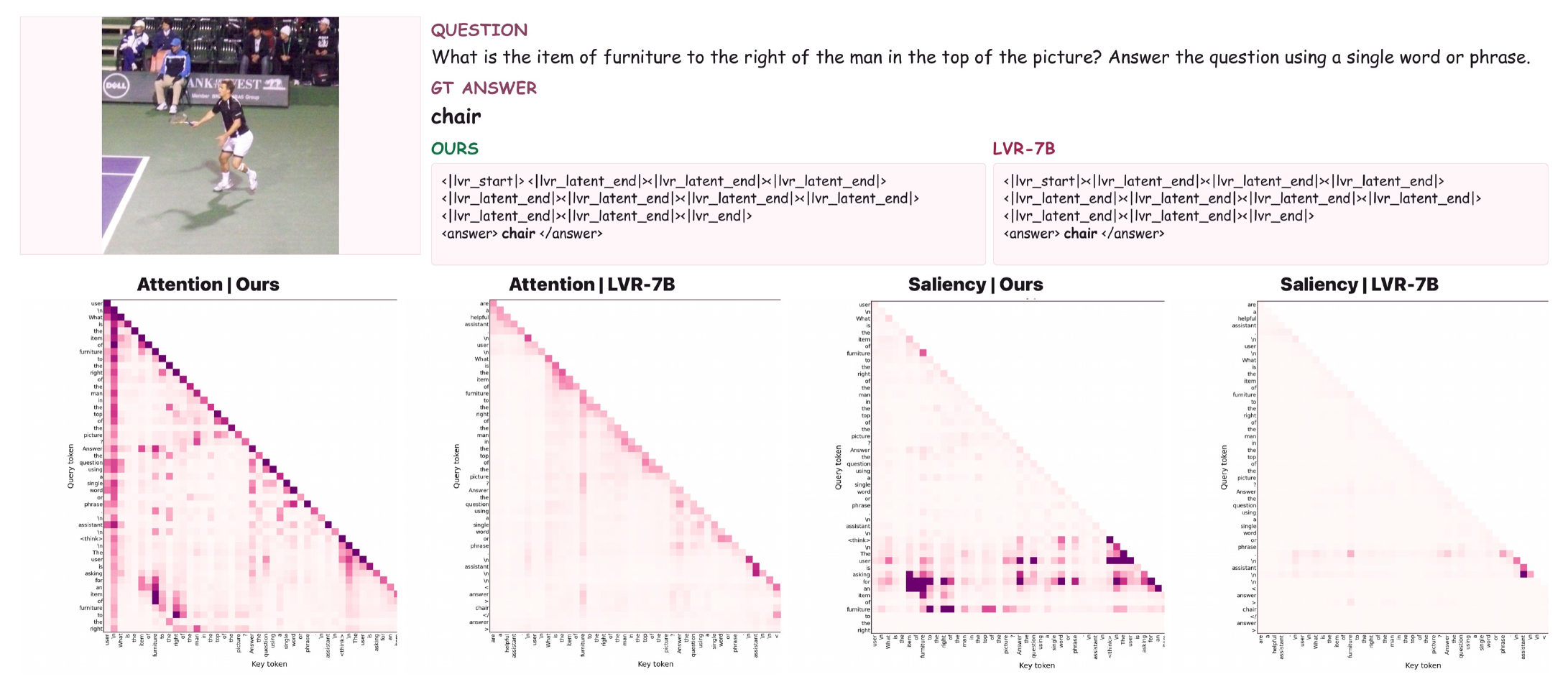}
  \caption{\textbf{Attention and saliency cases (01--03).}
  Top to bottom: the cap on a table, the bus beside a person, and the
  chair beside a tennis player.}
  \label{fig:appendix-token-saliency-01-03}
\end{figure}

\begin{figure}[!htbp]
  \centering
  \setlength{\parskip}{0pt}
  \includegraphics[width=\linewidth]{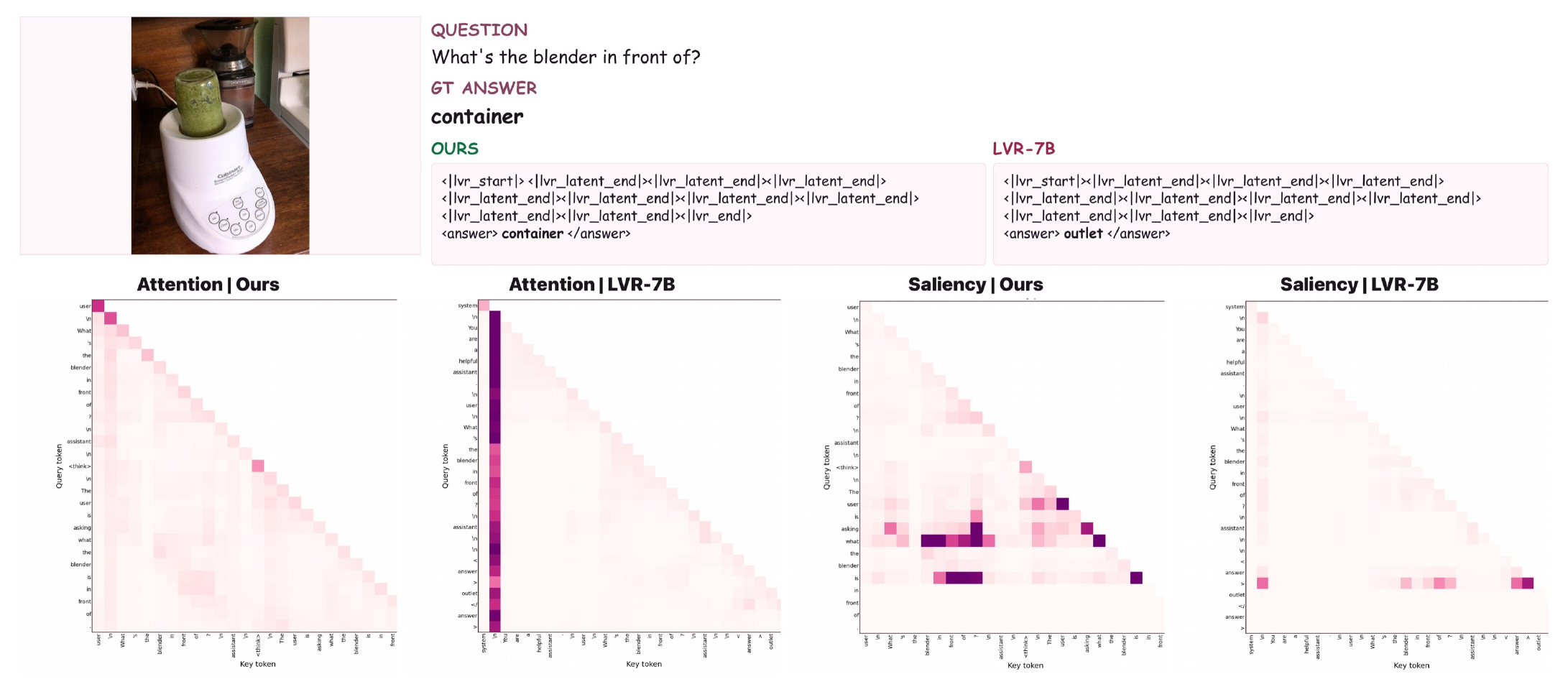}
  \par\vspace{4pt}
  \includegraphics[width=\linewidth]{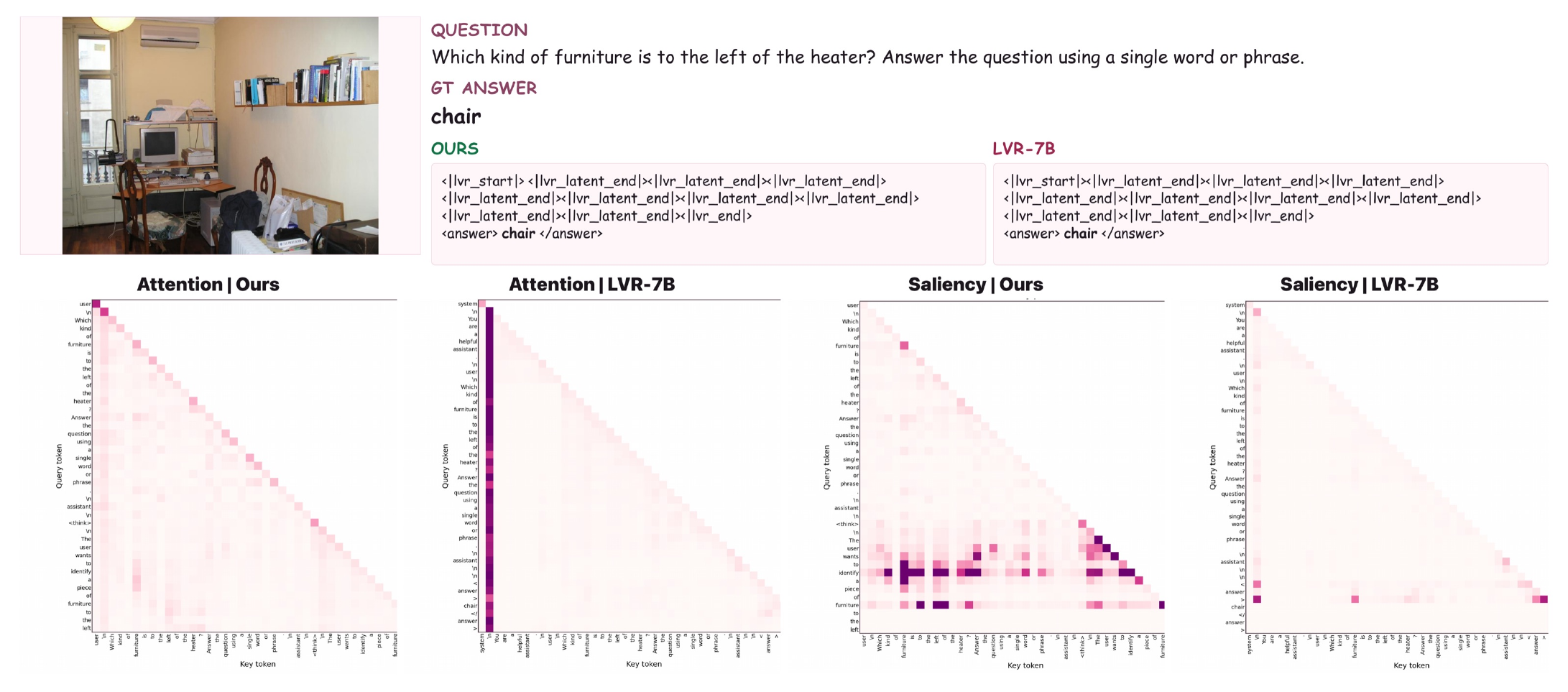}
  \par\vspace{4pt}
  \includegraphics[width=\linewidth]{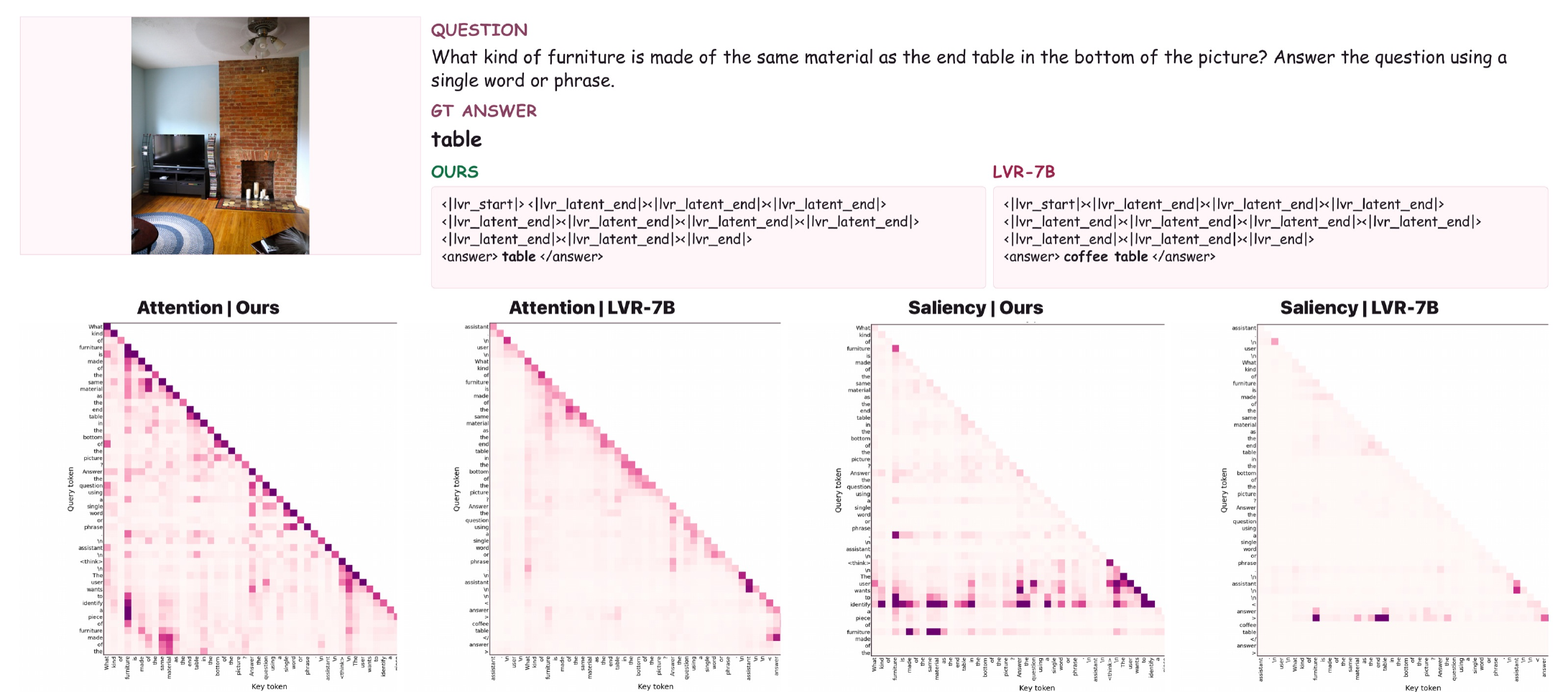}
  \caption{\textbf{Attention and saliency cases (04--06).}
  Top to bottom: the object behind the blender, the furniture left of the
  heater, and a furniture-material comparison.}
  \label{fig:appendix-token-saliency-04-06}
  \label{fig:appendix-token-saliency-04-07}
\end{figure}

\begin{figure}[!htbp]
  \centering
  \setlength{\parskip}{0pt}
  \includegraphics[width=\linewidth]{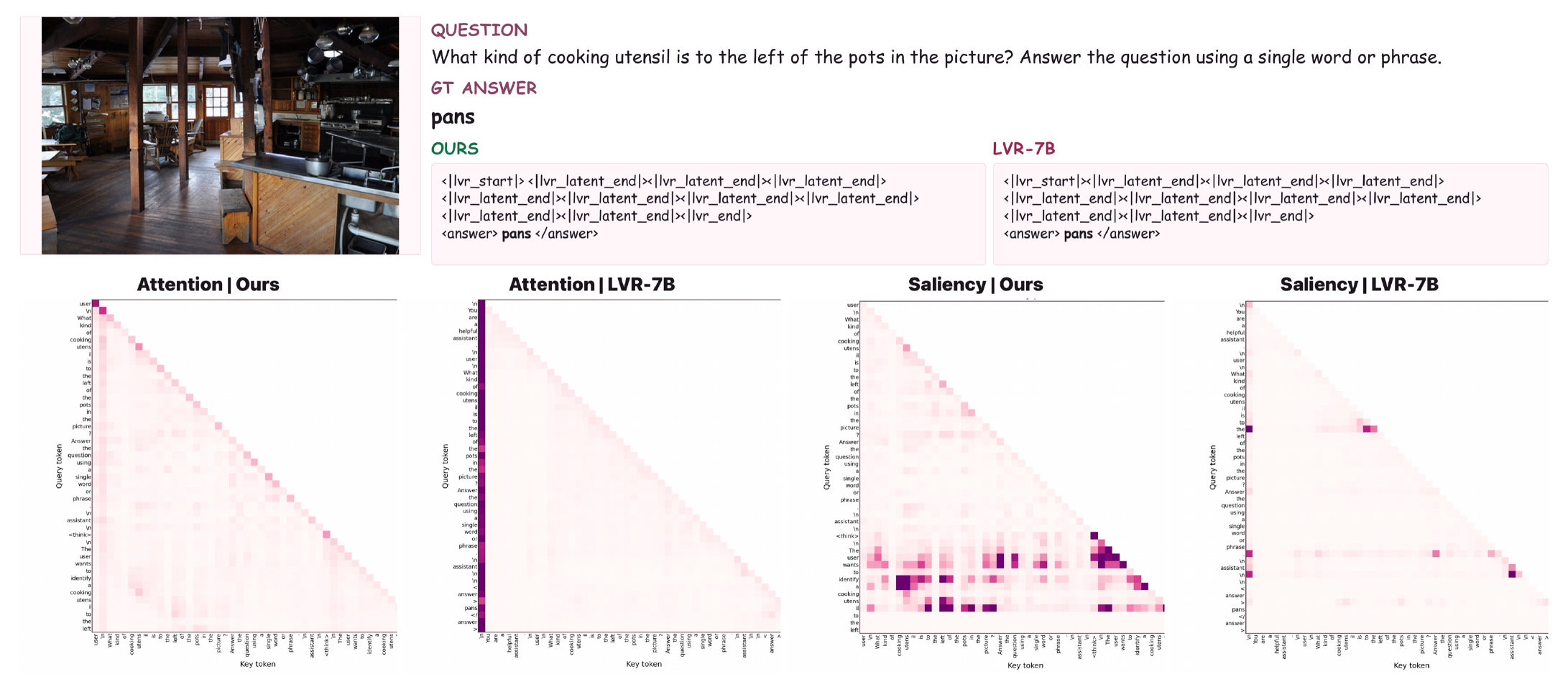}
  \par\vspace{4pt}
  \includegraphics[width=\linewidth]{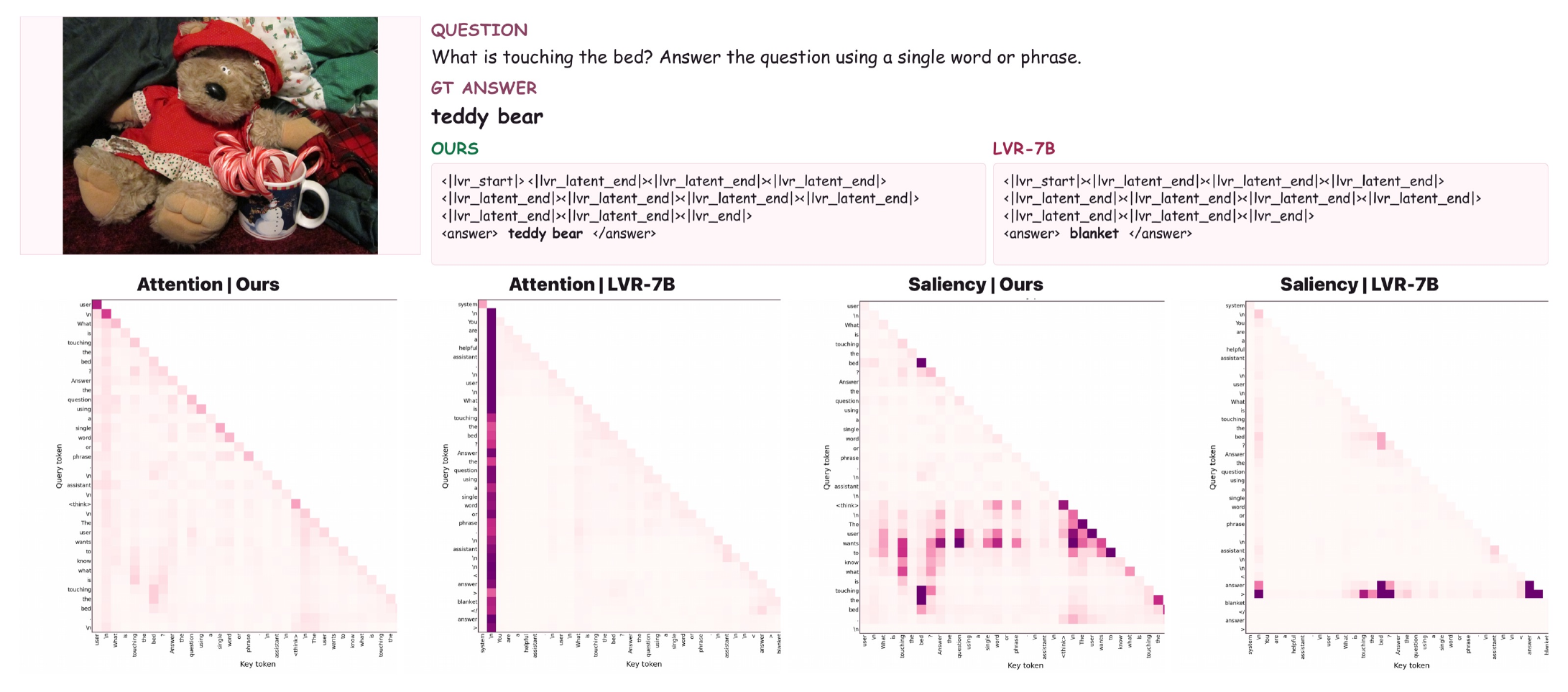}
  \par\vspace{4pt}
  \includegraphics[width=\linewidth]{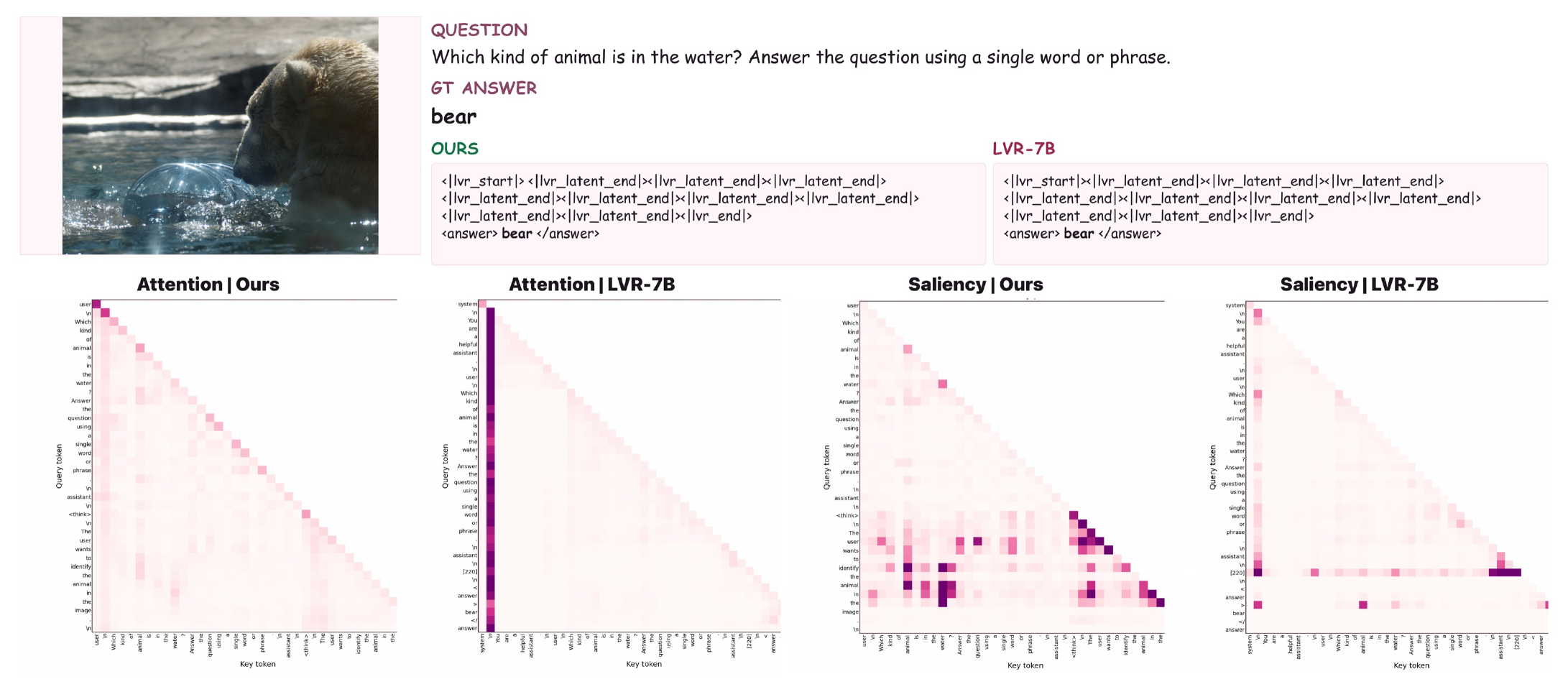}
  \caption{\textbf{Attention and saliency cases (07--09).}
  Top to bottom: pans left of the pots, the teddy-bear answer comparison,
  and a bear in water. In the middle example (Case 08), ReaLVR answers
  ``teddy bear'' and LVR-7B answers ``blanket'' to ``What is touching the bed?''}
  \label{fig:appendix-token-saliency-07-09}
  \label{fig:token-saliency-full}
  \label{fig:appendix-token-saliency-07-09-10}
  \label{fig:appendix-token-saliency-09-12}
\end{figure}

\begin{figure}[!htbp]
  \centering
  \setlength{\parskip}{0pt}
  \includegraphics[width=\linewidth]{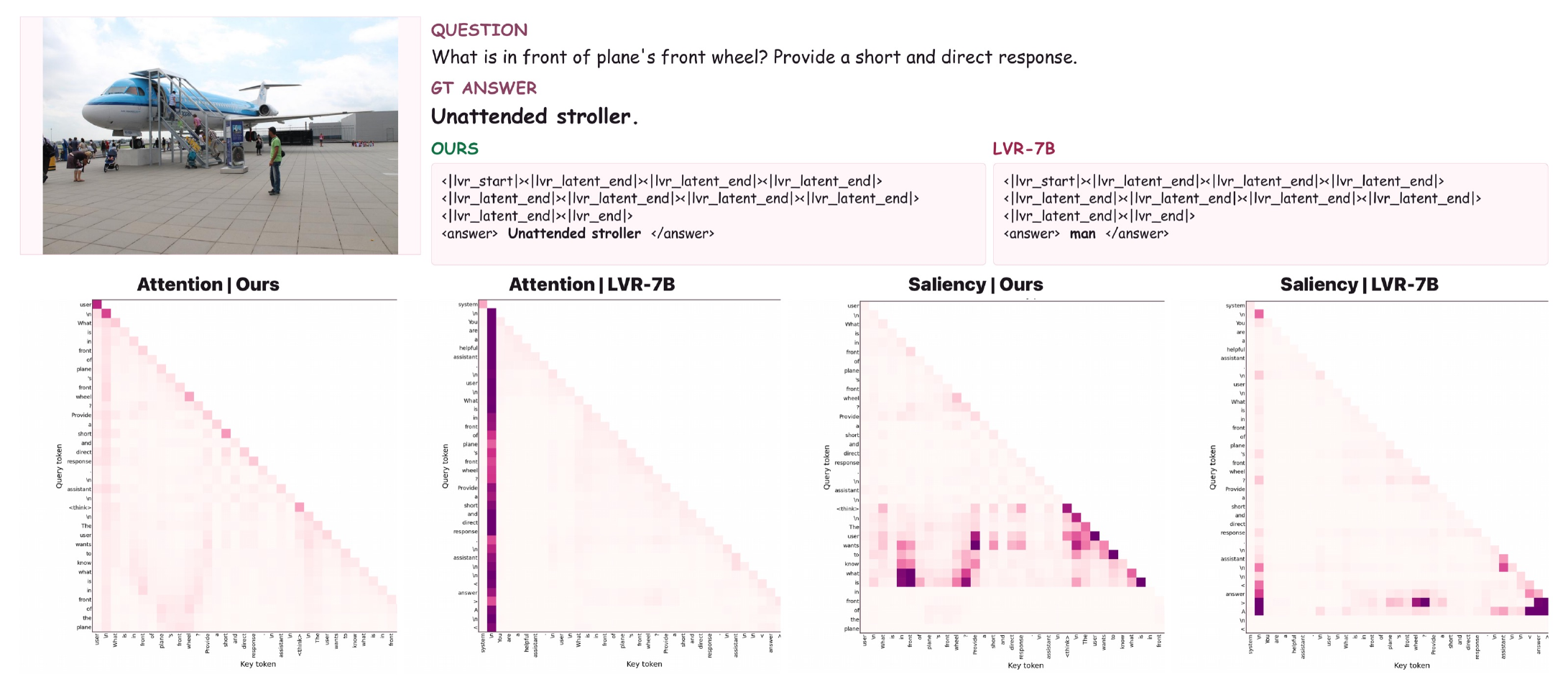}
  \par\vspace{4pt}
  \includegraphics[width=\linewidth]{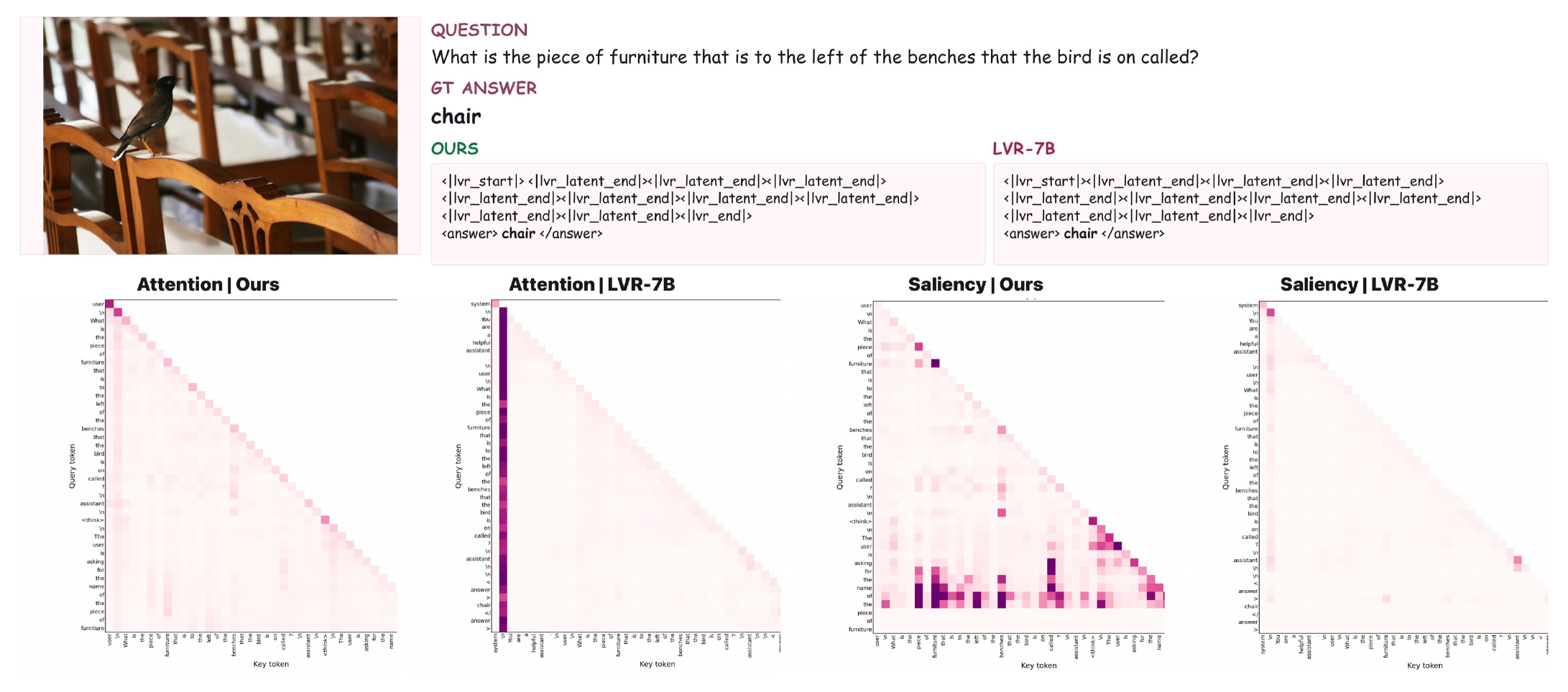}
  \par\vspace{4pt}
  \includegraphics[width=\linewidth]{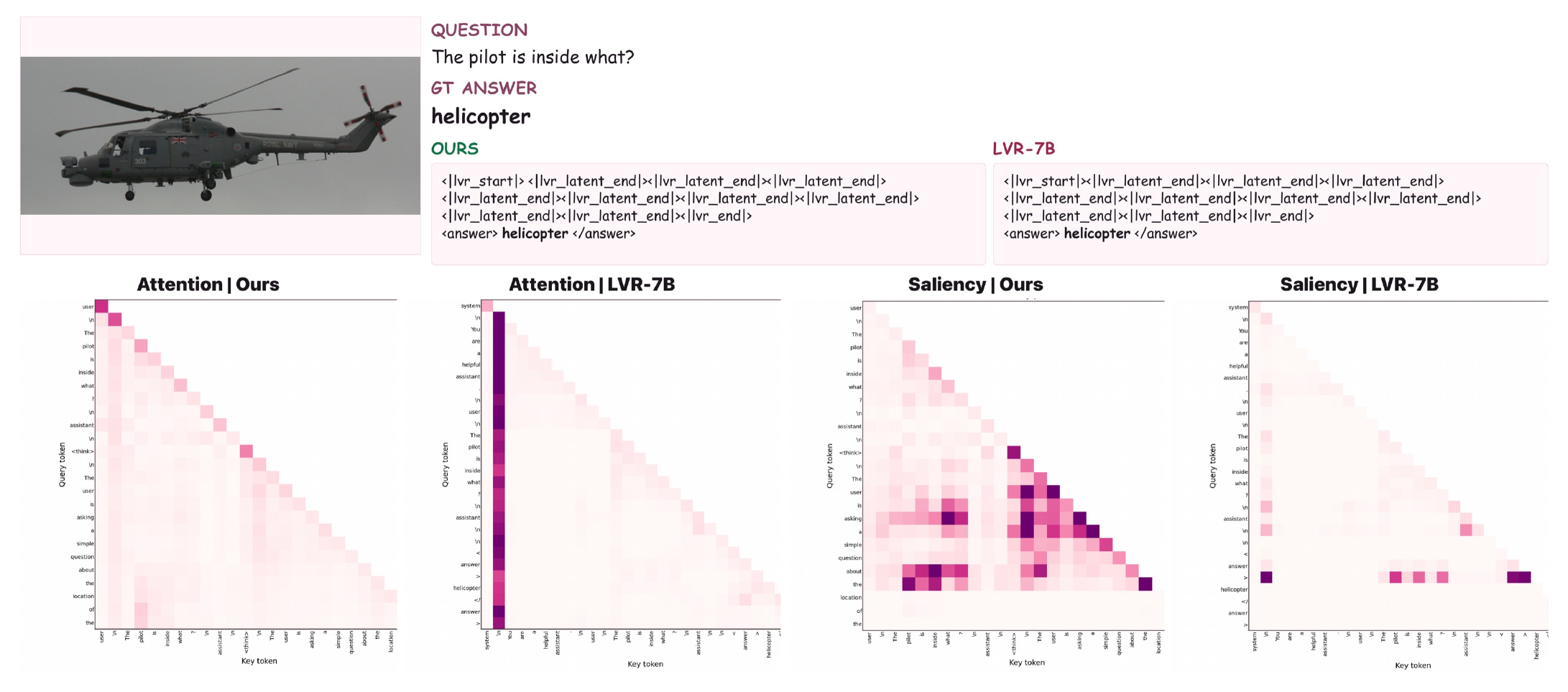}
  \caption{\textbf{Attention and saliency cases (10--12).}
  Top to bottom: the object in front of an airplane's front wheel,
  the chair-and-bird scene, and the vehicle containing the pilot.}
  \label{fig:appendix-token-saliency-10-12}
  \label{fig:appendix-token-saliency-11-12}
\end{figure}

Figures~\ref{fig:appendix-token-saliency-01-03}--\ref{fig:appendix-token-saliency-10-12}
compare twelve image--question cases at a common scale. Each includes the
answers and four labeled attention/saliency panels, using the token axes and
attribution protocol in Appendix~\ref{app:diag-saliency}.

\FloatBarrier
\subsection{Task Examples Across Five Benchmarks}
\label{app:benchmark-case-studies}
\begin{figure}[!htbp]
  \centering
  \includegraphics[width=\linewidth]{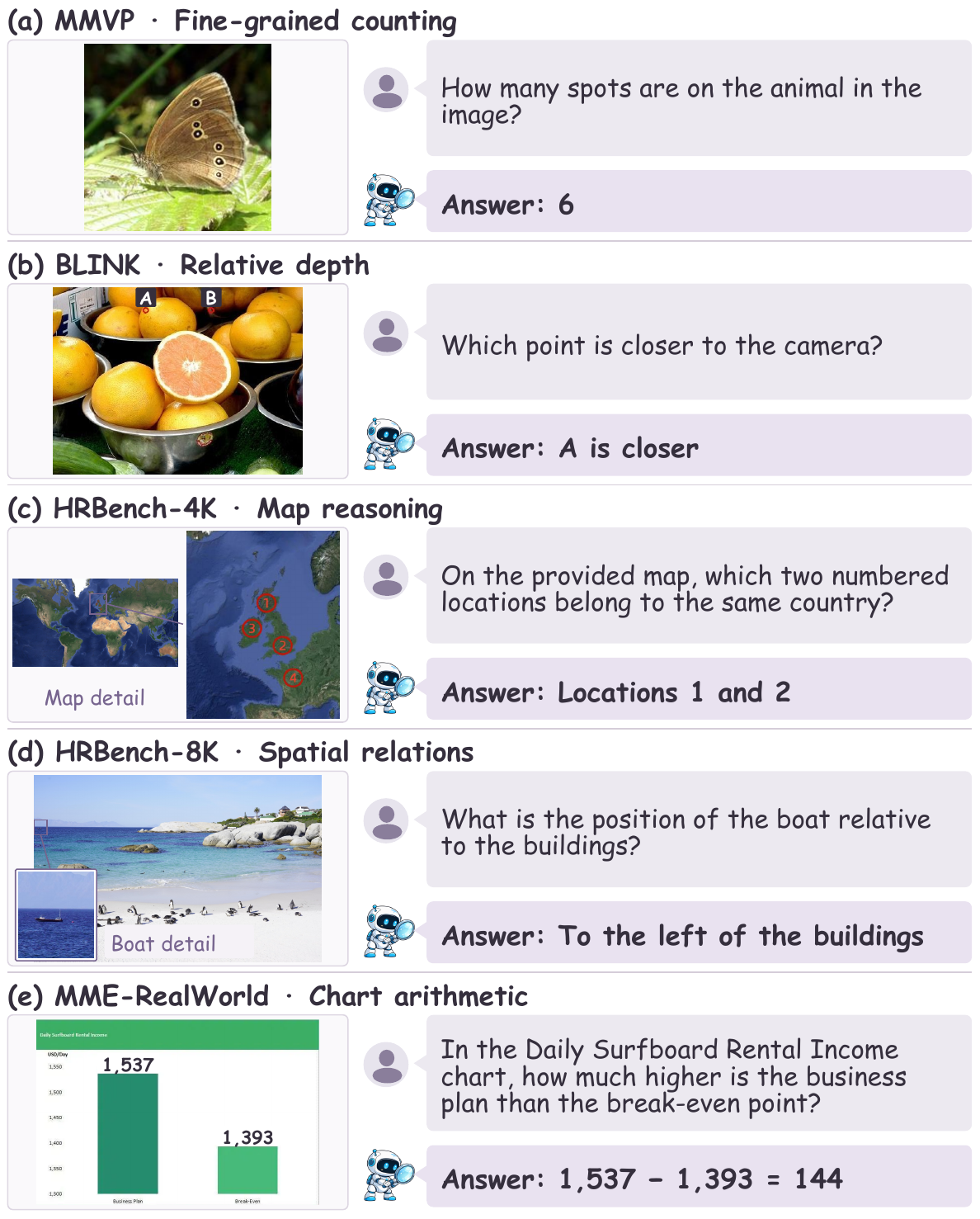}
  \caption{\textbf{One question from each of the five benchmarks.}
  Images, questions, and answers come from the official
  datasets~\citep{tong2024eyes,fu2024blink,wang2025hrbench,zhang2025mmerealworld}.
  Robot bubbles show annotated answers, not recorded ReaLVR predictions;
  the chart calculation is added for clarity.
  Insets enlarge source-image details, and the chart is cropped to the
  relevant panel. Point and value labels are enlarged for readability.
  Multiple-choice options are omitted, and the chart question is shortened.}
  \label{fig:five-benchmark-cases}
\end{figure}

Figure~\ref{fig:five-benchmark-cases} illustrates the evidence required by
each benchmark. Beyond counting and depth comparison, the map question
relates numbered locations to countries, and the 8K scene requires finding
a small boat before judging its position relative to distant buildings.
The chart question combines reading numerical values with subtraction:
$1{,}537-1{,}393=144$.

\FloatBarrier
\subsection{Position-wise Latent Variation}
\begin{figure}[!htb]
  \centering
  \includegraphics[width=\linewidth]{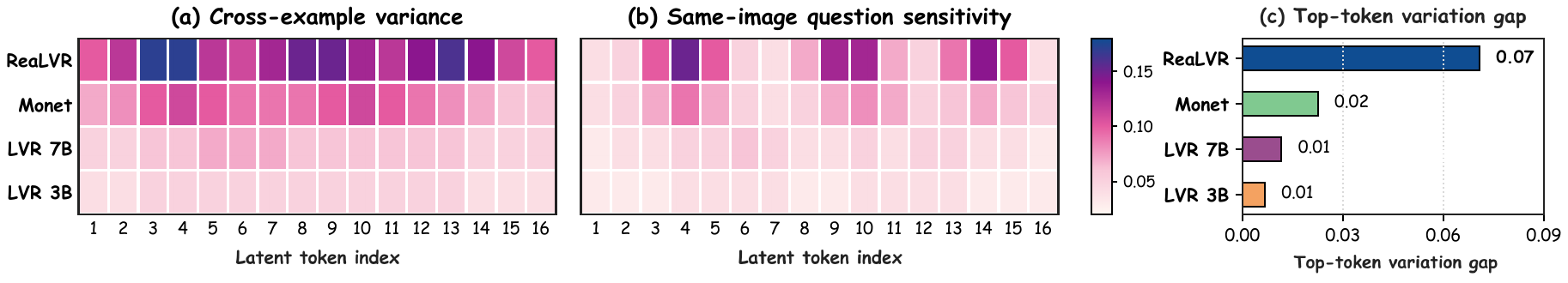}
  \caption{\textbf{Position-wise latent variation.}
  (a) Cross-example variation. (b) Variation across questions about the same
  image. Heatmap columns index latent positions $1$--$16$.
  (c) Reported top-token variation gap: $0.07$ for ReaLVR, $0.02$ for Monet,
  and $0.01$ for each LVR variant. ReaLVR concentrates more of the measured
  variation at particular latent positions.}
  \label{fig:slot-variance-map}
\end{figure}

\clearpage
\section{Attention Analysis Designs}
\label{app:attention-designs}
The following designs illustrate three complementary questions: where latent
states read visual evidence, how evidence selection changes with the question,
and which latent states support the answer. They complement the diagnostic
definitions in Appendices~\ref{app:diag-target-region}
and~\ref{app:diag-latent-dependence}.

\subsection{Visual Evidence and Answer Readout}
\label{app:design-evidence-readout}
\begin{figure}[!htbp]
  \centering
  \includegraphics[width=\linewidth]{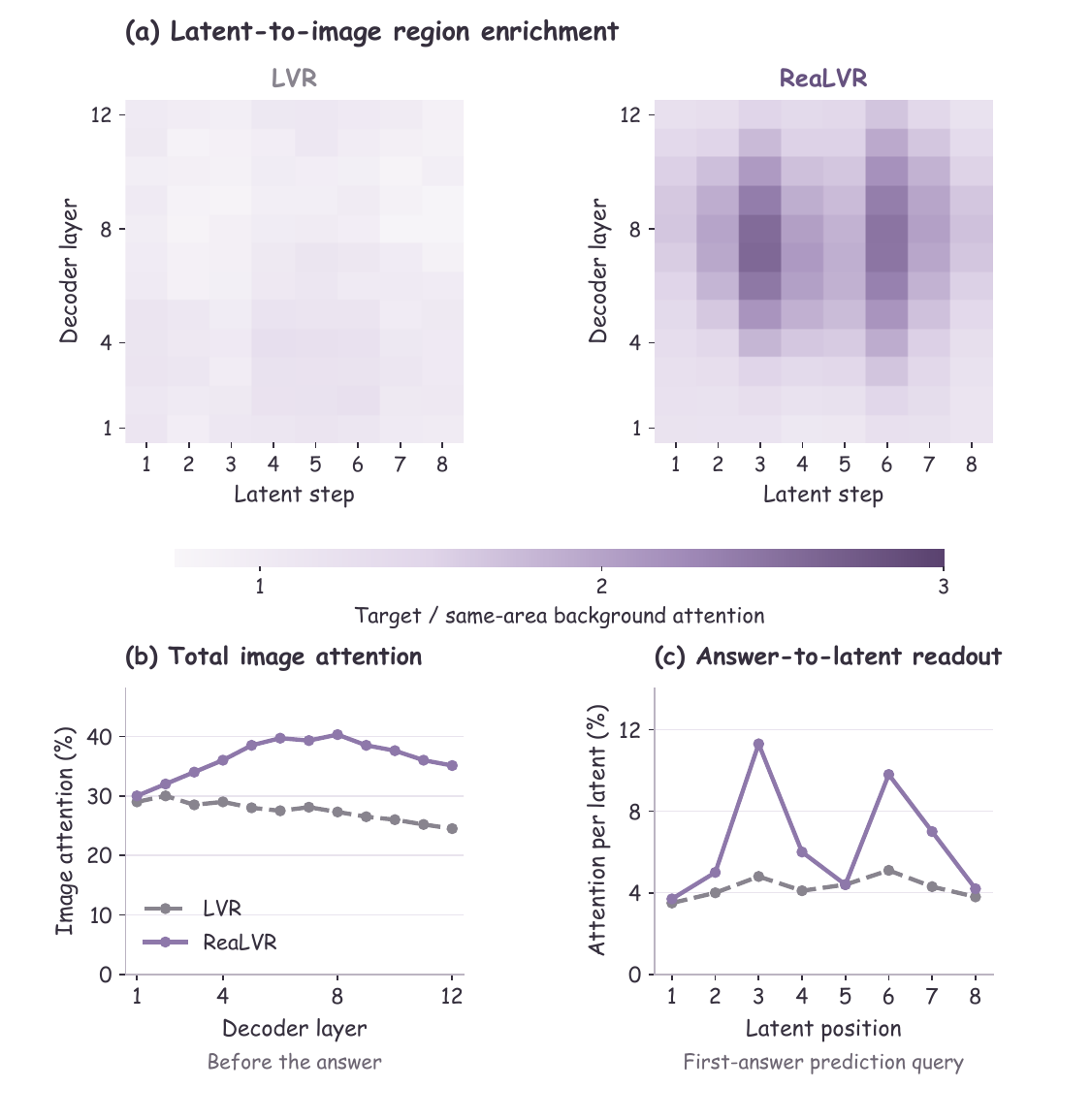}
  \caption{\textbf{Visual evidence and answer readout.}
  (a) Target-region attention relative to same-area background windows,
  shown by decoder layer and latent step; $1$ means no preference.
  Both heatmaps share a color scale.
  (b) Raw attention mass assigned to all image tokens before the answer.
  (c) Raw attention to each latent state from the query used to predict the
  first answer token, before that token is supplied.
  The panels illustrate aggregate post-softmax quantities.}
  \label{fig:design-evidence-readout}
\end{figure}

\clearpage
\subsection{Question-Dependent Evidence Selection}
\label{app:design-question-switch}
\begin{figure}[!htbp]
  \centering
  \includegraphics[width=\linewidth]{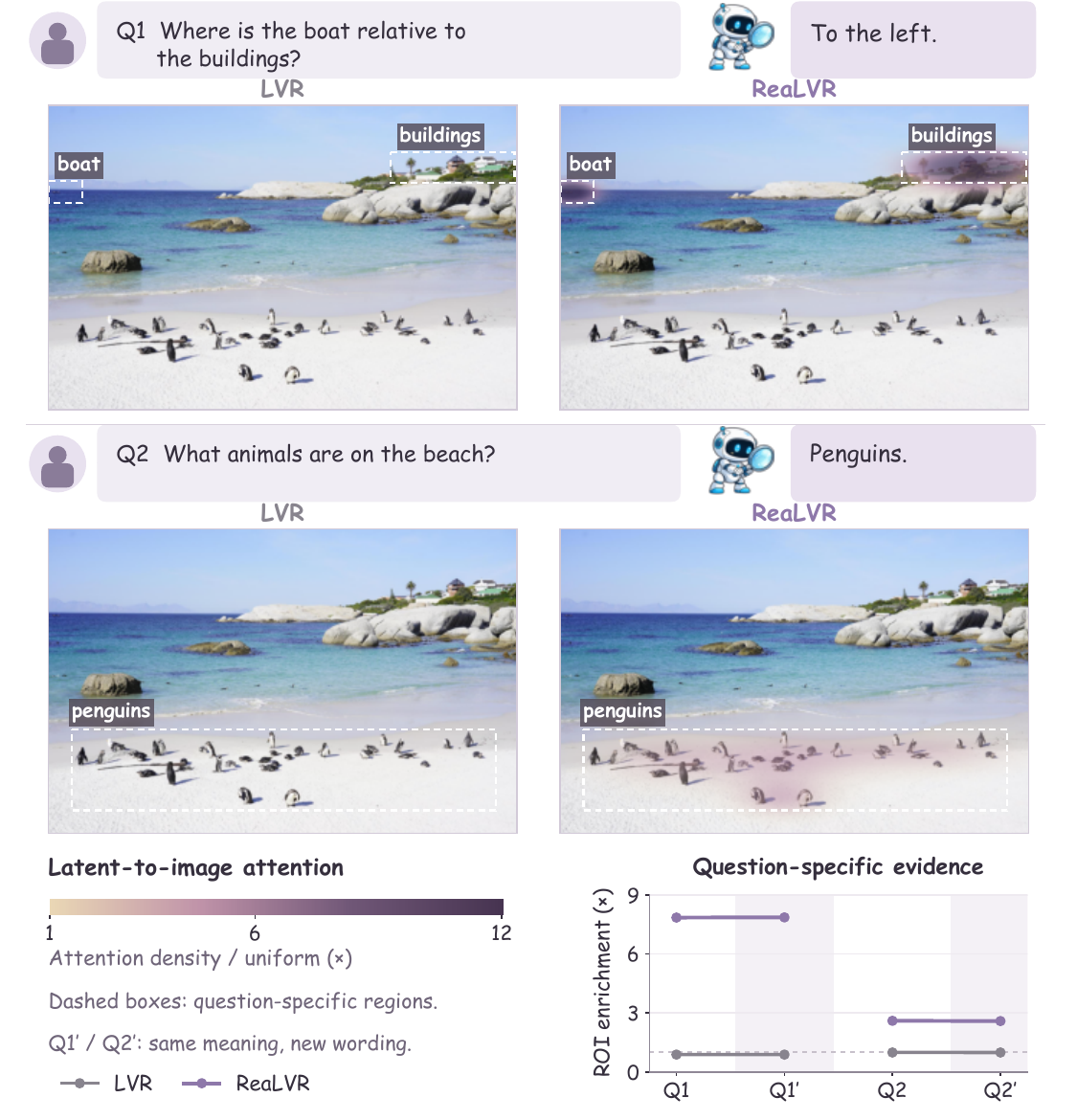}
  \caption{\textbf{Question-dependent evidence selection.}
  The photograph is HRBench-8K example~798~\citep{wang2025hrbench}.
  Q1 adapts its spatial-relation question; Q2 and both paraphrases are
  author-written. Dashed boxes mark the manually specified regions for
  the boat-and-buildings question and the animal question.
  All maps share a density scale normalized to a full-image mean of $1$.
  The chart reports mean density within each question's region and compares
  that question with its paraphrase ($Q1'$ or $Q2'$).
  This uniform baseline differs from the matched-background baseline in
  Figure~\ref{fig:design-evidence-readout}.}
  \label{fig:design-question-switch}
\end{figure}

\clearpage
\subsection{Latent Selection and Fixed-Context Intervention}
\label{app:design-latent-intervention}
\begin{figure}[!htbp]
  \centering
  \includegraphics[width=0.98\linewidth]{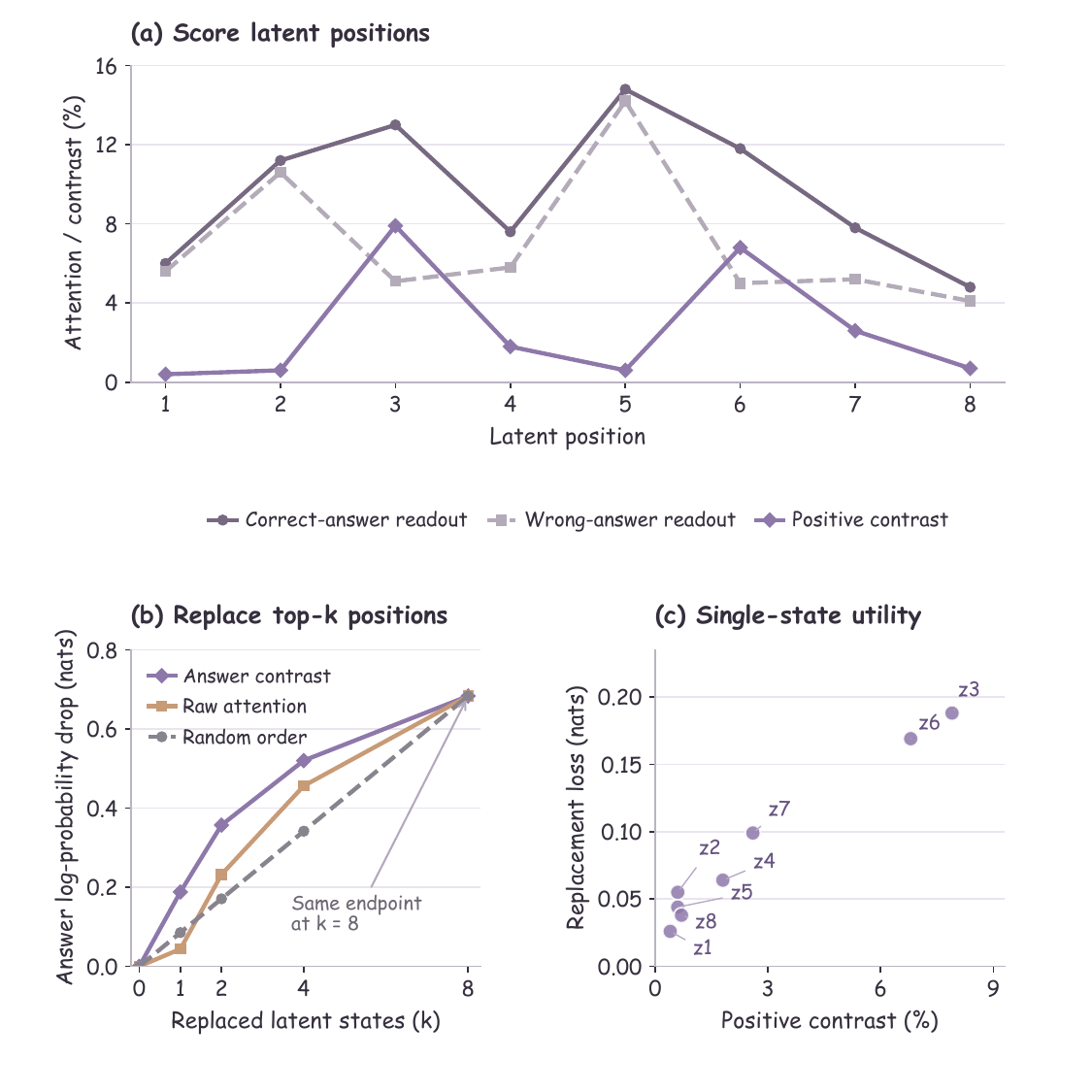}
  \caption{\textbf{Latent selection and fixed-context intervention.}
  (a) Teacher-forced readout after supplying correct or incorrect answers;
  positive contrast is the unnormalized positive part of their difference.
  (b) Correct-answer log-probability loss after replacing positions selected
  by contrast, raw correct-answer attention, or random ordering, with the
  other latent inputs fixed. (c) Loss from replacing each single latent position, plotted against that
position's positive contrast. Results are averaged over 1000 [BLINK] examples
on \texttt{Qwen2.5-VL-7B}. All strategies replace all eight positions at
$k=8$ and therefore coincide at the endpoint. Losses are in nats.}
  \label{fig:design-latent-intervention}
\end{figure}

\section{Paired Counterfactual and Mass-Matched Mechanism Tests}
\label{app:mechanism-tests}

This section reports a paired counterfactual comparison,
fixed-context latent exchange and a mass-matched test of the proposed
evidence pathway.

\subsection{Paired counterfactual behavior}
\label{app:complete-counterfactual}

For edit type $e$, let $(x_i^0,x_i^1)$ be the original and edited images
with the same question, and let $(y_i^0,y_i^1)$ be their canonical
ground-truth answers. We partition the pairs into
$\mathcal C_e=\{i:y_i^0\ne y_i^1\}$ and
$\mathcal U_e=\{i:y_i^0=y_i^1\}$. The four edit types are color change,
object removal, shape swap, and spatial swap; each has 512 pairs before
this partition. We evaluate Direct, LVR-RL, and ReaLVR on the identical
pairs with the same prompt, image processing, answer parser, and decoding
settings. Direct uses the same pretrained backbone, Stage~2 examples,
answer reward, and update budget but omits the latent span.

For $\mathcal C_e$, we report accuracy on each view, the fraction with
two parseable but different predictions, and the \emph{strict paired
correct flip},
\begin{equation}
  F_{\mathrm{correct}}(e)
  =\frac{1}{|\mathcal C_e|}
   \sum_{i\in\mathcal C_e}
   \mathbf{1}[\widehat y_i^0=y_i^0\ \land
                 \widehat y_i^1=y_i^1].
  \label{eq:strict-paired-flip}
\end{equation}
Because $y_i^0\ne y_i^1$, this event requires a correct answer change.
For $\mathcal U_e$, the false-flip rate counts parseable predictions that
change although the answer does not. Parse failures count as incorrect
and not as valid prediction changes. Accuracy, prediction change, and
strict correct flip use $|\mathcal C_e|$ as their denominator; false flip
uses $|\mathcal U_e|$. Parse coverage is the fraction of all $2\times512$
view-level predictions with a parseable answer.

\paragraph{Protocol.} Each pair is evaluated as a two-turn conversation. The first turn presents $x_i^0$ with question $q_i$; the second turn presents the edited image $x_i^1$ with the same $q_i$, keeping the first-turn exchange in context. This tests whether a model revises its answer when the visual evidence changes within a dialogue, rather than repeating its earlier prediction. \texttt{MMVP} instead scores each image in a separate conversation, where no earlier answer is available to anchor the prediction.

\begin{table*}[t]
\caption{\textbf{Complete paired counterfactual evaluation.}
The first four rate columns use changed-answer pairs $\mathcal C_e$;
false flip uses unchanged-answer pairs $\mathcal U_e$. Each edit type is
evaluated on the same pairs for all methods. Rates are percentages
computed from single deterministic greedy rollouts on the partitioned sets
$\mathcal C_e$ and $\mathcal U_e$; counts are in parentheses (parse
coverage counts view-level predictions).}
\label{tab:complete-counterfactual}
\centering
\appendixWideTableSetup
\begin{adjustbox}{max width=\linewidth}
\begin{tabular}{llccccccc}
\toprule
\textbf{Edit} & \textbf{Method} & $|\mathcal C_e|/|\mathcal U_e|$ &
\shortstack{\textbf{Original}\\\textbf{acc.} ($A_0$)} & \shortstack{\textbf{Edited}\\\textbf{acc.} ($A_1$)} &
\shortstack{\textbf{Prediction}\\\textbf{change} ($\Delta$)} &
\shortstack{\textbf{Strict paired}\\\textbf{correct flip}\\($F_{\mathrm{correct}}$)} &
\shortstack{\textbf{False flip}\\\textbf{on} $\mathcal U_e$} &
\shortstack{\textbf{Parse}\\\textbf{coverage}} \\
\midrule
Color change & Direct & 442 / 70 & \shortstack{75.8\\(335)} & \shortstack{70.4\\(311)} & \shortstack{61.3\\(271)} & \shortstack{53.6\\(237)} & \shortstack{11.4\\(8)} & \shortstack{99.8\\(1022)} \\
 & LVR-RL & 442 / 70 & \shortstack{78.5\\(347)} & \shortstack{8.1\\(36)} & \shortstack{6.1\\(27)} & \shortstack{4.8\\(21)} & \shortstack{2.9\\(2)} & \shortstack{99.8\\(1022)} \\
 & \textbf{ReaLVR} & 442 / 70 & \shortstack{\textbf{85.3}\\(377)} & \shortstack{\textbf{81.2}\\(359)} & \shortstack{\textbf{80.5}\\(356)} & \shortstack{\textbf{73.1}\\(323)} & \shortstack{\textbf{5.7}\\(4)} & \shortstack{\textbf{100.0}\\(1024)} \\
\midrule
Object removal & Direct & 417 / 95 & \shortstack{74.3\\(310)} & \shortstack{68.6\\(286)} & \shortstack{63.8\\(266)} & \shortstack{52.3\\(218)} & \shortstack{12.6\\(12)} & \shortstack{99.5\\(1019)} \\
 & LVR-RL & 417 / 95 & \shortstack{77.2\\(322)} & \shortstack{11.3\\(47)} & \shortstack{14.6\\(61)} & \shortstack{6.7\\(28)} & \shortstack{6.3\\(6)} & \shortstack{99.8\\(1022)} \\
 & \textbf{ReaLVR} & 417 / 95 & \shortstack{\textbf{84.4}\\(352)} & \shortstack{\textbf{79.1}\\(330)} & \shortstack{\textbf{77.5}\\(323)} & \shortstack{\textbf{69.8}\\(291)} & \shortstack{\textbf{6.3}\\(6)} & \shortstack{\textbf{100.0}\\(1024)} \\
\midrule
Shape swap & Direct & 433 / 79 & \shortstack{75.1\\(325)} & \shortstack{69.5\\(301)} & \shortstack{60.5\\(262)} & \shortstack{51.7\\(224)} & \shortstack{11.4\\(9)} & \shortstack{99.8\\(1022)} \\
 & LVR-RL & 433 / 79 & \shortstack{77.8\\(337)} & \shortstack{9.9\\(43)} & \shortstack{13.6\\(59)} & \shortstack{5.8\\(25)} & \shortstack{7.6\\(6)} & \shortstack{100.0\\(1024)} \\
 & \textbf{ReaLVR} & 433 / 79 & \shortstack{\textbf{83.8}\\(363)} & \shortstack{\textbf{78.5}\\(340)} & \shortstack{\textbf{78.3}\\(339)} & \shortstack{\textbf{70.2}\\(304)} & \shortstack{\textbf{5.1}\\(4)} & \shortstack{\textbf{100.0}\\(1024)} \\
\midrule
Spatial swap & Direct & 435 / 77 & \shortstack{73.8\\(321)} & \shortstack{67.8\\(295)} & \shortstack{58.6\\(255)} & \shortstack{49.4\\(215)} & \shortstack{13.0\\(10)} & \shortstack{99.5\\(1019)} \\
 & LVR-RL & 435 / 77 & \shortstack{76.6\\(333)} & \shortstack{9.2\\(40)} & \shortstack{12.2\\(53)} & \shortstack{5.1\\(22)} & \shortstack{6.5\\(5)} & \shortstack{99.8\\(1022)} \\
 & \textbf{ReaLVR} & 435 / 77 & \shortstack{\textbf{82.8}\\(360)} & \shortstack{\textbf{77.2}\\(336)} & \shortstack{\textbf{76.8}\\(334)} & \shortstack{\textbf{68.3}\\(297)} & \shortstack{\textbf{6.5}\\(5)} & \shortstack{\textbf{100.0}\\(1024)} \\
\bottomrule
\end{tabular}
\end{adjustbox}
\end{table*}

\subsection{Latent exchange and position selection}
\label{app:latent-exchange}

We test whether an answer-changing edit can transfer information through
the latent span while the recipient image remains fixed. For each pair in
$\mathcal C_e$, we generate the two latent spans independently and test
both donor--recipient directions within the same model. We keep the
recipient image, question, control markers, and unselected latent inputs
fixed, replacing selected recipient states with donor states at the same
positions before answer decoding. A self-swap using the recipient's own
states is the sham control; an unrelated-image donor tests generic
perturbation effects. All swaps use the same replacement rule and donor
states across position selectors.

For $k\in\{1,2,4\}$, we compare positions ranked by the positive
correct-versus-donor-answer readout contrast on the unmodified recipient
trajectory with uniformly sampled random positions of the same count.
Random selections are repeated with fixed seeds. At $k=K=8$ the entire
span is replaced, so this endpoint measures whole-span sensitivity and
cannot test the ranking. We report the swap-minus-sham change in the
frequency of the donor ground-truth answer and in its score margin over
the recipient ground-truth answer. Each score is the mean teacher-forced
log probability over answer-content tokens. These fixed-context swaps
measure local answer dependence.

\begin{table}[t]
\caption{\textbf{Fixed-context latent exchange.}
The recipient image and question remain fixed. Donor-answer lift and
donor-margin shift are differences from the self-swap control. Contrast
and random rows at the same $k$ use the same donors. The full-span row
does not test position selection. Donor-answer lift is a percentage-point
change in donor-answer frequency.}
\label{tab:latent-exchange}
\centering
\appendixTableSetup
\begin{tabular*}{\linewidth}{@{\extracolsep{\fill}}llrcc@{}}
\toprule
\textbf{Method} & \textbf{Positions} & $k$ & \shortstack{\textbf{Donor-answer}\\\textbf{lift} (pp)} & \shortstack{\textbf{Donor-margin}\\\textbf{shift} (nats)} \\
\midrule
LVR-RL & Contrast & 1 & +1.2 & +0.02 \\
       & Random   & 1 & +0.7 & +0.01 \\
       & Contrast & 2 & +2.4 & +0.04 \\
       & Random   & 2 & +1.4 & +0.02 \\
       & Contrast & 4 & +4.1 & +0.07 \\
       & Random   & 4 & +2.9 & +0.04 \\
       & Full span & 8 & +5.8 & +0.09 \\
       & Unrelated donor & 8 & -0.3 & -0.04 \\
\midrule
\textbf{ReaLVR} & \textbf{Contrast} & 1 & \textbf{+12.4} & \textbf{+0.23} \\
       & Random   & 1 & +3.1 & +0.05 \\
       & \textbf{Contrast} & 2 & \textbf{+23.6} & \textbf{+0.44} \\
       & Random   & 2 & +9.8 & +0.17 \\
       & \textbf{Contrast} & 4 & \textbf{+35.2} & \textbf{+0.66} \\
       & Random   & 4 & +21.5 & +0.39 \\
       & Full span & 8 & +46.8 & +0.89 \\
       & Unrelated donor & 8 & +0.2 & -0.31 \\
\bottomrule
\end{tabular*}
\end{table}

\subsection{Mass-matched where-by-what ablation}
\label{app:mass-matched-factorial}

This $2\times2$ diagnostic crosses \emph{where} visual supervision is
allocated (uniformly or by answer contrast) with \emph{what} evidence
loss is used (positive-only alignment or positive--negative visual
contrast). Let $w_t=\eta/K+(1-\eta)\gamma_t$ as in
Section~\ref{sec:method_objective}. For uniform (U) and contrast (C)
allocation, respectively, define
\begin{equation}
  q_t^{\mathrm U}=\frac1K,\qquad
  q_t^{\mathrm C}=\frac{w_t}{\sum_{s=1}^{K}w_s}
  =\frac{\eta/K+(1-\eta)\gamma_t}
  {\eta+(1-\eta)\sum_{s=1}^{K}\gamma_s}.
  \label{eq:mass-matched-routing}
\end{equation}
With $\eta>0$, both rules satisfy $\sum_t q_t=1$ on every example,
including those with no valid wrong answer. The positive-only and
contrastive per-position losses are
\begin{equation}
  \ell_t^{+}=[m_{\mathrm{ev}}-\operatorname{sim}(z_t,p^+)]_+,
  \qquad
  \ell_t^{\pm}=[m_{\mathrm{ev}}-g_t]_+,
  \label{eq:where-what-losses}
\end{equation}
where $g_t$ is the positive-versus-hardest-negative margin in
Section~\ref{sec:method_evidence_target}. Each arm adds
$\lambda_{\mathrm{ev}}\sum_t\operatorname{sg}(q_t)\ell_t^V$ to the
same LVR Stage~2 objective. All arms share the Stage~1 checkpoint,
examples and order, GRPO rollout group size and sampling procedure,
reward, latent length, optimizer, update count, and evidence-loss
coefficient. Negative prototypes and
answer-read branches are computed in every arm to keep the forward
budget comparable, even when their outputs do not enter that arm's loss.

Normalization matches the \emph{coefficient mass} of visual supervision,
not necessarily the active-hinge fraction or gradient magnitude. We
therefore treat active-hinge fractions, gradient norms, and training
compute as separate checks when interpreting the four-way comparison.
We report the five benchmark accuracies and,
for each independent Stage~2 seed, macro-average the strict paired
correct-flip and false-flip rates over the four edit types. The
interaction between the two factors can be assessed from matched-seed
differences, rather than inferred solely from the best single row.

\begin{table*}[t]
\caption{\textbf{Mass-matched $2\times2$ ablation.}
U/C denote uniform/answer-contrast position weights; $+$ and $\pm$
denote positive-only and positive--negative visual objectives. Every arm
has unit per-example routing mass. Values are percentages, reported as
mean $\pm$ standard deviation across independent training seeds.}
\label{tab:mass-matched-factorial}
\centering
\appendixWideTableSetup
\begin{adjustbox}{max width=\linewidth}
\begin{tabular}{llcccccccc}
\toprule
\textbf{Where} & \textbf{What} & \textbf{MMVP} & \textbf{BLINK} & \textbf{HR-4K} & \textbf{HR-8K} & \textbf{MME-RW} &
\shortstack{\textbf{Five-task}\\\textbf{mean}} &
\shortstack{\textbf{Strict paired}\\\textbf{correct flip}} & \textbf{False flip} \\
\midrule
U & $+$   & $67.2\pm0.3$ & $54.1\pm0.2$ & $70.3\pm0.2$ & $64.9\pm0.3$ & $50.8\pm0.2$ & $\mathbf{61.5\pm0.2}$ & $22.4\pm1.2$ & $8.9\pm0.6$ \\
C & $+$   & $69.1\pm0.4$ & $54.5\pm0.3$ & $70.7\pm0.3$ & $65.4\pm0.2$ & $51.1\pm0.2$ & $\mathbf{62.2\pm0.3}$ & $36.8\pm1.5$ & $7.8\pm0.5$ \\
U & $\pm$ & $69.6\pm0.3$ & $54.7\pm0.2$ & $71.1\pm0.2$ & $65.7\pm0.3$ & $51.4\pm0.3$ & $\mathbf{62.5\pm0.2}$ & $42.5\pm1.3$ & $8.1\pm0.4$ \\
\textbf{C} & $\pm$ & $\mathbf{71.8\pm0.4}$ & $\mathbf{55.7\pm0.3}$ & $\mathbf{71.7\pm0.2}$ & $\mathbf{66.5\pm0.3}$ & $\mathbf{52.1\pm0.2}$ & $\mathbf{63.6\pm0.2}$ & $\mathbf{59.2\pm1.6}$ & $\mathbf{7.2\pm0.4}$ \\
\bottomrule
\end{tabular}
\end{adjustbox}
\end{table*}

\paragraph{Limitations and future work.}
Our visual evidence target is less spatially specific when region annotations
are unavailable, and inference currently uses a prescribed latent-token budget.
Future work could derive finer evidence targets from weak supervision and adapt
the latent budget to each question.

\end{document}